\documentclass{article}
\PassOptionsToPackage{numbers, compress}{natbib}

\usepackage{configuration/iclr2025_conference,times}
\usepackage[hyperindex,breaklinks,colorlinks,citecolor=darkblue]{hyperref} %

\usepackage{stfloats}
\usepackage[utf8]{inputenc}
\usepackage[T1]{fontenc}
\usepackage{booktabs}
\usepackage{amsfonts}       %
\usepackage{nicefrac}       %
\usepackage{microtype}      %
\usepackage{xcolor}
\usepackage[flushleft]{threeparttable}
\usepackage{color}
\usepackage{mathtools}
\usepackage{transparent}
\usepackage{url}
\usepackage{float}
\usepackage{subfigure}
\usepackage{graphicx}
\usepackage{multirow}
\usepackage{xspace}
\usepackage{natbib}
\usepackage{enumitem}
\usepackage[font=small]{caption}
\usepackage{diagbox}
\usepackage{wrapfig}
\usepackage[toc, page, header]{appendix}
\usepackage{amsmath,amsthm,amssymb}
\newtheorem{theorem}{Theorem}[section]

\newtheorem{remark}[theorem]{Remark}

\providecommand{\norm}[1]{\left\lVert#1\right\rVert}

\providecommand{\R}{\mathbb{R}} 

\DeclareMathOperator*{\argmax}{arg\,max}

\providecommand{\0}{\mathbf{0}}
\providecommand{\1}{\mathbf{1}}

\providecommand{\kk}{\mathbf{k}}

\providecommand{\ww}{\mathbf{w}}
\providecommand{\xx}{\mathbf{x}}
\providecommand{\yy}{\mathbf{y}}

\providecommand{\mG}{\mathbf{G}}

\providecommand{\mI}{\mathbf{I}}

\providecommand{\mR}{\mathbf{R}}

\providecommand{\mW}{\mathbf{W}}

\providecommand{\cL}{\mathcal{L}}

\newenvironment{talign*}
{\csname align*\endcsname}
{\endalign}
\usepackage{amssymb}
\usepackage{sidecap}

\usepackage{algorithm}
\usepackage[noend]{algpseudocode}

\newcommand{\mycaptionof}[2]{\captionof{#1}{#2}}

\errorcontextlines\maxdimen

\makeatletter
\newcommand*{\algrule}[1][\algorithmicindent]{\makebox[#1][l]{\hspace*{.5em}\thealgruleextra\vrule height \thealgruleheight depth \thealgruledepth}}%
\newcommand*{\thealgruleextra}{}
\newcommand*{\thealgruleheight}{.75\baselineskip}
\newcommand*{\thealgruledepth}{.25\baselineskip}

\newcount\ALG@printindent@tempcnta
\def\ALG@printindent{%
	\ifnum \theALG@nested>0
	\ifx\ALG@text\ALG@x@notext
	\else
		\unskip
		\addvspace{-1pt}
		\ALG@printindent@tempcnta=1
		\loop
		\algrule[\csname ALG@ind@\the\ALG@printindent@tempcnta\endcsname]%
		\advance \ALG@printindent@tempcnta 1
		\ifnum \ALG@printindent@tempcnta<\numexpr\theALG@nested+1\relax
			\repeat
		\fi
	\fi
}%
\usepackage{etoolbox}
\patchcmd{\ALG@doentity}{\noindent\hskip\ALG@tlm}{\ALG@printindent}{}{\errmessage{failed to patch}}
\makeatother

\newbox\statebox
\newcommand{\myState}[1]{%
	\setbox\statebox=\vbox{#1}%
	\edef\thealgruleheight{\dimexpr \the\ht\statebox+1pt\relax}%
	\edef\thealgruledepth{\dimexpr \the\dp\statebox+1pt\relax}%
	\ifdim\thealgruleheight<.75\baselineskip
		\def\thealgruleheight{\dimexpr .75\baselineskip+1pt\relax}%
	\fi
	\ifdim\thealgruledepth<.25\baselineskip
		\def\thealgruledepth{\dimexpr .25\baselineskip+1pt\relax}%
	\fi
	\State #1%
	\def\thealgruleheight{\dimexpr .75\baselineskip+1pt\relax}%
	\def\thealgruledepth{\dimexpr .25\baselineskip+1pt\relax}%
}

\usepackage{xspace}  
\usepackage{bm}

\usepackage{makecell}

\newcommand{\algmoe}{Dynamic Mixture of Experts\xspace}
\newcommand{\algabbr}{\textsc{DynMoE}\xspace}

\title{Dynamic Mixture of Experts: An Auto-Tuning Approach for Efficient Transformer Models}

\author{Yongxin Guo$^{1,}$\thanks{Equal contributions.}{\quad\,}Zhenglin Cheng$^{4,5,6,}$$^\ast${\quad\,}Xiaoying Tang$^{1,2,3,}$$^\dagger${\quad\,}Zhaopeng Tu$^8${\quad\,}Tao Lin$^{5,7,}$\thanks{Tao Lin and Xiaoying Tang are corresponding authors.}\\
$^1$School of Science and Engineering, The Chinese University of Hong Kong, Shenzhen 518172, China\\
$^2$Shenzhen Institute of Artificial Intelligence and Robotics for Society (AIRS), Shenzhen, China\\
$^3$Guangdong Provincial Key Laboratory of Future Networks of Intelligence, Shenzhen, China\\
$^4$Zhejiang University\quad$^5$School of Engineering, Westlake University\quad$^6$SII\\
$^7$Research Center for Industries of the Future, Westlake University\quad$^8$Tencent AI Lab
}

\iclrfinalcopy
\begin{document}

\definecolor{darkblue}{rgb}{0.0, 0.0, 0.55}
\definecolor{myblue}{rgb}{0,0.45,0.74}
\definecolor{myred}{rgb}{0.85,0.33,0.1}

\maketitle

\begin{abstract}
  The Sparse Mixture of Experts (SMoE) has been widely employed to enhance the efficiency of training and inference for Transformer-based foundational models, yielding promising results.
  However, the performance of SMoE heavily depends on the choice of hyper-parameters, such as the number of experts and the number of experts to be activated (referred to as top-$k$), resulting in significant computational overhead due to the extensive model training by searching over various hyper-parameter configurations.
  As a remedy, we introduce the \algmoe (\algabbr) technique.
  \algabbr incorporates (1) a novel gating method that enables each token to automatically determine the number of experts to activate.
  (2) An adaptive process automatically adjusts the number of experts during training.
  Extensive numerical results across Vision, Language, and Vision-Language tasks demonstrate the effectiveness of our approach to achieve competitive performance compared to GMoE for vision and language tasks, and MoE-LLaVA for vision-language tasks, while maintaining efficiency by activating fewer parameters.
  Our code is available at \url{https://github.com/LINs-lab/DynMoE}.
  \looseness=-1
\end{abstract}

\section{Introduction}

The scalable nature of Transformer models~\citep{kaplan2020scaling} has gained remarkable successes across a spectrum of applications, ranging from language~\citep{achiam2023gpt,touvron2023llama,touvron2023llama2} and vision~\cite{kirillov2023segment,peebles2023scalable} to cross-modality domains~\citep{liu2024visual,li2022blip,li2023blip}.
To further enhance performance while maintaining high efficiency, Sparse Mixture of Experts (SMoE) has emerged as a promising technique that significantly reduces computation costs during both training and inference stages~\citep{fedus2022switch,lepikhin2020gshard,zhang2022moefication}, and has been shown to achieve comparable or superior performance compared to traditional dense models~\citep{li2022sparse,jiang2024mixtral,dai2024deepseekmoe}.

\begin{figure}[!t]
    \centering

    \subfigure[Performance Fluctuation Illustration]{\includegraphics[width=.4\textwidth]{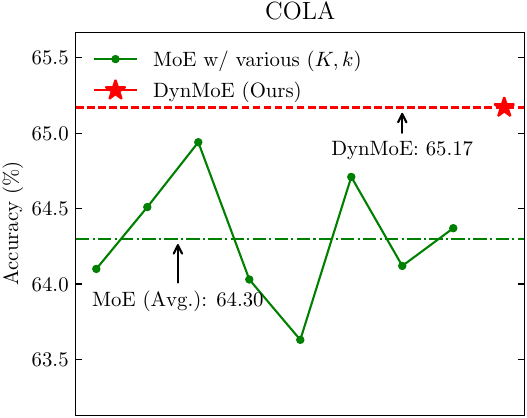}
        \label{fig:illustration-variance}
    }
    \subfigure[Performance-Efficiency Illustration]{\includegraphics[width=.44\textwidth]{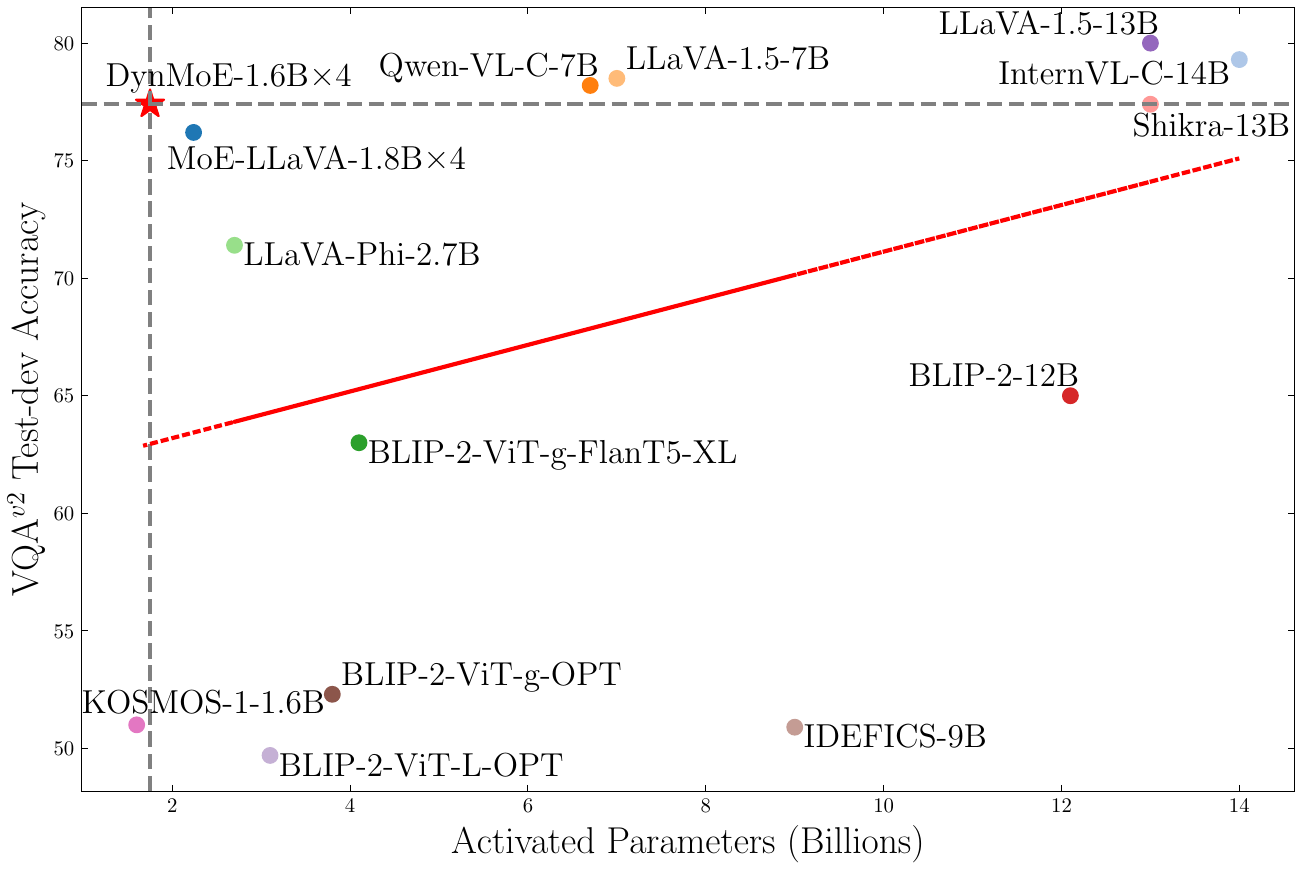}
        \label{fig:illustration-efficiency}
    }
    \caption{
        \textbf{Illustration of performance and efficiency of \algabbr.} In Figure~\ref{fig:illustration-variance}, we carried out experiments on GLUE benchmark~\citep{wang2018glue}, employing BERT-large~\citep{devlin2019bert} as backbone.
        In Figure\ref{fig:illustration-efficiency}, we follow the MoE-LLaVA~\citep{lin2024moe} settings, the $x$-axis represents the number of activated parameters, while the $y$-axis shows the performance on the Visual Question Answering (VQA) task.
    }
    \label{fig:illustration-sparse}
\end{figure}

Despite its success, SMoE has an unavoidable drawback: \textit{the performance of SMoE heavily relies on the choice of hyper-parameters}, such as the number of activated experts per token, referred as top-$k$, and the number of experts~\citep{clark2022unified,fan2024towards,yang2021m6}, denoted as $K$. As illustrated in Figure~\ref{fig:illustration-variance}, the performance discrepancy of MoE models under various configurations can be approximately 1\%-3\%.
Notably, \textit{identifying the optimal hyper-parameter without a sufficient number of ablation studies is challenging.}
As the size of the models continues to grow, this limitation could result in a significant waste of computational resources, and in turn, could hinder the efficiency of training MoE-based models in practice.
\looseness=-1

To tackle the above problems, the objective of this paper is to explore a novel training technique for MoE models, with the aim of addressing the following core question:

\begin{center}
    \textit{Is it possible to develop a MoE training strategy that can \textbf{automatically} determine the number of experts and the number of activated experts per token during the training process?}
\end{center}

Hence, we introduce the \algmoe (\algabbr) method, which addresses the aforementioned question through the introduction of two innovative components: (1) a top-any gating method that enables each token to autonomously determine the number of experts to activate, thereby allowing different tokens to activate varying numbers of experts; (2) an adaptive training process that dynamically adjusts the number of experts, increasing it when the current quantity is inadequate and removing redundant experts as necessary.
Additionally, we introduce a new auxiliary loss function specifically designed to encourage sparsity when employing the top-any gating approach. This loss encourages different experts to be diverse, rather than mandating that all experts be activated with the same frequency.
We summarize the contributions of this paper as follows:
\begin{itemize}[nosep, leftmargin=12pt]
    \item Introducing \algabbr, a novel method frees the burden of pivotal hyper-parameter selection for MoE training, which is capable of autonomously determining the number of experts and the number of experts to be activated per token.
    \item Conducting extensive empirical experiments across Vision, Language, and Vision-Language tasks.
          The results illustrate that \algabbr achieves comparable or superior performance and efficiency compared to the well-tuned MoE settings (Figure~\ref{fig:illustration-efficiency}).
\end{itemize}

\section{Related Works}
The Sparse Mixture of Experts (SMoE) approach~\citep{eigen2013learning,shazeer2017outrageously,lepikhin2020gshard} has been proven to effectively enhance the training and inference efficiency of foundational models. Contemporary studies primarily modify the MLP layer of transformer models into multiple expert models and employ a gating network to determine which expert to select. They only choose a subset of experts for each token during both training and inference~\citep{lepikhin2020gshard,fedus2022switch}. Recently, the SMoE structure has shown success in various research areas. For instance, GMoE~\citep{li2023sparse} has demonstrated that SMoE can enhance generalization performance in vision tasks. Large Language Models (LLMs) have also employed MoE to simultaneously reduce training and inference costs while improving model performance~\citep{fedus2022switch,jiang2024mixtral,dai2024deepseekmoe,ren2023pangu,lin2024moe}. However, most of these models employ standard SMoE structures and apply the SMoE to various tasks. Our paper focuses on improving the MoE training process, which can be easily integrated with these methods.

Recently, some attempts have been made to improve the architecture of MoE models. For example, researchers have investigated the benefits of sample-wise~\citep{ramachandran2018diversity,gross2017hard} and token-wise~\citep{shazeer2017outrageously,riquelme2021scaling,fedus2022switch} routing. Some studies introduce load balancing loss to ensure that the experts are activated an equal number of times~\citep{lepikhin2020gshard,fedus2022switch}. Expert choice routing~\citep{zhou2022mixture} addresses load balance by allowing experts to choose tokens; however, this approach also suffers from dropped tokens. SoftMoE~\citep{puigcerver2023sparse} uses a slot mechanism to simultaneously resolve the issues of load balance and dropped tokens.
Nevertheless, these approaches also require pre-defined hyperparameters, such as the number of experts or the number of experts to be activated.
Some studies enable tokens to activate a varying number of experts~\citep{huang2024harder,yang2024xmoe,huang2024harder,yang2024xmoe}. However, these approaches either rely on modifying the routing mechanism from top-$k$ to top-$p$ (which introduces the additional hyperparameter $p$), or use dense training during the initial stages, neither of which provide an optimal implementation.
In this paper, we tackle this problem by presenting \algabbr, an algorithm that automatically determines the number of activated experts for each token and dynamically adds or removes experts during the training process. Furthermore, we introduce a new auxiliary loss function that ensures sparsity when utilizing the \algabbr algorithm.

\section{Method}

\begin{wrapfigure}{r}{.45\textwidth}
    \centering
    \includegraphics[width=.45\textwidth]{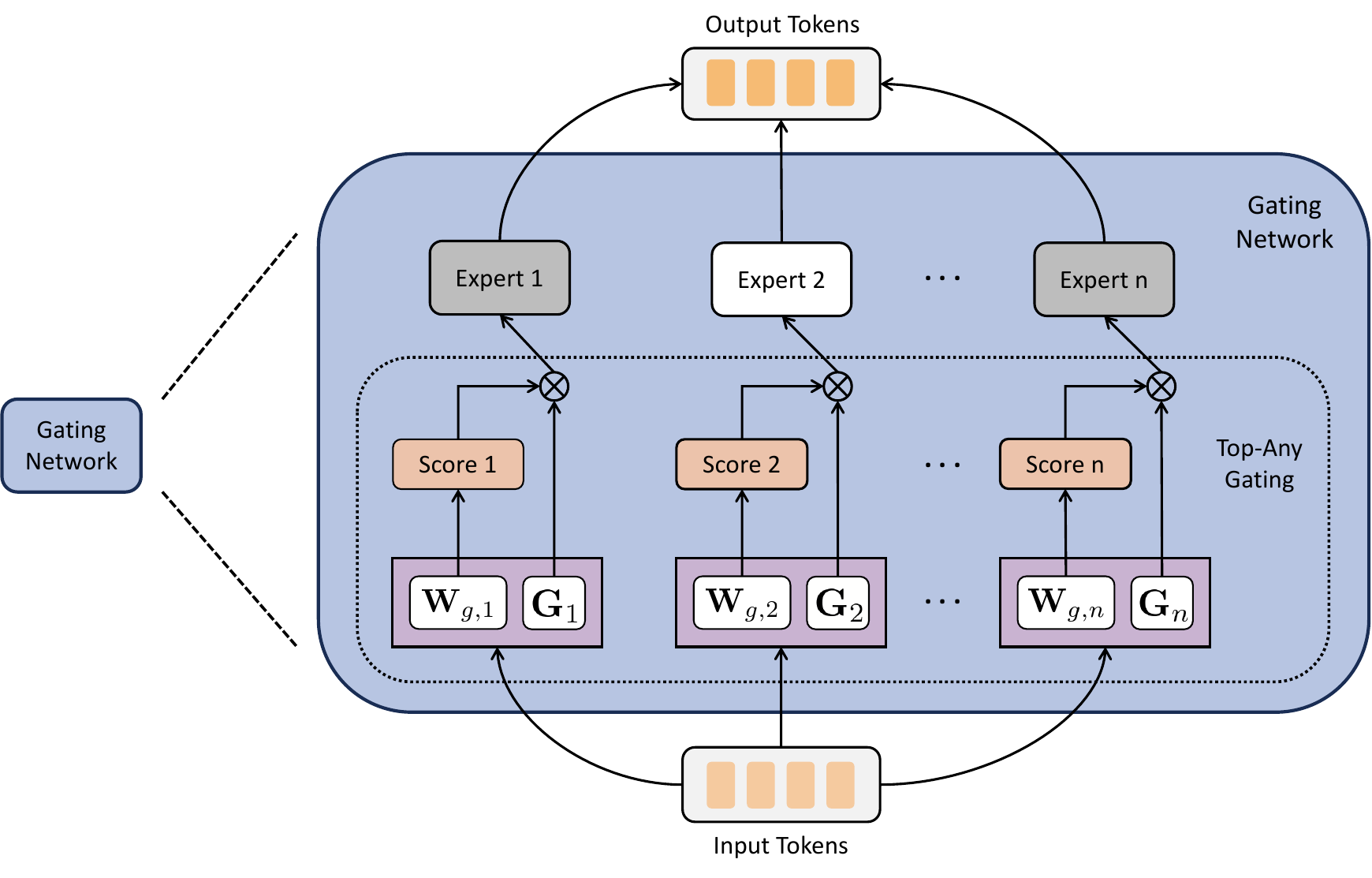}
    \caption{
        \small
        \textbf{Illustration of the top-any gating method.}
        The input tokens pass through the gating weights $\mW_{g,e}$ corresponding to each expert $e$, obtaining the gating scores.
        The scores surpass gates $\mG_e$ will activate the subsequent expert.
        Finally, the expert outputs are combined to produce the output tokens.
        \looseness=-1
    }
    \label{fig:top-any-gating}
    \vspace{-1em}
\end{wrapfigure}

In this section, we introduce the \algmoe (\algabbr), an algorithm capable of automatically determining the number of experts and the number of experts to be activated for both training and inference stages.
This is achieved through the incorporation of two crucial components:
\begin{itemize}[nosep, leftmargin=16pt]
    \item[(1)] \emph{The top-any gating method} (Figure~\ref{fig:top-any-gating}), which models the gating mechanism as a multi-label classification problem, allowing tokens to decide the number of experts to be activated on their own.
          This enables different tokens to activate varying numbers of experts, including the option to activate no experts.
    \item[(2)] \emph{A carefully designed adaptive process} that adds new experts when tokens choose to not activate any existing experts, and removes any surplus experts that have not been activated by any tokens.
\end{itemize}
The overall process is summarized in Algorithm~\ref{alg:algorithm-framework-general}.

\subsection{Top-Any Gating}
In this section, we present the superior gating method to eliminate the need for tuning the top-$k$ value.
We further improve the test-time inference procedure and introduce an additional auxiliary loss to prevent token dropping and boost efficiency.

\paragraph{Traditional top-$k$ gating and the limitations.}
The traditional top-$k$ gating method takes the token embedding $\xx$ as input and employs an additional gating network $g$ to predict the gating scores. These gating scores are then used to determine which experts will be activated for the input tokens.
Typically, given token $\xx \in \R^{d}$ as input, the gating process is defined as the follows~\citep{rajbhandari2022deepspeed,hwang2023tutel}:
\begin{align}
    g(\xx) \in \R^{K} := \text{softmax}(\mW_g^{T} \xx) \,,
\end{align}
where $\mW_g \in \R^{d \times K}$ is the parameter of the gating network, and $K$ is the number of experts.
Then the output of the MoE layer is defined by
\begin{align}
    \yy = \frac{1}{\sum_{e \in \text{Top-}k \left( g(\xx) \right) } g(\xx)_e } \sum_{e \in \text{Top-}k \left( g(\xx) \right) } g(\xx)_e E_e(\xx) \,,
\end{align}
where $ E_e(\xx) \in \R^{d} $ is the output of $e$-th expert given input $\xx$, and $g(\xx)_e$ is the $e$-th entry of $g(\xx)$.

Despite the considerable success of the top-$k$ gating method in enhancing training and inference efficiency, two limitations persist:
\begin{enumerate}[nosep, leftmargin=12pt]
    \item \emph{The value of $k$ must be fine-tuned to optimize model performance.}
          As demonstrated in Figure~\ref{fig:illustration-variance}, the performance of MoE models can vary significantly with different top-$k$ values.
          This observation has also been noted in recent studies~\citep{clark2022unified, fan2024towards, yang2021m6}.
          Consequently, substantial computational resources are needed to identify the optimal value of $k$.
    \item \emph{The top-$k$ gating approach assumes that each token must activate the same number of experts, which may not always hold in practice.}
          For instance, when considering different tasks, there could exist tokens shared by all tasks and those specific to certain tasks, i.e. different tokens could activate different numbers of experts.
\end{enumerate}

\paragraph{Addressing the limitations of top-$k$ gating by tuning-free top-any gating.}
To address the aforementioned limitations, we propose the \emph{\textbf{top-any gating method}}, which does not require a pre-defined value of $k$ and allows different tokens to activate varying numbers of experts during both training and inference stages.

The design of the top-any gating method draws inspiration from the multi-label classification problem.
We consider each expert as an individual class and calculate the classification (gating) score for each class (expert) independently.
Subsequently, all classes (experts) with scores exceeding the threshold are deemed positive (activated).
In detail, given the expert representation matrix $\mW_g \in \R^{d \times K}$, where the $k$-th row of $\mW_g$ acts as the representation of expert $k$, and an input token $\xx \in \R^{d}$, the key steps of top-any gating can be formulated by the following equation:
\begin{align}
    s(\xx) & = \frac{\left \langle \xx, \mW_{g} \right \rangle}{\norm{\xx} \norm{\mW_{g}}} \,, \label{equ:gating-sim} \\
    g(\xx) & = \text{sign} \left( \sigma \left( s(\xx) \right) - \sigma( \mG ) \right) \,, \label{equ:gating-score}
\end{align}
where $\mW_g \in \R^{d \times K}$ and $\mG \in \R^{K}$.
To illustrate, we first compute the cosine similarities between the token and the expert representation matrix $\mW_g$ and obtain the similarity score $s(\xx) \in \R^{K}$.
Then the sigmoid function $\sigma$ is applied to the similarity score $s(\xx)$ to obtain the scores between $0$ and $1$.
Finally, experts with similarity scores greater than the trainable per-expert threshold $\mG$ are considered to activate experts for the token $\xx$.
It is important to note that the sign function does not support back-propagation, and thus we customize the back-propagation process of this part by directly copying the gradient of $g(\xx)$ to $\sigma \left( s(\xx) \right) - \sigma ( \mG )$ to effectively bypass the sign function.

Given the gating score $g(\xx) \in \R^{K}$, the number of activated experts is then defined by
\begin{align}
    k := \text{sum} \left( g(\xx) \right)  \,, \label{equ:gating-k}
\end{align}
where $k$ represents the number of experts to be activated for token $\xx$.
The model output of the MoE layer with the top-any gating method can be derived as follows
\begin{align}
    \yy = \frac{1}{k} \sum_{g(\xx)_e > 0} E_{e}(\xx) \,. \label{equ:gating-outputs}
\end{align}

\begin{remark}[Discussion on not to consider the magnitude of scores when averaging the expert outputs.]
    \label{remark:not-use-weights}
    In our top-any gating approach, the scores of different experts are calculated independently. As a result, the scores of different experts may have different scales and ranges. For instance, there may be cases where the scores of Expert 1 are within the range of (0.1, 0.2), but the scores of Expert 2 are within the range of (0.8, 0.9). To avoid this mismatch, we have decided not to consider the magnitude of scores in Equation~\eqref{equ:gating-outputs}. Ablation studies can be found in Table~\ref{tab:weights-ablations}.
\end{remark}

\paragraph{Improving the top-any gating during test-time to prevent token dropping.}
To facilitate the design of the adaptive expert number process, we did not impose a minimum value on $k$.
Consequently, some tokens may not activate any experts.
To address this issue, during model performance evaluation, we modify the top-any gating to enable top-$1$ gating for tokens that do not choose to activate any experts.
In detail, for the input token $\xx$ with $\text{sum}(g(\xx)) = 0$, the modified gating score $\tilde{g}(\xx)$ is obtained by
\begin{align}
    \tilde{g}(\xx)_k =
    \begin{split}
        \left \{            \begin{array}{ll}
                                0               & k \not = \argmax_{k} \sigma (s(\xx)) \,, \\
                                \sigma (s(\xx)) & k = \argmax_{k} \sigma (s(\xx)) \,.
                            \end{array}
        \right.
    \end{split}
\end{align}

\paragraph{Guarding efficiency for top-any gating by auxiliary loss.}
The primary goal of using MoE models is to improve the training and inference efficiency.
However, in the absence of a cap on the maximum number of activated experts, tokens might activate all experts, which is counterproductive to our primary goal.

Using an auxiliary loss as a regularization over experts may alleviate our issue.
However, existing auxiliary loss methods~\citep{lepikhin2020gshard,fedus2022switch,wu2024multi} are primarily designed to ensure load balancing across experts and thus cannot align with our objectives.
While activating all experts can indeed achieve load balancing, it contradicts our aim of improving efficiency by limiting the number of activated experts.
Therefore, we need a solution that not only ensures load balancing but also restricts the number of activated experts~\footnote{We also conducted experiments incorporating other auxiliary losses with \algabbr, as shown in Table~\ref{tab:dynmoe-load-balance-efficiency-sparse}.}.
\looseness=-1

As a remedy, we propose a new auxiliary loss, namely \emph{sparse and simple gating loss}, as shown in~\eqref{equ:gating-loss}.
The \emph{diversity loss} and \emph{simplicity loss} in~\eqref{equ:gating-loss} work together to improve the efficiency of the model by addressing different aspects of the expert representations.
On one hand, the \emph{diversity loss} encourages independence among the $\mW_g$ representations of various experts.
It serves two purposes: First, it prevents a high degree of similarity between experts, thereby enhancing the model's representational capacity;
Second, it guides tokens to avoid simultaneous activation of all experts, thereby promoting sparse gating for improved efficiency.
On the other hand, the \emph{simplicity loss} normalizes $\mW_g$ to avoid excessively large values within the matrix, which helps maintain numerical stability and prevents overfitting due to extreme parameter values.
The detailed loss function is defined as follows:
\begin{align}
    \textstyle
    \cL = \underbrace{\norm{\mW_g^{T} \mW_g - \mI_K}_2}_{\emph{diversity loss}} + \underbrace{\frac{1}{K} \sum_{e=1}^{K} \norm{\ww_{g, e}}_2}_{\emph{simplicity loss}} \,, \label{equ:gating-loss}
\end{align}
where $\mI_K$ is the identity matrix with dimension $K$, and $\ww_{g, e} \in \R^{d}$ is the $e$-th element of $\mW_g$, indicating the representation of the $e$-th expert.

\subsection{Adaptive Training Process}

\begin{figure}
    \centering
    \includegraphics[width=0.7\textwidth]{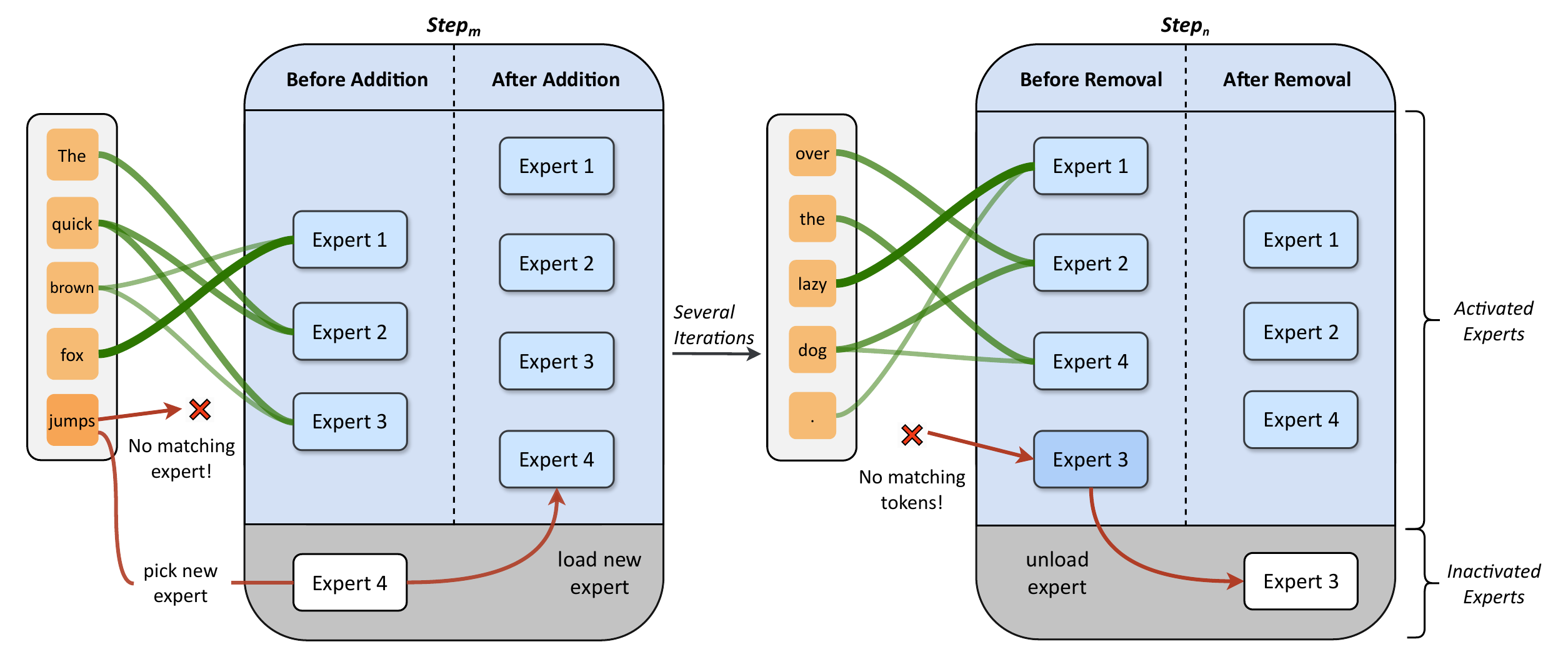}
    \caption{
        \textbf{Elaboration on the adaptive training process.}
        We visualize the adaptive training process of \algabbr, including record routing, experts adding, and experts removing.
        The green strip connecting the token and the expert indicates records of a token routing to an expert.
        The red arrow at the bottom part of the figure shows where and when expert addition and removal happens.
    }
    \label{fig:adaptive-proc}
\end{figure}

In this section, we elaborate on the adaptive training process, which is designed to automatically determine the number of experts.
As illustrated in Figure~\ref{fig:adaptive-proc}, the adaptive process consists of three parts, namely
(1) \emph{Routing Recording}: recording the routing results during training;
(2) \emph{Adding Experts}: adding new experts when tokens choose not to activate any existing experts;
and (3) \emph{Removing Experts}: removing experts that have not been chosen by any tokens.
To promising efficiency and avoiding burden communication, we only check if experts required to be added or removed every 100-300 iterations.

\paragraph{\emph{Routing Recording}.}
To facilitate the removal and addition of experts, it is essential to track the routing status.
Specifically, we record two key pieces of information for each MoE layer: (1) For each expert $e$, we record the time at which expert $e$ is activated, denoted as $\mR_{E} \in \R^{K}$ (as shown in Line 9 of Algorithm~\ref{alg:algorithm-framework-general}).
(2) For input data that does not activate any expert, we compute the sum of their embeddings $\xx$ as $\mR_{S} \in \R^{d}$ (as outlined in Line 10 of Algorithm~\ref{alg:algorithm-framework-general}).
Note that this approach simplifies the expert addition process: by using the token embeddings to initialize the expert representation $\mW_g$, we can achieve a high similarity score between these tokens and the new experts, ensuring that the new expert will be activated by these tokens when added.

As demonstrated in Algorithm~\ref{alg:algorithm-framework-general}, we utilize $flag_{s}$ and $flag_{f}$ to determine when to start and stop routing recording.
Users can control these two flags as needed.

\paragraph{\emph{Adding Experts} when there exist tokens that choose not to activate any experts.}
We add new experts when the recorded $\mR_{S} \not = \mathbf{0}$, as some tokens do not activate any experts and $\mR_{S}$ is the sum of these tokens.
Therefore, given $K$ activated experts and new expert $K + 1$, we initialize $\mW_{g, K + 1} = \frac{\mR_{S}}{\norm{\mR_{S}}}$ and $\mG_{K+1} = \mathbf{0}$.
Moreover, due to the device constrain, the maximum number of experts should be constrained. We set the maximum number of experts to 16 for vision and language tasks, and 4 for vision-language tasks in practice. Discussions on additional strategies for initializing new experts can be found in Appendix~\ref{sec:additiona-vision-task}.

\paragraph{\emph{Removing Experts} when there exist experts not activated by any token.}
We remove experts when there is an expert $e$ such that $\mR_{E}^{e} = \mathbf{0}$ (as shown in Line 13 in Algorithm~\ref{alg:algorithm-framework-general}), which indicates that there is no token choose to activate the expert $e$.

\section{Experiments}
\setlength{\parskip}{0.4pt plus0.4pt minus0.4pt}
In this section, we carry out experiments to address the following questions:
\begin{itemize}[leftmargin=12pt,nosep]
    \item \textbf{Q1}: Can \algabbr achieve competitive performance among different MoE settings? See \ref{answer1}.
    \item \textbf{Q2}: Can \algabbr handle tasks with varying modalities and scales? See \ref{answer2}.
    \item \textbf{Q3}: Will the model trained by \algabbr maintain sparsity to ensure efficiency? See \ref{answer3}.
    \item \textbf{Q4}: Can \algabbr offer insights that could guide the design of MoE models? See \ref{answer4}.
\end{itemize}
Additional numerical results, including (1) detailed results on vision and language tasks, (2) ablation studies on auxiliary losses, (3) comparison to top-p gating baselines, (4) pretraining and fine-tuning results on more vision tasks, (5) training efficiency evaluation, (6) overhead of introducing top-any gating, and (7) ablation studies on the weights for averaging expert outputs, can be found in the Appendix~\ref{sec:additional-experiments}.

\subsection{Experiment Setup} To answer the above four questions, we conduct experiments on Vision, Language, and Vision-Language tasks. The details are shown in the following.
\begin{itemize}[leftmargin=12pt, nosep]
    \item \textbf{Vision Task.} For the vision tasks, we follow the same settings as in GMoE~\citep{li2023sparse}. We employ the pre-trained ViT-S/16~\cite{dosovitskiy2020image} model and evaluate it on the DomainBed~\citep{gulrajani2020search} benchmark.
          Our experiments encompass four Domain Generalization datasets: PACS~\citep{li2017deeper}, VLCS~\citep{albuquerque2019generalizing}, OfficeHome~\citep{venkateswara2017deep}, and DomainNet~\citep{peng2019moment}.
          All results are reported using the train-validation selection criterion.
    \item \textbf{Language Task.} The language tasks adhere to the same settings as those in MoEfication~\citep{zhang2022moefication} and EMoE~\citep{qiu2023emergent}.
          The MoE models are built upon the BERT-large~\citep{devlin2019bert} architecture using the MoEfication method and are fine-tuned on GLUE~\citep{wang2018glue} tasks, which include COLA~\citep{warstadt2019neural}, QNLI~\citep{wang2018glue}, RTE~\citep{bentivogli2009fifth}, MNLI~\citep{xu2020clue}, and MRPC~\citep{dolan2005automatically}.
          For each MoE setting, we tune the learning rates in \{2e-5, 3e-5, 5e-5\} and report the best results.
    \item \textbf{Vision-Language Task.} The vision-language tasks follows the setting in MoE-LLaVA~\citep{lin2024moe}, where we use StableLM-2-1.6B~\citep{bellagente2024stable}, Qwen-1.8B~\citep{bai2023qwen} and Phi-2-2.7B~\citep{hughesPhi2SurprisingPower2023} as backbone language models, and use clip-vit-large-patch14-336~\citep{radford2021learning} as the vision encoder.
          The models are evaluated on image understanding benchmarks including VQA-v2~\citep{goyal2017making}, GQA~\citep{hudson2019gqa}, VisWiz~\citep{gurari2018vizwiz}, ScienceQA-IMG~\citep{lu2022learn}, TextVQA~\citep{singh2019towards}, POPE~\citep{li2023evaluating}, MME~\citep{yin2023survey}, MMBench~\citep{liu2023mmbench}, LLaVA-Bench (in-the-Wild)~\citep{liu2024visual}, and MM-Vet~\citep{yu2023mm}. Furthermore, we keep routing records in our model during testing time. For each benchmark, we collect the number of experts' activations per MoE layer and total processed tokens during testing. The hyper-parameter settings are the same to MoE-LLaVA for fair comparision.
\end{itemize}

\subsection{\label{answer1}A1: \algabbr Achieves Competitive Performance among Various MoE Settings}
\begin{figure}
    \centering
    \includegraphics[width=\textwidth]{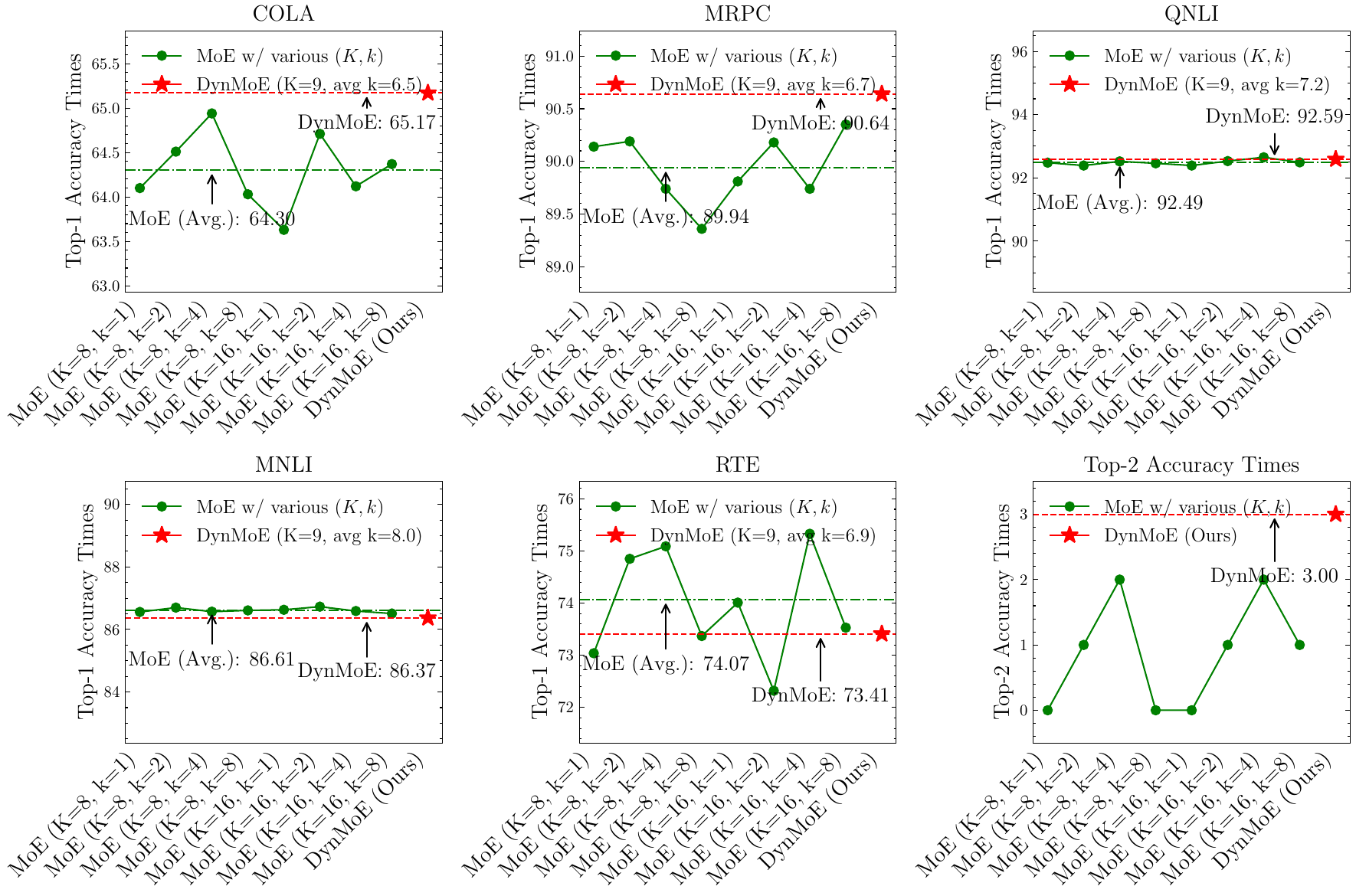}
    \caption{
        \textbf{Performance of \algabbr on language tasks.}
        We conduct experiments on the GLUE benchmark.
        The $x$-axis represents MoE settings with varying $K$ and top-$k$ values.
        The $y$-axis denotes the model's performance.
        Dashed lines indicate the average performance across different settings, as well as the performance of \algabbr. For all the MoE settings, we tune the learning rates in \{2e-5, 3e-5, 5e-5\} and report the best results. We also report the times when each MoE setting attains the top-2 best results across all configurations.
        \looseness=-1
    }
    \label{fig:language}
\end{figure}

In this section, we carry out experiments on the GLUE benchmark~\citep{wang2018glue}, varying the number of experts ($K$) and the value of top-$k$.
The results of these experiments can be observed in Figure~\ref{fig:language}.
More detailed results of each MoE setting can be found in Tables~\ref{tab:detailed-results-cola}-\ref{tab:detailed-results-rte} of Appendix.
\looseness=-1

\paragraph{The performance of \algabbr surpasses the average performance among various MoE settings.}
As seen in Figure~\ref{fig:language}, we can observe that
\begin{enumerate}[nosep, leftmargin=12pt]
    \item The \algabbr outperforms the average performance for various $K$ and top-$k$ values in most tasks. \algabbr also achieves the highest number of top-1/2 best performances among all MoE settings, demonstrating its competitive performance.
    \item The performance fluctuates considerably with different $K$ and top-$k$ values, such as up to 3.0\% on the RTE task and 1.3\% on the COLA task.
          \algabbr overcomes this issue by not requiring pre-defined $K$ and top-$k$ values.
    \item  The performance gain of specific $K$ and top-$k$ choice is not consistent among tasks.
          For instance, the $K = 16, k = 4$ setting performs well on QNLI but poorly on MRPC.
          In contrast, the \algabbr always achieve competitive performance among tasks.
\end{enumerate}

\subsection{\label{answer2}A2: \algabbr Can Handle Vision, Language, and Vision-Language Tasks}

In addition to Language tasks, we also conduct experiments on Vision and Vision-Language tasks to verify the performance of \algabbr on different modalities and task scales.
The results can be found in Tables~\ref{tab:vision}, and~\ref{tab:vision-language}.

\paragraph{The effectiveness of \algabbr remains consistent in both Vision and Vision-Language tasks.}
Compared to the standard MoE,
we can observe the following:
\textbf{A.} \algabbr outperforms standard MoE with well-tuned learning rate, number of experts, and top-$k$~\citep{qiu2023emergent} in Vision tasks. The performance difference between \algabbr and another well-tuned MoE setting in~\citep{li2023sparse}, falls within the range of random fluctuation.
\textbf{B.} When using StableLM-1.6B and Phi-2-2.7B as the backbone, the performance of \algabbr-LLaVA surpasses that of MoE-LLaVA.
\textbf{C.} With Qwen-1.8B as the backbone, the performance of \algabbr-LLaVA remains comparable to MoE-LLaVA. In this setting, the average top-$k$ of \algabbr-LLaVA (avg $k = 1.86$) is also close to the MoE-LLaVA setting ($k=2$).
\textbf{D.} In the BERT experiments (Figure~\ref{fig:language}), \algabbr generally activate more experts for each token compared to larger scale MoE-LLaVA experiments (Table~\ref{tab:vision-language}). This observation aligns with the BERT experiments results obtained when using a fixed k value, i.e., k=4 generally performs better among the set \{1,2,4,8\}.
\looseness=-1

\begin{table}[!t]
    \centering
    \caption{
        \textbf{Performance of \algabbr on vision tasks}:
        Our study investigates the performance of \algabbr on vision tasks using the DomainBed benchmark, with ViT-small serving as the backbone model.
        The effectiveness of GMoE is elucidated based on meticulously tuned results as presented in the previous works~\cite{li2023sparse} and~\cite{qiu2023emergent}.
        In our implementation of \algabbr, we configure the maximum number of experts to $8$, with an initial setting of 6 experts.
        The number of experts is dynamically adjusted in each iteration for \algabbr.
        We also report the performance of \algabbr using Gshard loss~\citep{lepikhin2020gshard} as the auxiliary loss.
    }
    \resizebox{.9\textwidth}{!}{
        \begin{tabular}{l c c c c c}
            \toprule
            Algorithms                                           & PACS & VLCS & OfficeHome & DomainNet & Average \\
            \midrule
            GMoE (in~\cite{li2023sparse})                        & 88.1 & 80.2 & 74.2       & 48.7      & 72.8    \\
            GMoE (carefully tuned~\citep{qiu2023emergent})       & 87.7 & 79.6 & 73.1       & -         & -       \\
            \midrule
            GMoE (with \algabbr, Gshard Loss)                    & 88.4 & 79.4 & 73.6       & 47.4      & 72.2    \\
            GMoE (with \algabbr, Diverse and Simple Gating Loss) & 87.6 & 80.3 & 73.5       & 48.2      & 72.4    \\
            \bottomrule
        \end{tabular}
    }
    \label{tab:vision}
\end{table}

\begin{table}[!t]
    \centering
    \caption{
        \textbf{Performance of \algabbr on vision-language tasks}: Our study investigates the performance of \algabbr-LLaVA on image understanding benchmarks.
        Evaluation Benchmarks include VQA-v2; GQA; VisWiz; SQA$^I$ (ScienceQA-IMG); VQA$^T$ (TextVQA); POPE; MME; MMB (MMBench); LLaVA$^W$ (LLaVA-Bench (in-the-Wild)); MM-Vet.
        For a fair comparison, we set the maximum number of experts to 4 for \algabbr-LLaVA (the same as the number of experts in MoE-LLaVA) and set the initial number of experts to $2$.
        $N_{A}$ indicates the number of activated parameters.
    }
    \resizebox{1.\textwidth}{!}{
        \begin{tabular}{l|c c|ccccc|ccccc}
            \toprule
            Algorithms                           & $N_{A}$                    & VQA$^{v2}$                  & GQA                         & VisWiz                      & SQA$^I$                     & VQA$^T$                     & POPE                        & MME                           & MMB                         & LLaVA$^W$                   & MM-Vet                      \\
            \midrule
            \textit{Dense}                                                                                                                                                                                                                                                                                                                                                                  \\
            \makecell[r]{LLaVA-1.5 (Vicuna-13B)} & \textcolor{lightgray}{13B} & \textcolor{lightgray}{80.0} & \textcolor{lightgray}{63.3} & \textcolor{lightgray}{53.6} & \textcolor{lightgray}{71.6} & \textcolor{lightgray}{61.3} & \textcolor{lightgray}{85.9} & \textcolor{lightgray}{1531.3} & \textcolor{lightgray}{67.7} & \textcolor{lightgray}{70.7} & \textcolor{lightgray}{35.4} \\
            \makecell[r]{LLaVA-1.5 (Vicuna-7B)}  & \textcolor{lightgray}{7B}  & \textcolor{lightgray}{78.5} & \textcolor{lightgray}{62.0} & \textcolor{lightgray}{50.0} & \textcolor{lightgray}{66.8} & \textcolor{lightgray}{58.2} & \textcolor{lightgray}{85.9} & \textcolor{lightgray}{1510.7} & \textcolor{lightgray}{64.3} & \textcolor{lightgray}{63.4} & \textcolor{lightgray}{30.5} \\
            \makecell[r]{LLaVA-Phi (Phi-2-2.7B)} & 2.7B                       & 71.4                        & -                           & 35.9                        & 68.4                        & 48.6                        & 85.0                        & 1335.1                        & 59.8                        & -                           & 28.9                        \\
            \midrule
            \textit{Sparse (StableLM-1.6B)}                                                                                                                                                                                                                                                                                                                                                 \\
            \makecell[r]{MoE-LLaVA                                                                                                                                                                                                                                                                                                                                                          \\($K=4,k=2$)} & \makecell{2.06B} &  76.7 & 60.3 & 36.2 & 62.6 & 50.1 & 85.7 & 1318.2 & 60.2 & 86.8 & 26.9 \\
            \cmidrule(lr){2-12}
            \makecell[r]{\algabbr-LLaVA                                                                                                                                                                                                                                                                                                                                                     \\(avg $k = 1.25$)} & 1.75B & 77.4 & 61.4 & 40.6 & 63.4 & 48.9 & 85.7 & 1300.9 & 63.2 & 86.4 & 28.1 \\
            \midrule
            \textit{Sparse (Qwen-1.8B)}                                                                                                                                                                                                                                                                                                                                                     \\
            \makecell[r]{MoE-LLaVA                                                                                                                                                                                                                                                                                                                                                          \\($K=4,k=2$)} & \makecell{2.24B} & 76.2 & 61.5 & 32.6 & 63.1 & 48.0 & 87.0 & 1291.6 & 59.7 & 88.7 & 25.3 \\
            \cmidrule(lr){2-12}
            \makecell[r]{\algabbr-LLaVA                                                                                                                                                                                                                                                                                                                                                     \\(avg $k = 1.86$)} & 2.19B & 76.4 & 60.9 & 32.4 & 63.2 & 47.5 & 85.8 & 1302.4 & 61.3 & 89.2 & 24.2 \\
            \midrule
            \textit{Sparse (Phi-2-2.7B)}                                                                                                                                                                                                                                                                                                                                                    \\
            \makecell[r]{MoE-LLaVA                                                                                                                                                                                                                                                                                                                                                          \\($K=4,k=2$)} & \makecell{3.62B} & 77.6 & 61.4 & 43.9 & 68.5 & 51.4 & 86.3 & 1423.0 & 65.2 & 94.1 & 34.3\\
            \cmidrule(lr){2-12}
            \makecell[r]{\algabbr-LLaVA                                                                                                                                                                                                                                                                                                                                                     \\(avg $k = 1.68$)} & 3.35B & 77.9 & 61.6 & 45.1 & 68.0 & 51.8 & 86.0 & 1429.6 & 66.6 & 95.6 & 33.6 \\
            \bottomrule
        \end{tabular}
    }
    \normalsize
    \label{tab:vision-language}
\end{table}

\subsection{\label{answer3}A3: \algabbr Maintains Efficiency by Activating Less Parameters}
In this section, we aim to demonstrate that although we did not enforce sparsity on the \algabbr models, the trained \algabbr models are still sparse, promising improved inference efficiency.

\paragraph{\algabbr-LLaVA activates fewer parameters compared to MoE-LLaVA.}
In Table~\ref{tab:vision-language}, we display the number of activated parameters in the "$N_A$" column.
When using StabeLM-1.6B as the backbone, \algabbr-LLaVA activates approximately 15.0$\%$ fewer parameters than MoE-LLaVA.
For Qwen-1.8B, \algabbr-LLaVA activates about 2.2$\%$ fewer parameters than MoE-LLaVA.
For Phi-2-2.7B, \algabbr-LLaVA activates about 7.5$\%$ fewer parameters than MoE-LLaVA.
In these three cases, the reduction in activated parameters does not compromise the model's performance.

\paragraph{Ablation studies on the value of top-$k$ during test.}
In Table~\ref{tab:ablation-test-top-k}, we examine the performance of \algabbr-LLaVA when using different top-$k$ values during the testing phase.
The results indicate that (1) The original \algabbr-LLaVA outperforms other settings in most cases while activating the fewest number of parameters.
(2) Compared to the StableLM-1.6B backbone, \algabbr-LLaVA trained with the Qwen-1.8B backbone sometimes favors activating two experts.
This observation aligns with the fact that \algabbr-LLaVA also chooses to activate about $2$ experts (see Table~\ref{tab:vision-language}).

\begin{figure}
    \centering
    \includegraphics[width=1.0\textwidth]{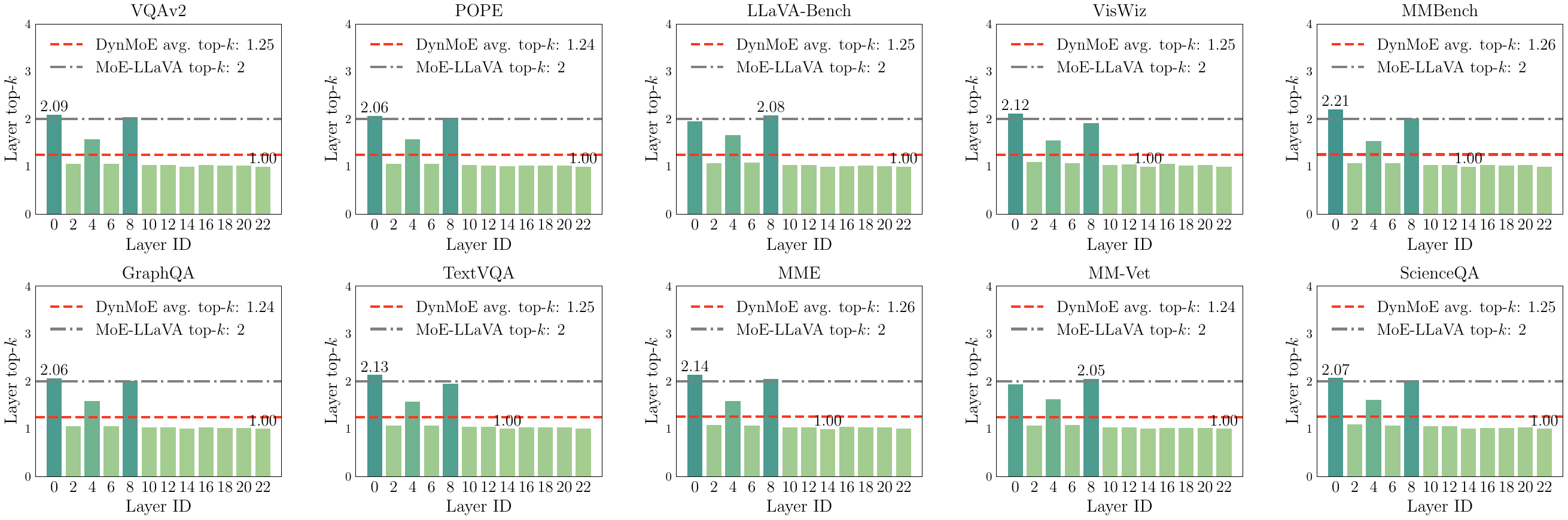}
    \caption{\textbf{Average top-$k$ activated experts of \algabbr on vision-language benchmarks.} We record average top-$k$ activated experts for each MoE layer when using StableLM-1.6B as the language model backbone.}
    \label{fig:avg-topk-stablelm}
\end{figure}

\begin{table}[!t]
    \centering
    \caption{\textbf{Ablation studies on the value of top-$k$ during test.}
        We train the models using \algabbr and set different values of top-$k$ during the test.
        Training and evaluation settings are identical to that of Table~\ref{tab:vision-language}.}
    \vspace{-.5em}
    \resizebox{1.\textwidth}{!}{
        \begin{tabular}{l|c|ccccc|ccccc}
            \toprule
            Algorithms             & $N_{A}$ & VQA$^{v2}$ & GQA  & VisWiz & SQA$^I$ & VQA$^T$ & POPE & MME    & MMB  & LLaVA$^W$ & MM-Vet \\
            \midrule
            \textit{StableLM-1.6B}                                                                                                        \\
            \algabbr-LLaVA         & 1.75B   & 77.4       & 61.4 & 40.6   & 63.4    & 48.9    & 85.7 & 1300.9 & 63.2 & 86.4      & 28.1   \\
            \algabbr-LLaVA ($k=2$) & 2.06B   & 76.9       & 61.0 & 39.1   & 62.1    & 49.2    & 85.7 & 1320.4 & 62.4 & 73.6      & 28.2   \\
            \algabbr-LLaVA ($k=3$) & 2.47B   & 76.8       & 60.7 & 37.0   & 62.6    & 48.9    & 85.5 & 1306.9 & 62.5 & 74.0      & 26.8   \\
            \algabbr-LLaVA ($k=4$) & 2.89B   & 76.8       & 60.5 & 34.8   & 61.9    & 49.0    & 85.8 & 1321.9 & 61.9 & 75.8      & 27.8   \\
            \midrule     \textit{Qwen-1.8B}                                                                                               \\
            \algabbr-LLaVA         & 2.19B   & 76.2       & 61.5 & 32.6   & 63.1    & 48.0    & 87.0 & 1291.6 & 59.7 & 88.7      & 25.3   \\
            \algabbr-LLaVA ($k=2$) & 2.24B   & 76.2       & 60.8 & 33.8   & 62.2    & 47.7    & 87.5 & 1281.3 & 60.4 & 91.3      & 23.0   \\
            \algabbr-LLaVA ($k=3$) & 2.65B   & 76.2       & 60.5 & 32.2   & 62.9    & 48.1    & 88.4 & 1263.7 & 60.7 & 87.8      & 23.4   \\
            \algabbr-LLaVA ($k=4$) & 3.05B   & 75.7       & 60.0 & 31.6   & 62.8    & 48.3    & 88.1 & 1263.4 & 61.0 & 86.7      & 23.7   \\
            \midrule
            \textit{Phi-2-2.7B}                                                                                                           \\
            \algabbr-LLaVA         & 3.35B   & 77.9       & 61.6 & 45.1   & 68.0    & 51.8    & 86.0 & 1429.6 & 66.6 & 95.6      & 33.6   \\
            \algabbr-LLaVA ($k=2$) & 3.62B   & 77.8       & 61.5 & 41.6   & 67.6    & 51.8    & 85.5 & 1433.5 & 66.8 & 95.1      & 32.7   \\
            \algabbr-LLaVA ($k=3$) & 4.46B   & 77.7       & 61.8 & 42.0   & 68.0    & 52.3    & 86.3 & 1438.1 & 66.8 & 94.3      & 30.8   \\
            \algabbr-LLaVA ($k=4$) & 5.30B   & 77.5       & 61.4 & 41.7   & 68.0    & 52.4    & 87.0 & 1431.5 & 66.5 & 95.8      & 32.8   \\
            \bottomrule
        \end{tabular}
    }
    \label{tab:ablation-test-top-k}
    \vspace{-1em}
\end{table}

\begin{figure}[!t]
    \centering
    \subfigure[Activation frequency \footnotesize{(Qwen)}]{
        \includegraphics[width=.315\textwidth]{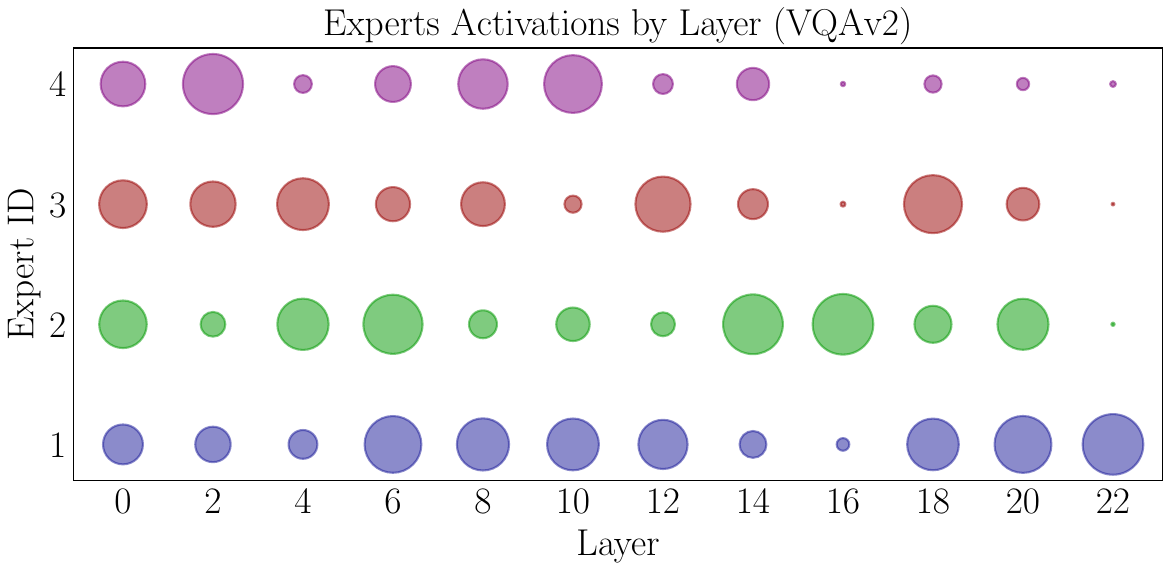}
    }
    \subfigure[Activation frequency \footnotesize{(StableLM)}]{
        \includegraphics[width=.315\textwidth]{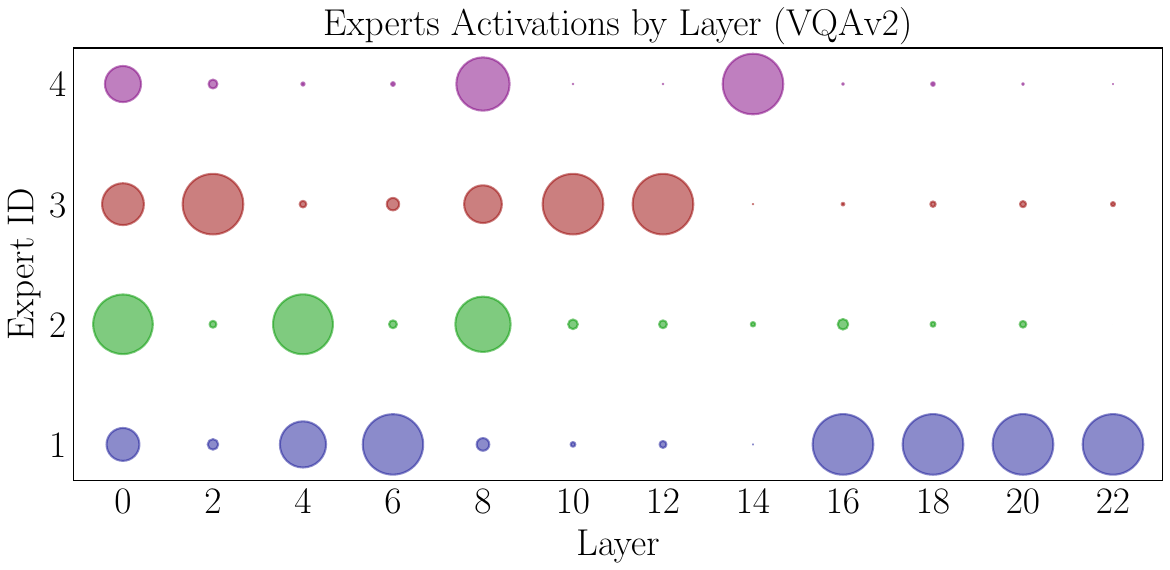}
    }
    \subfigure[Activation frequency \footnotesize{(Phi-2)}]{
        \includegraphics[width=.315\textwidth]{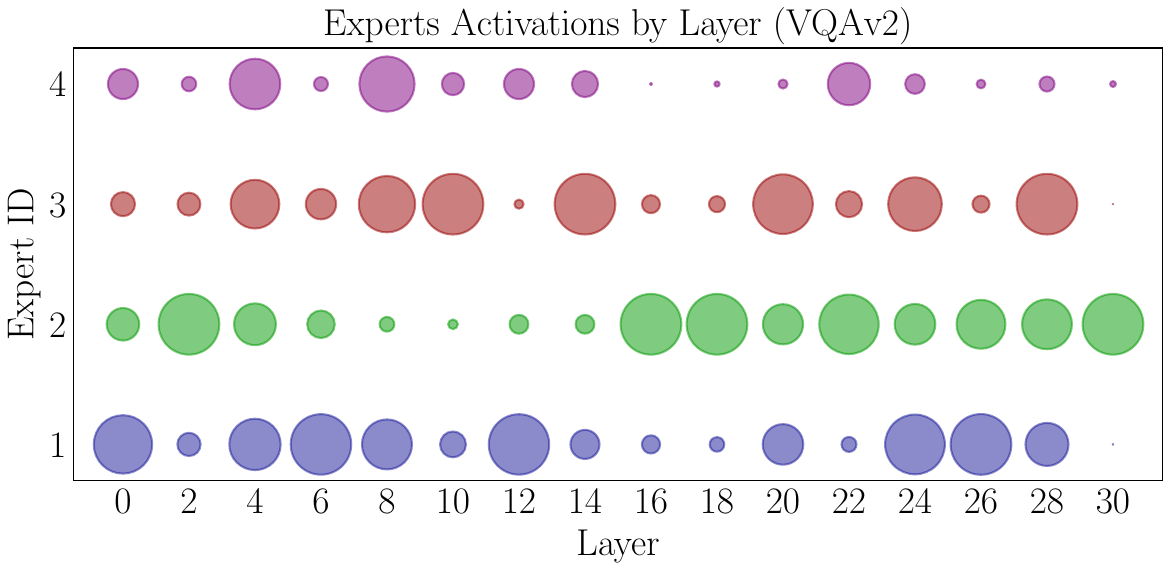}
    }
    \vspace{-0.5em}
    \caption{
        \textbf{Statistics of expert activation frequency in different layers.}
        We report the frequency of expert activations in various layers for the VQA task.
        Larger circles indicate experts that are activated more frequently.
    }
    \label{fig:expert-activation-freq}
    \vspace{-1.5em}
\end{figure}

\begin{table}[!t]
    \centering
    \caption{
        \textbf{Efficiency evaluation of \algabbr comparing to MoE-LLaVA.} We conduct experiments on single A100 GPU (80 GB) paired with 16 CPUs using identical environment and identical inference configurations.
        We report the performance of MoE-LLaVA using DeepSpeed's top-2 gating implementation.
        The symbols $\downarrow$ and $\uparrow$ indicate that lower and higher values, respectively, denote better performance. The results in this table are averaged over 5 trials, with the first sample excluded to avoid measuring unnecessary memory allocations. Other metrics are reported in Table~\ref{tab:inference-efficiency-appendix}.
    }
    \setlength{\tabcolsep}{2pt}
    {\fontsize{10}{8}\selectfont
        \resizebox{1.\textwidth}{!}{
            \begin{tabular}{l c c c c c c c}
                \toprule
                Model                              & Memory $\downarrow$ & Throughput $\uparrow$ & First Token Latency $\downarrow$ & Wall-clock Time $\downarrow$ & Routing Time $\downarrow$    & Gating Time $\downarrow$     & Expert Passing Time $\downarrow$ \\[1.5pt]
                                                   & (GB)                & (token / second)      & (ms)                             & (second / sample)            & (ms / sample $\times$ layer) & (ms / sample $\times$ layer) & (ms / sample $\times$ layer)     \\
                \midrule
                Dense-LLaVA (StableLM-1.6B)        & 3.68                & 32                    & 72                               & 4.7                          & -                            & -                            & -                                \\
                MoE-LLaVA (StableLM-1.6B$\times$4) & 5.98                & 27                    & 137                              & 6.2                          & 0.04                         & 1.23                         & 1.30                             \\
                \algabbr (StableLM-1.6B$\times$4)  & 5.98                & 30                    & 124                              & 5.7                          & 0.52                         & 0.81                         & 1.17                             \\
                \bottomrule
            \end{tabular}
        }}
    \label{tab:inference-efficiency}
\end{table}

\paragraph{Inference efficiency of \algabbr.} To further evaluate the inference efficiency of DynMoE, we have compared its FLOPs, MACs, speed, and memory usage to those of MoE-LLaVA. The results in Table~\ref{tab:inference-efficiency} show that:
(1) \algabbr chieves higher throughput, lower latency, and reduced wall-clock time compared to MoE-LLaVA, indicating improved efficiency.
(2) The top-any gating introduces additional cost in the router, but the gating and expert passing steps are more efficient than in MoE-LLaVA.
(3) In the current implementation, all experts, loaded or unloaded, occupy GPU memory, resulting in the same memory usage as MoE-LLaVA. Offloading unloaded experts from GPU memory could improve efficiency.

\begin{figure}
    \centering
    \includegraphics[width=0.45\linewidth]{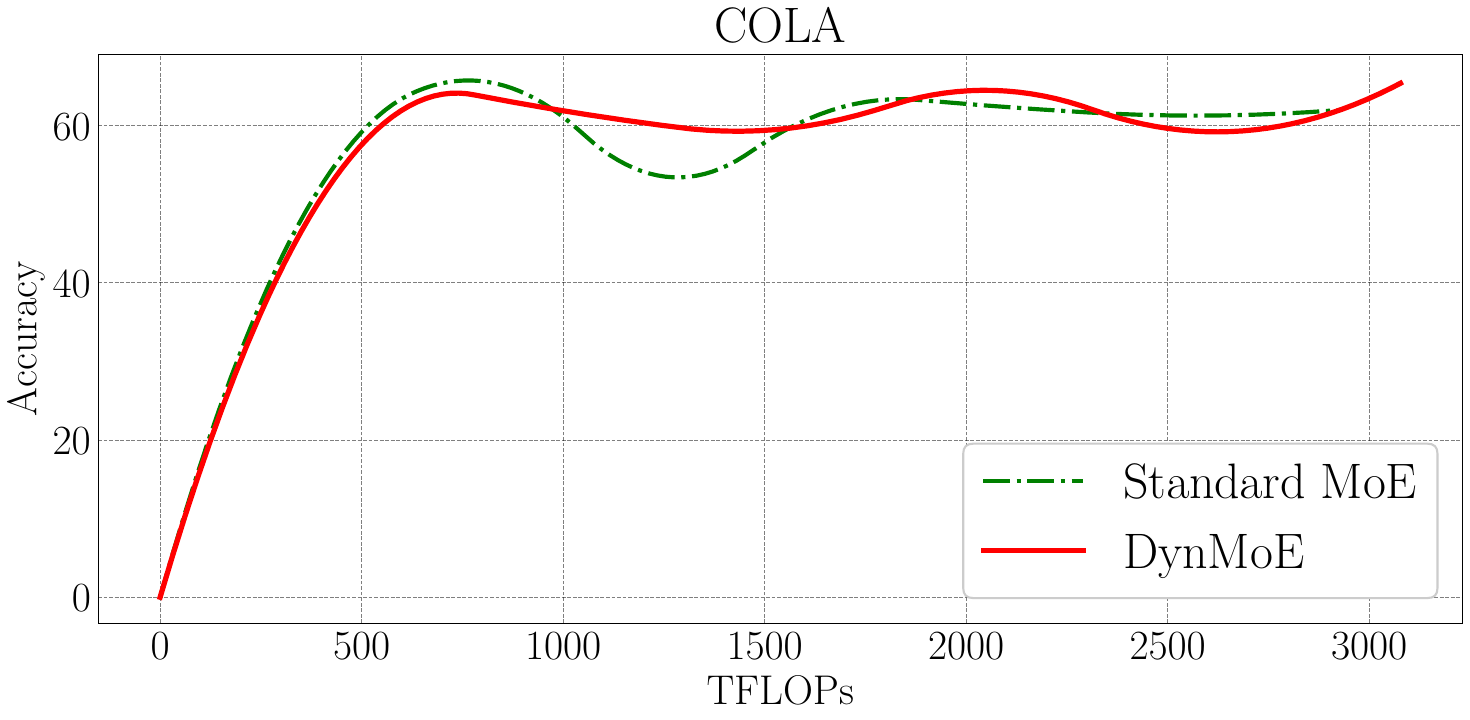}
    \includegraphics[width=0.45\linewidth]{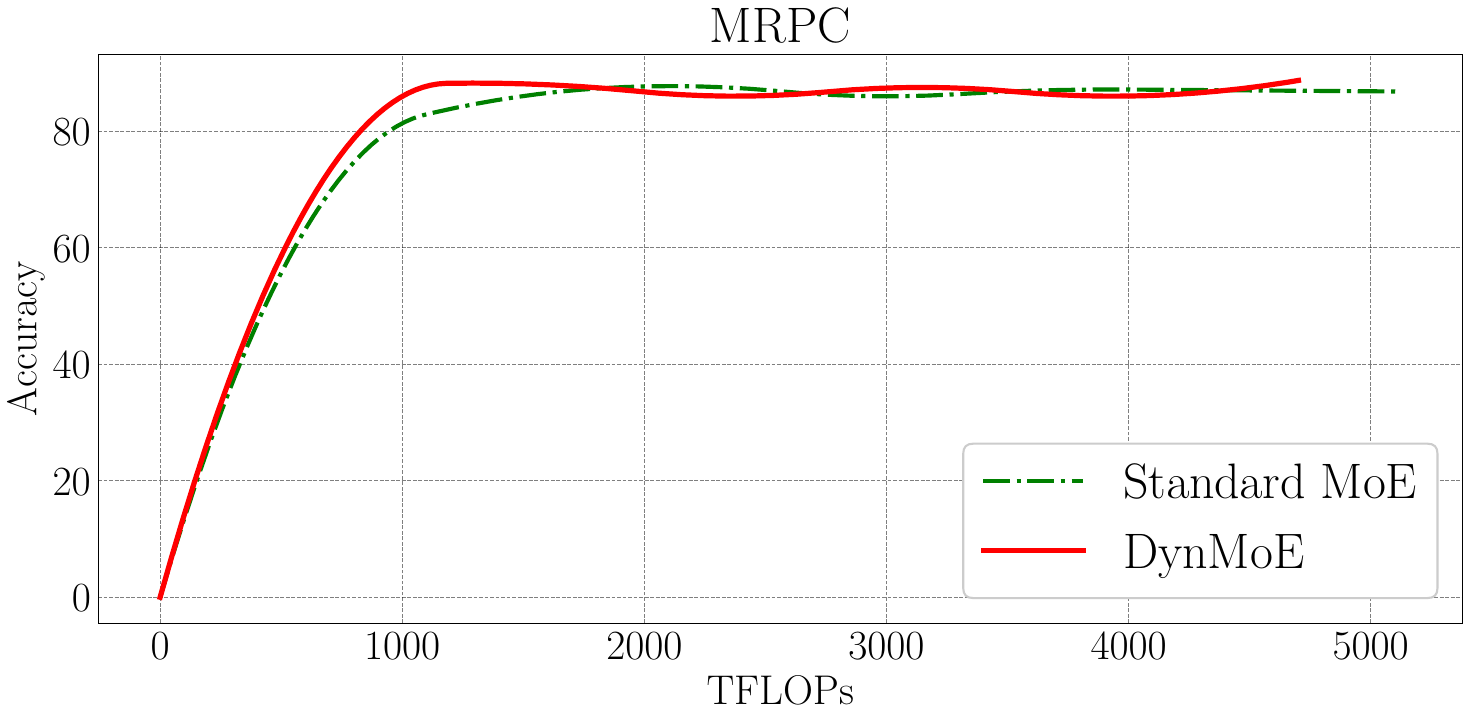}
    \vspace{-1em}
    \caption{\textbf{Convergence curve w.r.t. training FLOPs.}  We present the convergence curve with respect to training FLOPs for \algabbr and the best-performance MoE setting on the GLUE benchmark. }
    \vspace{-1.5em}
    \label{fig:performance-flops-illustration}
\end{figure}

\paragraph{Training efficiency of \algabbr.} We present training FLOPs for both
Language (Figure~\ref{fig:performance-flops-illustration}) and Vision-Language (Table~\ref{tab:inference-efficiency-appendix}) experiments.
The results show that \algabbr achieves comparable or lower FLOPs than standard MoE, ensuring both efficiency and performance without extensive parameter tuning.

\subsection{\label{answer4}A4: \algabbr Provide Insights on MoE Architecture Design}

\paragraph{MoE structure is required for bottom layer rather than top layer.}
In Figures~\ref{fig:avg-topk-stablelm} and~\ref{fig:expert-activation-freq}, we present the average top-$k$ of \algabbr-LLaVA and the frequency of expert activation across various layers.
Our observations indicate that:
(1) In the top layer (the layer closest to the LM prediction head), tokens tend to select the same expert, while in the bottom layer, tokens activate all experts uniformly.
This suggests that there is no need to convert the top layer to MoE layer, whereas the bottom layer should be transformed into MoE layer.
(2) Different LLM backbones may exhibit distinct expert activation frequency patterns.
For the StableLM backbone, most MoE layers activate only one expert, whereas for the Phi-2 backbone, experts are more likely to be activated uniformly.

\paragraph{Shared experts exist in each MoE layer.} Figures~\ref{fig:gates-stablelm}-~\ref{fig:gates-phi} display the threshold $\mG$ values for each MoE layer. We notice that typically, one expert per layer has a significantly lower threshold, making it more easier to be activated. This observation is consistent with Deepseek-MoE's~\citep{dai2024deepseekmoe} design of incorporating shared experts for all tokens in each MoE layer.\looseness=-1

\section{Conclusion and Future Works}
\label{sec:conclusion}
In this paper, we introduce \algabbr, a method that automatically determines both the number of experts and the number of experts to activate. Our results show that \algabbr delivers comparable or even superior performance across various MoE model configurations, while maintaining efficiency. This demonstrates \algabbr's potential to save researchers time and computational resources in hyperparameter tuning. Additionally, our visualizations reveal interesting insights, such as the reduced number of experts needed for the top layers, which could inspire future advancements in MoE model design.
For future work, as discussed in \citet{han2021dynamic}, MoE can be considered a dynamic model because different tokens may activate different experts, thereby enabling adaptive computation and enhancing the model's ability to adapt to input data. While \algabbr addresses dynamic challenges through adaptive top-$k$ selection and an adaptive number of experts, exploring integration with other dynamic techniques, such as layer skipping \citep{zhao2024dynamic}, would also be valuable.
Moreover, the current adaptive process and top-any gating method are not sufficiently efficient. Developing more optimized implementations, such as designing CUDA kernels, would be valuable in the future.

\clearpage
\section*{Acknowledgments}
This work is supported in part by the funding from Shenzhen Institute of Artificial Intelligence and Robotics for Society, in part by the Shenzhen Key Lab of Crowd Intelligence Empowered Low-Carbon Energy Network (Grant No. ZDSYS20220606100601002), in part by Shenzhen Stability Science Program 2023, and in part by the Guangdong Provincial Key Laboratory of Future Networks of Intelligence (Grant No. 2022B1212010001).
This work is also supported in part by the Research Center for Industries of the Future (RCIF) at Westlake University, Westlake Education Foundation, and Westlake University Center for High-performance Computing.

\bibliography{paper}
\bibliographystyle{configuration/iclr2025_conference}

\clearpage
\onecolumn
{
    \hypersetup{linkcolor=black}
    \parskip=0em
    \renewcommand{\contentsname}{Contents of Appendix}
    \tableofcontents
    \addtocontents{toc}{\protect\setcounter{tocdepth}{3}}
}

\section{Experiment Settings}
\label{sec:experiment-setting-appendix}

We conduct experiments on Vision, Language, and Vision-Language tasks. The detailed experiment settings are shown in the following.
\begin{itemize}[leftmargin=12pt, nosep]
    \item \textbf{Vision Task.} For the vision tasks, we follow the same settings as in GMoE~\citep{li2023sparse}. We employ the pre-trained ViT-S/16~\cite{dosovitskiy2020image} model and evaluate it on the DomainBed~\citep{gulrajani2020search} benchmark.
          Our experiments encompass four Domain Generalization datasets: PACS~\citep{li2017deeper}, VLCS~\citep{albuquerque2019generalizing}, OfficeHome~\citep{venkateswara2017deep}, and DomainNet~\citep{peng2019moment}.
          All results are reported using the train-validation selection criterion.
          We conduct all experiments on a single RTX 3090 GPU, and the reported results are averaged over three random seeds. For \algabbr, we set the maximum number of experts to 8 and the initial number of experts to 6. The adaptive process is executed for each iteration.
    \item \textbf{Language Task.} The language tasks adhere to the same settings as those in MoEfication~\citep{zhang2022moefication} and EMoE~\citep{qiu2023emergent}.
          The MoE models are built upon the BERT-large~\citep{devlin2019bert} architecture using the MoEfication method and are fine-tuned on GLUE~\citep{wang2018glue} tasks, which include COLA~\citep{warstadt2019neural}, QNLI~\citep{wang2018glue}, RTE~\citep{bentivogli2009fifth}, MNLI~\citep{xu2020clue}, and MRPC~\citep{dolan2005automatically}. We conduct all experiments on a single RTX 3090 GPU, and the reported results are averaged over three random seeds. For \algabbr, we set the maximum number of experts to 8 and the initial number of experts to 6. For each epoch, we begin recording routing at 1/3 of the epoch and complete recording routing and execute the adaptive process at 2/3 of the epoch.
    \item \textbf{Vision-Language Task.} The vision-language tasks follows the setting in MoE-LLaVA~\citep{lin2024moe}, where we use StableLM-2-1.6B~\citep{bellagente2024stable}, Qwen-1.8B~\citep{bai2023qwen} and Phi-2~\citep{hughesPhi2SurprisingPower2023} as backbone language models, and use clip-vit-large-patch14-336~\citep{radford2021learning} as the vision encoder. We conduct model training on 8 A100 (80G) GPUs, completing within 2 days, detailed hyper-parameters setting are shown in Table~\ref{tab:training-hparams}. The models are evaluated on image understanding benchmarks including VQA-v2~\citep{goyal2017making}, GQA~\citep{hudson2019gqa}, VisWiz~\citep{gurari2018vizwiz}, ScienceQA-IMG~\citep{lu2022learn}, TextVQA~\citep{singh2019towards}, POPE~\citep{li2023evaluating}, MME~\citep{yin2023survey}, MMBench~\citep{liu2023mmbench}, LLaVA-Bench (in-the-Wild)~\citep{liu2024visual}, and MM-Vet~\citep{yu2023mm}. Furthermore, we keep routing records in our model during testing time. For each benchmark, we collect the number of experts' activations per MoE layer and total processed tokens during testing.
\end{itemize}

\begin{table}[]
    \centering
    \caption{\textbf{Detailed training hyper-parameters and configuration.}}
    \label{tab:training-hparams}
    \begin{tabular}{@{}lccc@{}}
        \toprule
        \multirow{2}{*}{Config} & \multicolumn{3}{c}{Models}                                                                  \\ \cmidrule(l){2-4}
                                & StableLM                                    & Qwen                  & Phi-2                 \\ \midrule
        Maximum experts         & \multicolumn{3}{c}{4}                                                                       \\ \midrule
        Deepspeed               & Zero2                                       & Zero2                 & Zero2\_offload        \\
        Data                    & \multicolumn{3}{c}{LLaVA-Finetuning}                                                        \\
        Image resolution        & \multicolumn{3}{c}{336 $\times$ 336}                                                        \\
        Image encoder           & \multicolumn{3}{c}{CLIP-Large/336}                                                          \\
        Feature select layer    & \multicolumn{3}{c}{-2}                                                                      \\
        Image projector         & \multicolumn{3}{c}{Linear layers with GeLU}                                                 \\
        Epoch                   & \multicolumn{3}{c}{1}                                                                       \\
        Learning rate           & \multicolumn{3}{c}{2e-5}                                                                    \\
        Learning rate schedule  & \multicolumn{3}{c}{Cosine}                                                                  \\
        Weight decay            & \multicolumn{3}{c}{0.0}                                                                     \\
        Batch size per GPU      & 8                                           & 8                     & 4                     \\
        GPU                     & 4 $\times$ A100 (80G)                       & 8 $\times$ A100 (80G) & 8 $\times$ A100 (80G) \\
        Precision               & \multicolumn{3}{c}{Bf16}                                                                    \\ \bottomrule
    \end{tabular}
\end{table}

\section{Detailed Algorithm Framework}

\begin{algorithm}[!t]
    \small
    \begin{algorithmic}[1]
        \small
        \Require{Input data $\xx$, initial gating network parameters $\mW_g$, $\mG$, and $\tau$, experts $E_1, \cdots, E_K$, start record routing flag $flag_{s}$, finish record routing flag $flag_{f}$.}
        \Ensure{MoE layer output $\yy$, auxiliary loss value.}
        \If{$flag_{s}$}
        \myState{Set routing flag $flag_{rout} = 1$.}
        \myState{Initialize routing records by $\mR_{\text{rout}} = \0_{K}$.}
        \myState{Initialize non-activate sample records $\mR_{\text{sam}} = \0_{d}$.}
        \EndIf
        \myState{Get the gating outputs $g(\xx)$ and $\kk$ by Eq~\eqref{equ:gating-score}} and~\eqref{equ:gating-k}.
        \myState{Get MoE layer output $\yy$ by Eq~\eqref{equ:gating-outputs}.}
        \myState{Calculate auxiliary loss by Eq~\eqref{equ:gating-loss}.}
        \If{$flag_{rout} = 1$}
        \myState{$\mR_{E} = \mR_{E} + \text{sum}(g(\xx), \text{dim}=0)$.}
        \myState{$\mR_{S} = \mR_{S} + \sum_{i=1}^{N} \1_{\kk_i = 0} \xx_i$}
        \EndIf
        \If{$flag_{f}$}
        \myState{$flag_{rout} = 0$.}
        \If{Exists $e$ that $\mR_{\text{E}}^{e} = \mathbf{0}$}
        \myState{Remove experts $e$.}
        \EndIf
        \If{$\mR_{\text{S}, e} \not = \mathbf{0}$}
        \myState{Add new expert $K + 1$ with expert representation $\mW_{g, K + 1} = \mR_{S} / \norm{\mR_{S}}$.}
        \EndIf
        \EndIf
    \end{algorithmic}
    \mycaptionof{algorithm}{\small Pseudo code of \algabbr on each iteration and MoE layer.}
    \label{alg:algorithm-framework-general}
\end{algorithm}

\section{Additional Experiments}
\label{sec:additional-experiments}

\subsection{Detailed Results on Language and Vision Tasks}

In this section, we present the detailed results of our experiments on the GLUE benchmark~\citep{wang2018glue} in Table~\ref{tab:language} and on the DomainNet dataset in Table~\ref{tab:domainnet}. These results demonstrate that incorporating the specially designed diversity and simplicity loss significantly enhances the model's performance.

Moreover, we present the detailed results using different learning rates on the GLUE benchmark in Tables~\ref{tab:detailed-results-cola}-~\ref{tab:detailed-results-rte}.

\begin{table}[!t]
    \centering
    \caption{
        \textbf{Detailed performance of \algabbr and various MoE settings on COLA dataset}
    }
    \setlength{\tabcolsep}{2pt}
    {\fontsize{10}{8}\selectfont
    \resizebox{1.\textwidth}{!}{
        \begin{tabular}{l c c c c c c c c c c}
            \toprule
            COLA                                           & $K=8, k=1$ & $K=8, k=2$ & $K=8, k=4$ & $K=8, k=8$ & $K=16, k=1$ & $K=16, k=2$ & $K=16, k=4$ & $K=16, k=8$ & DynMoE \\
            \midrule
            lr = 2e-5 & 64.10 & 64.51 & 64.94 & 43.00 & 63.63 & 64.71 & 64.12 & 64.37 & 65.17  \\
            lr = 3e-5 & 63.86 & 62.10 & 64.73 & 64.03 & 61.76 & 22.04 & 63.42 & 63.13 & 62.80  \\
            lr = 5e-5 & 41.83 & 39.68 & 62.63 & 0.00 (fail) & 37.26 & 38.30 & 20.24 & 25.79 & 40.68  \\
            \bottomrule
        \end{tabular}
    }}
    \vspace{-1em}
    \label{tab:detailed-results-cola}
\end{table}

\begin{table}[!t]
    \centering
    \caption{
        \textbf{Detailed performance of \algabbr and various MoE settings on MRPC dataset}
    }
    \setlength{\tabcolsep}{2pt}
    {\fontsize{10}{8}\selectfont
    \resizebox{1.\textwidth}{!}{
        \begin{tabular}{l c c c c c c c c c c}
            \toprule
            MPRC                                       & $K=8, k=1$ & $K=8, k=2$ & $K=8, k=4$ & $K=8, k=8$ & $K=16, k=1$ & $K=16, k=2$ & $K=16, k=4$ & $K=16, k=8$ & DynMoE \\
            \midrule
            lr = 2e-5 & 89.74 & 89.63 & 89.74 & 89.36 & 88.07 & 89.02 & 89.74 & 89.56 & 89.57  \\
            lr = 3e-5 & 90.14 & 90.19 & 89.50 & 88.67 & 89.81 & 90.18 & 89.38 & 90.35 & 90.64  \\
            lr = 5e-5 & 88.70 & 84.62 & 88.72 & 84.48 & 88.30 & 89.08 & 87.40 & 79.95 & 90.09  \\
            \bottomrule
        \end{tabular}
    }}
    \vspace{-1em}
    \label{tab:detailed-results-mrpc}
\end{table}

\begin{table}[!t]
    \centering
    \caption{
        \textbf{Detailed performance of \algabbr and various MoE settings on QNLI dataset}
    }
    \setlength{\tabcolsep}{2pt}
    {\fontsize{10}{8}\selectfont
    \resizebox{1.\textwidth}{!}{
        \begin{tabular}{l c c c c c c c c c c}
            \toprule
            QNLI                                           & $K=8, k=1$ & $K=8, k=2$ & $K=8, k=4$ & $K=8, k=8$ & $K=16, k=1$ & $K=16, k=2$ & $K=16, k=4$ & $K=16, k=8$ & DynMoE \\
            \midrule
            lr = 2e-5 & 92.48 & 84.94 & 92.52 & 92.46 & 92.39 & 92.51 & 92.65 & 92.49 & 92.39  \\
            lr = 3e-5 & 92.45 & 92.39 & 92.01 & 78.39 & 78.22 & 92.53 & 92.50 & 92.31 & 92.59  \\
            lr = 5e-5 & 50.54 & 64.46 & 78.13 & 64.43 & 50.54 & 50.54 & 64.27 & 64.43 & 75.50  \\
            \bottomrule
        \end{tabular}
    }}
    \vspace{-1em}
    \label{tab:detailed-results-qnli}
\end{table}

\begin{table}[!t]
    \centering
    \caption{
        \textbf{Detailed performance of \algabbr and various MoE settings on MNLI dataset}
    }
    \setlength{\tabcolsep}{2pt}
    {\fontsize{10}{8}\selectfont
    \resizebox{1.\textwidth}{!}{
        \begin{tabular}{l c c c c c c c c c c}
            \toprule
            MNLI                                           & $K=8, k=1$ & $K=8, k=2$ & $K=8, k=4$ & $K=8, k=8$ & $K=16, k=1$ & $K=16, k=2$ & $K=16, k=4$ & $K=16, k=8$ & DynMoE \\
            \midrule
            lr = 2e-5 & 86.56 & 86.70 & 86.57 & 86.61 & 86.63 & 86.73 & 86.55 & 86.51 & 86.37  \\
            lr = 3e-5 & 86.46 & 52.40 & 69.40 & 69.35 & 69.57 & 68.47 & 86.59 & 69.47 & 52.34  \\
            lr = 5e-5 & 51.44 & 35.45 & 35.45 & 35.45 & 35.45 & 34.54 & 35.45 & 34.24 & 51.68  \\
            \bottomrule
        \end{tabular}
    }}
    \vspace{-1em}
    \label{tab:detailed-results-mnli}
\end{table}

\begin{table}[!t]
    \centering
    \caption{
        \textbf{Detailed performance of \algabbr and various MoE settings on RTE dataset}
    }
    \setlength{\tabcolsep}{2pt}
    {\fontsize{10}{8}\selectfont
    \resizebox{1.\textwidth}{!}{
        \begin{tabular}{l c c c c c c c c c c}
            \toprule
            RTE                                           & $K=8, k=1$ & $K=8, k=2$ & $K=8, k=4$ & $K=8, k=8$ & $K=16, k=1$ & $K=16, k=2$ & $K=16, k=4$ & $K=16, k=8$ & DynMoE \\
            \midrule
            lr = 2e-5 & 73.04 & 70.52 & 74.13 & 74.37 & 74.01 & 66.19 & 75.33 & 72.56 & 72.80  \\
            lr = 3e-5 & 72.44 & 74.85 & 75.09 & 73.53 & 73.16 & 72.32 & 75.21 & 73.53 & 73.41  \\
            lr = 5e-5 & 58.48 & 54.39 & 62.45 & 65.10 & 63.78 & 63.06 & 58.84 & 63.66 & 65.22  \\
            \bottomrule
        \end{tabular}
    }}
    \label{tab:detailed-results-rte}
\end{table}

\begin{table}[!h]
    \centering
    \caption{\textbf{Performance of \algabbr on language tasks}: Our study investigates the performance of \algabbr on language tasks using the GLUE~\citep{wang2018glue} benchmark, with BERT-large serving as the backbone model. The baselines including traditional MoE methods with different number of experts $K$ and top-$k$. In our implementation of \algabbr, we configure the maximum number of experts to 16, with an initial setting of 8 experts. The number of experts is dynamically adjusted in each epoch for \algabbr.  The $-$ represents experiment failure, final results could not be obtained using Gshard loss.}
    \resizebox{1.\textwidth}{!}{
    \begin{tabular}{l c c c c c c}
        \toprule
        Algorithms            & COLA  & MRPC  & QNLI  & MNLI  & RTE   & Average \\
        \midrule
        MoE ($K = 8, k = 1$)  & 64.10$\pm 0.94$ & 90.14$\pm 0.60$ & 92.48$\pm 0.21$ & 86.56$\pm 0.06$ & 73.04$\pm 2.13$ & 81.26   \\
        MoE ($K = 8, k = 2$)  & 64.51$\pm 0.81$ & 90.19$\pm 0.17$ & 92.39$\pm 0.08$ & 86.70$\pm 0.23$ & 74.85$\pm 1.96$ & 81.73   \\
        MoE ($K = 8, k = 4$)  & 64.94$\pm 0.62$ & 89.74$\pm 0.99$ & 92.52$\pm 0.12$ & 86.57$\pm 0.28$ & 75.09$\pm 1.84$ & 81.77   \\
        MoE ($K = 8, k = 8$)  & 64.03$\pm 0.54$ & 89.36$\pm 0.09$ & 92.46$\pm 0.09$ & 86.61$\pm 0.26$ & 74.37$\pm 0.78$ & 81.37   \\
        MoE ($K = 16, k = 1$) & 63.63$\pm 0.20$ & 89.81$\pm 0.30$ & 92.39$\pm 0.21$ & 86.63$\pm 0.17$ & 74.01$\pm 0.29$ & 81.29   \\
        MoE ($K = 16, k = 2$) & 64.71$\pm 1.21$ & 90.18$\pm 1.33$ & 92.53$\pm 0.07$ & 86.73$\pm 0.43$ & 72.32$\pm 3.54$ & 81.29   \\
        MoE ($K = 16, k = 4$) & 64.12$\pm 1.42$ & 89.74$\pm 0.40$ & 92.65$\pm 0.09$ & 86.59$\pm 0.16$ & 75.33$\pm 0.95$ & 81.69   \\
        MoE ($K = 16, k = 8$) & 64.37$\pm 1.14$ & 90.35$\pm 0.68$ & 92.49$\pm 0.11$ & 86.51$\pm 0.20$ & 73.53$\pm 2.21$ & 81.45   \\
        \midrule
        \algabbr, Gshard Loss & 64.88$\pm 0.86$ & 89.85$\pm 0.22$ & 92.42$\pm 0.07$ & -     & 73.41$\pm 0.68$ & -       \\
        \algabbr              & 65.17$\pm 0.26$ & 90.64$\pm 0.26$ & 92.59$\pm 0.08$ & 86.37$\pm 0.13$ & 73.41$\pm 1.96$ & 81.64   \\
        \bottomrule
    \end{tabular}
    }
    \label{tab:language}
\end{table}

\begin{table}[!h]
    \centering
    \caption{\textbf{Detailed results on DomainNet dataset}: We report the detailed test results on each domain of the DomainNet dataset.}
    \resizebox{1.\textwidth}{!}{
        \begin{tabular}{l c c c c c c c}
            \toprule
            Algorithms                                           & clip & info & paint & quick & real & sketch & Average \\
            \midrule
            GMoE (with \algabbr, Gshard Loss)                    & 66.8 & 23.8 & 54.1  & 15.9  & 68.7 & 54.9   & 47.4    \\
            GMoE (with \algabbr, Diverse and Simple Gating Loss) & 68.0 & 24.4 & 55.4  & 16.6  & 69.5 & 55.1   & 48.2    \\
            \bottomrule
        \end{tabular}
    }
    \label{tab:domainnet}
\end{table}

\subsection{Combine \algabbr with Load Balance and Efficiency Losses}
\label{sec:load-balance-effiency}

In Tables~\ref{tab:dynmoe-load-balance-efficiency-performance},~\ref{tab:dynmoe-load-balance-efficiency-activation-freq},~\ref{tab:dynmoe-load-balance-efficiency-sparse}, and~\ref{tab:dynmoe-load-balance-efficiency-top-k}, we report the following metrics:
\begin{itemize}[leftmargin=12pt,nosep]
    \item Performance (Table~\ref{tab:dynmoe-load-balance-efficiency-performance}): The performance of different settings.
    \item Load Balance (Table~\ref{tab:dynmoe-load-balance-efficiency-activation-freq}): The frequency with which each expert is activated, calculated as (expert activation time / total token count).
    \item Efficiency (Table~\ref{tab:dynmoe-load-balance-efficiency-sparse} and~\ref{tab:dynmoe-load-balance-efficiency-top-k}): The top-k values per layer and the top-k activation frequency, calculated as (number of tokens that activate k experts / total tokens).
\end{itemize}
We can find that
\begin{itemize}[leftmargin=12pt,nosep]
    \item Although it does not explicitly enforce load balancing, the original DynMoE achieves load balancing comparable to that of the standard top-2 MoE (Table~\ref{tab:dynmoe-load-balance-efficiency-activation-freq}).
    \item Adding the load balance loss slightly decreases the performance of DynMoE (Table~\ref{tab:dynmoe-load-balance-efficiency-performance}) while increasing the number of activated experts (Table~\ref{tab:dynmoe-load-balance-efficiency-sparse}). However, it improves load balancing (Table~\ref{tab:dynmoe-load-balance-efficiency-activation-freq}).
    \item Adding an additional efficiency loss on top of the load balance loss improves performance (Table~\ref{tab:dynmoe-load-balance-efficiency-performance}) and helps overcome some extreme cases, such as the reduction of the top-k values in the bottom layer from 2.88 to 2.02 (Table~\ref{tab:dynmoe-load-balance-efficiency-sparse}), and reduce the number of tokens that activate all 4 experts (Table~\ref{tab:dynmoe-load-balance-efficiency-top-k}). Moreover, the efficiency loss further enhances load balancing (Table~\ref{tab:dynmoe-load-balance-efficiency-activation-freq}).
\end{itemize}

\begin{table}[!t]
    \centering
    \caption{
        \textbf{Performance of \algabbr with load balance and efficiency losses.} We conduct experiments using the MoE-LLaVA setup, incorporating (1) a load-balancing loss and (2) an efficiency loss to enforce sparsity, as proposed by \citet{zheng2025learn}.
    }
    \setlength{\tabcolsep}{2pt}
    {\fontsize{10}{8}\selectfont
    \resizebox{1.\textwidth}{!}{  
    \begin{tabular}{l c c c c c c c c c c}  
        \toprule  
        Performance (StableLM) & VQAv2 & GQA & VizWiz & SQA & TextVQA & POPE & MME & MMBench \\
        \midrule  
        \algabbr & 77.4 & 61.4 & 40.6 & 63.4 & 48.9 & 85.7 & 1300.9 & 63.2 \\
        \algabbr + load balance & 77.1 & 61.6 & 37.0 & 61.4 & 50.3 & 85.3 & 1313.5 & 61.7 \\
        \algabbr + load balance + efficiency & 77.1 & 61.8 & 39.4 & 62.9 & 49.7 & 85.4 & 1321.2 & 61.9 \\
        \bottomrule  
    \end{tabular}  
    }}
    \label{tab:dynmoe-load-balance-efficiency-performance}
\end{table}

\begin{table}[!t]
    \centering
    \caption{
        \textbf{Activation frequency per expert of \algabbr with load balance and efficiency losses.} We conduct experiments using the MoE-LLaVA setup, incorporating (1) a load-balancing loss and (2) an efficiency loss to enforce sparsity, as proposed by \citet{zheng2025learn}. We report the activation frequency of each expert at layer 0 on the VQAv2 dataset.
    }
    \setlength{\tabcolsep}{2pt}
    {\fontsize{10}{8}\selectfont
    \resizebox{1.\textwidth}{!}{  
    \begin{tabular}{l c c c c c c c c c c}  
        \toprule  
        Activation Frequency per Expert (VQAv2, layer 0) & Expert 1 & Expert 2 & Expert 3 & Expert 4 \\
        \midrule  
        MoE (top-2) & 0.36 & 1.29 & 0.16 & 0.19 \\
        \algabbr & 0.29 & 0.97 & 0.48 & 0.35 \\
        \algabbr + load balance & 0.81 & 0.50 & 0.90 & 0.68 \\
        \algabbr + load balance + efficiency & 0.45 & 0.52 & 0.42 & 0.63 \\
        \bottomrule  
    \end{tabular}  
    }}
    \label{tab:dynmoe-load-balance-efficiency-activation-freq}
\end{table}

\begin{table}[!t]
    \centering
    \caption{
        \textbf{Sparsity of \algabbr with load balance and efficiency losses.} We conduct experiments using the MoE-LLaVA setup, incorporating (1) a load-balancing loss and (2) an efficiency loss to enforce sparsity, as proposed by \citet{zheng2025learn}. We report the average top-k value of each expert at each layer on the VQAv2 dataset.
    }
    \setlength{\tabcolsep}{2pt}
    {\fontsize{10}{8}\selectfont
    \resizebox{1.\textwidth}{!}{  
    \begin{tabular}{l c c c c c c c c c c c c c c c c c c c}  
        \toprule  
        Top-k per Layer (VQAv2) & Layer 0 & Layer 2 & Layer 4 & Layer 6 & Layer 8 & Layer 10 & Layer 12 & Layer 14 & Layer 16 & Layer 18 & Layer 20 & Layer 22 \\
        \midrule  
        \algabbr & 2.09 & 1.07 & 1.57 & 1.06 & 2.04 & 1.03 & 1.03 & 1.00 & 1.03 & 1.02 & 1.02 & 1.00 \\
        \algabbr + load balance & 2.88 & 1.25 & 1.59 & 1.27 & 1.26 & 1.13 & 1.77 & 1.70 & 1.12 & 1.33 & 1.30 & 1.00 \\
        \algabbr + load balance + efficiency & 2.02 & 1.25 & 1.81 & 1.57 & 1.65 & 1.20 & 1.47 & 2.30 & 1.07 & 1.37 & 1.82 & 1.00 \\
        \bottomrule  
    \end{tabular}  
    }}
    \label{tab:dynmoe-load-balance-efficiency-sparse}
\end{table}

\begin{table}[!t]
    \centering
    \caption{
        \textbf{Top-k frequency of \algabbr with load balance and efficiency losses.} We conduct experiments using the MoE-LLaVA setup, incorporating (1) a load-balancing loss and (2) an efficiency loss to enforce sparsity, as proposed by \citet{zheng2025learn}. We report the frequency of activating top-k experts for each configuration.
    }
    \setlength{\tabcolsep}{2pt}
    {\fontsize{10}{8}\selectfont
    \begin{tabular}{l c c c c c c c c c c c c c c c c c c c}  
        \toprule  
        Top-k Frequency (VQAv2) & Top-1 & Top-2 & Top-3 & Top-4 \\
        \midrule  
        \algabbr & 0.79 & 0.16 & 0.04 & 0.01 \\
        \algabbr + load balance & 0.65 & 0.26 & 0.06 & 0.03 \\
        \algabbr + load balance + efficiency & 0.58 & 0.32 & 0.09 & 0.01\\
        \bottomrule  
    \end{tabular}  
    }
    \label{tab:dynmoe-load-balance-efficiency-top-k}
\end{table}

\subsection{Comparision to Top-p Gating Baseline}

In Table~\ref{tab:dynmoe-top-p}, we compare \algabbr with the method proposed by \citet{huang2024harder}, which replaces the traditional top-k gating with top-p gating. We set $p=0.4$ as suggested in the original paper. Our results show that DynMoE achieves better performance without requiring the additional parameter.

\begin{table}
    \centering
    \caption{
        \textbf{Comparision to top-p gating method.} We conduct experiments using the MoE-LLaVA setup, and compare \algabbr to top-p gating method~\citep{huang2024harder}.
    }
    \begin{tabular}{l c c c c c c c c c c c c c}
        \toprule
        StableLM & VQAv2 & GQA & VizWiz & SQA & TextVQA & POPE & MME & MMBench \\
        \midrule
        \algabbr & 77.4 & 61.4 & 40.6 & 63.4 & 48.9 & 85.7 & 1300.9 & 63.2 \\
        top p (p=0.4) & 77.1 & 61.7 & 36.0 & 62.8 & 48.6 & 85.2 & 1332.9 & 62.3 \\
        \bottomrule
    \end{tabular}
    \label{tab:dynmoe-top-p}
\end{table}

\subsection{Numberical Results on More Vision Tasks}
\label{sec:additiona-vision-task}

In Tables~\ref{tab:tiny-imagenet},~\ref{tab:cifar-10}, and~\ref{tab:imagenet}, we show the performance of \algabbr on more vision tasks, and also investigate the impact of strategies on initialize the new experts, including 
\begin{itemize}
    \item \textbf{Average:} Averaging the parameters of existing experts to initialize the new expert.
    \item \textbf{W-Average:} Using weighted averaging of the parameters of existing experts, where the weights correspond to the number of experts to be activated.
    \item \textbf{Most activated:} Initializing the new expert using the parameters of the most frequently activated expert.
\end{itemize}
Results show that (1) DynMoE converges faster than standard MoE settings; (2) W-Average achieves the best performance in most cases; and (3) incorporating load balance loss and efficiency loss accelerates training while improving performance.

\begin{table}
    \centering
    \caption{
        \textbf{Fintuneing results on Tiny-ImageNet dataset.} We fine-tune the pretrained ViT-S~\citep{dosovitskiy2020image} model on the TinyImageNet~\citep{le2015tiny} dataset and report the accuracy every two epochs.
    }
    \setlength{\tabcolsep}{2pt}
    {\fontsize{10}{8}\selectfont
    \resizebox{1.\textwidth}{!}{  
    \begin{tabular}{l c c c c c c c c c c c c c}
        \toprule
        TinyImageNet (Finetune, ViT-S, 2 MoE layers) & E1 & E3 & E5 & E7 & E9 & E11 & E13 & E15 & E17 & E19 & E20 \\
        \midrule
        MoE (K = 8, k = 1) & 78.32 & 82.79 & 84.03 & 84.83 & 85.20 & 85.61 & 85.82 & 86.27 & 86.44 & 86.61 & 86.65 \\
        MoE (K = 8, k = 2) & 78.53 & 82.95 & 84.05 & 84.74 & 84.99 & 85.00 & 85.95 & 86.45 & 86.63 & 86.58 & 86.72 \\
        MoE (K = 8, k = 4) & 79.25 & 83.38 & 83.73 & 84.72 & 85.00 & 85.50 & 85.93 & 86.27 & 86.00 & 86.64 & 86.56 \\
        MoE (K = 8, k = 8) & 79.20 & 83.30 & 84.02 & 84.10 & 84.86 & 85.62 & 86.08 & 86.12 & 86.44 & 86.73 & 86.58 \\
        DynMoE (Original, avg topk=6.5) & 79.10 & 83.09 & 84.20 & 84.84 & 85.18 & 85.56 & 85.91 & 86.09 & 86.37 & 86.40 & 86.70 \\
        DynMoE (Average, avg topk=6.0) & 79.19 & 83.48 & 84.21 & 84.84 & 85.32 & 85.76 & 86.25 & 86.41 & 86.49 & 86.70 & 86.75 \\
        DynMoE (W-Average, avg topk=6.0) & 78.96 & 83.18 & 84.15 & 84.92 & 85.34 & 85.93 & 86.10 & 86.30 & 86.60 & 86.70 & 86.80 \\
        DynMoE (Most activated, avg topk=6.5) & 79.09 & 83.57 & 84.21 & 84.62 & 85.40 & 85.87 & 86.45 & 86.40 & 86.63 & 86.66 & 86.70 \\
        \bottomrule
    \end{tabular}
    }}
    \label{tab:tiny-imagenet}
\end{table}

\begin{table}
    \centering
    \caption{
        \textbf{Pretrain results in CIFAR-10 dataset.} We pretrain the ViT-S~\citep{dosovitskiy2020image} model on the CIFAR-10 dataset and report the accuracy every 100 epochs.
    }
    \setlength{\tabcolsep}{2pt}
    {\fontsize{10}{8}\selectfont
    \resizebox{.5\textwidth}{!}{  
    \begin{tabular}{l c c c c c c c c c c c c c}
        \toprule
        CIFAR10 (ViT-S, 2 MoE Layer, Acc per 100 Epoch) & 200 & 300 & 400 & 500 \\
        \midrule
        MoE (K = 8, k = 1) & 72.66 & 77.51 & 80.10 & 81.08 \\
        MoE (K = 8, k = 2) & 73.79 & 78.50 & 80.85 & 81.91 \\
        MoE (K = 8, k = 4) & 72.77 & 77.84 & 80.30 & 81.14 \\
        MoE (K = 8, k = 8) & 70.68 & 75.32 & 78.28 & 79.11 \\
        DynMoE (Original, avg topk=7) & 74.84 & 79.24 & 81.77 & 82.50 \\
        DynMoE (Average, avg topk=7) & 74.70 & 80.32 & 82.51 & 83.57 \\
        DynMoE (W-Average, avg topk=6.5) & 74.16 & 78.77 & 81.30 & 82.01 \\
        DynMoE (Most activated, avg topk=6.5) & 71.71 & 77.56 & 80.08 & 80.58 \\
        \bottomrule
    \end{tabular}
    }}
    \label{tab:cifar-10}
\end{table}

\begin{table}
    \centering
    \caption{
        \textbf{Pretrain results on ImageNet dataset.} 
        We train ViT-S from scratch for 200 epochs on ImageNet dataset~\citep{deng2009imagenet} with a batch size of 512 across 8 A100 GPUs. In the ViT-S architecture, the layers [0, 3, 6, 9] are replaced with MoE layers, and the maximum number of experts per layer is 4. The learning rate is set to 1e-4, and we use the Adam optimizer with parameters [0.9, 0.99] and cosine learning schedule, while the weight decay is set to 5e-5. During evaluation, we set the batch size to 128 and use 1 A100 GPU. DynMoE-A indicates DynMoE with a simple and diverse gating loss, and DynMoE-B indicates DynMoE with simple and diverse gating loss + load balance loss + efficiency loss.
    }
    \setlength{\tabcolsep}{2pt}
    {\fontsize{10}{8}\selectfont
    \resizebox{1.\textwidth}{!}{  
    \begin{tabular}{l c c c c c c c c c c c c c}
        \toprule
        & Train Time & Train Wall-Clock Time & Evaluation Time & Inference Time on Routing & Inference Time on Gating & Inference Time on Expert Passing & Acc@1 \\
        & (s/batch) & (days) & (s/batch) & (ms / batch and MoE layer) & (ms / batch and MoE layer) & (ms / batch and MoE layer) & (\%) \\
        \midrule
        DeepSpeed-MoE (top-1) & 0.39 & 2.3 & 0.53 & 0.06 & 14.54 & 94.32 & 64.7 \\
        DeepSpeed-MoE (top-2) & 0.63 & 3.7 & 1.05 & 0.06 & 50.78 & 186.47 & 67.3 \\
        DynMoE-A (k=1.63) & 0.51 & 2.8 & 0.99 & 0.53 & 36.45 & 186.18 & 66.5 \\
        DynMoE-B (k=1.25) & 0.41 & 2.5 & 0.82 & 0.46 & 31.04 & 148.11 & 68.5 \\
        \bottomrule
    \end{tabular}
    }}
    \label{tab:imagenet}
\end{table}

\subsection{Efficiency Evaluation}

In Table~\ref{tab:train-efficiency}, we compare the training efficiency of \algabbr with standard top-1 and top-2 gating MoE models. The results show that \algabbr trains faster than top-2 gating but slower than top-1 gating.

In Table~\ref{tab:overhead}, we show the overhead introduced by the top-any gating method. The results indicate that the primary source of overhead comes from the routing process, which likely explains why DynMoE (with forced top-1) is noticeably slower than MoE (also with forced top-1).

\begin{table}
    \centering
    \caption{
        \textbf{Training efficiency comparision.} 
    }
    \setlength{\tabcolsep}{2pt}
    {\fontsize{10}{8}\selectfont
    \resizebox{1.\textwidth}{!}{  
    \begin{tabular}{l c c c c c c c c c c c c c}
        \toprule
        Time & MoE-LLaVA & ViT-S, ImageNet & ViT-S, ImageNet \\
        (s/batch) & (Train, 4 A100, batch size=32) & (Train, 8 A100, batch size=512) & (Test, 4 A100, batch size=256) \\
        \midrule
        Top-1 MoE & 1.31 & 0.39 & 0.18 \\
        Top-2 MoE & 1.60 & 0.63 & 0.32 \\
        DynMoE & 1.48 & 0.51 & 0.27 \\
        \bottomrule
    \end{tabular}
    }}
    \label{tab:train-efficiency}
\end{table}

\begin{table}
    \centering
    \caption{
        \textbf{Overhead of top-any gating.} 
        We report the time taken for key steps in top-any routing, including routing, gating, and expert passing.
    }
    \setlength{\tabcolsep}{2pt}
    {\fontsize{10}{8}\selectfont
    \begin{tabular}{l c c c c c c c c c c c c c}
        \toprule
        Time (per sample and MoE layer) & Routing (ms) & Gating (ms) & Expert Passing (ms) \\
        \midrule
        MoE-LLaVA (enforced top-1) & 0.04 & 0.60 & 1.08 \\
        DynMoE (enforced top-1) & 0.45 & 0.75 & 1.10 \\
        \bottomrule
    \end{tabular}
    }
    \label{tab:overhead}
\end{table}

\subsection{Ablation Studies on Aggregation Weights}

In Remark~\ref{remark:not-use-weights}, we discussed why we did not consider the magnitude of scores when averaging the expert outputs. In Table~\ref{tab:weights-ablations}, we show that using different expert scores significantly reduce the model performance.

\begin{table}
    \centering
    \caption{
        \textbf{Ablation studies on aggregation weights.} 
        We conduct experiments using the MoE-LLaVA setup, and the "weighted scores" indicates averaging the expert outputs based on the gating outputs.
    }
    \setlength{\tabcolsep}{2pt}
    {\fontsize{10}{8}\selectfont
    \begin{tabular}{l c c c c c c c c c c c c c}
        \toprule
        StableLM & VQAv2 & GQA & VizWiz & SQA & TextVQA & POPE & MME & MMBench \\
        \midrule
        ours & 77.4 & 61.4 & 40.6 & 63.4 & 48.9 & 85.7 & 1300.9 & 63.2 \\
        weighted scores & 73.9 & 57.4 & 32.1 & 61.3 & 46.9 & 84.2 & 1176.8 & 52.1 \\
        \bottomrule
    \end{tabular}
    }
    \label{tab:weights-ablations}
\end{table}

\section{Additional Visualization Results}

\subsection{Activation Frequency}

\begin{figure}
    \centering
    \subfigure[GraphQA]{
        \includegraphics[width=.45\textwidth]{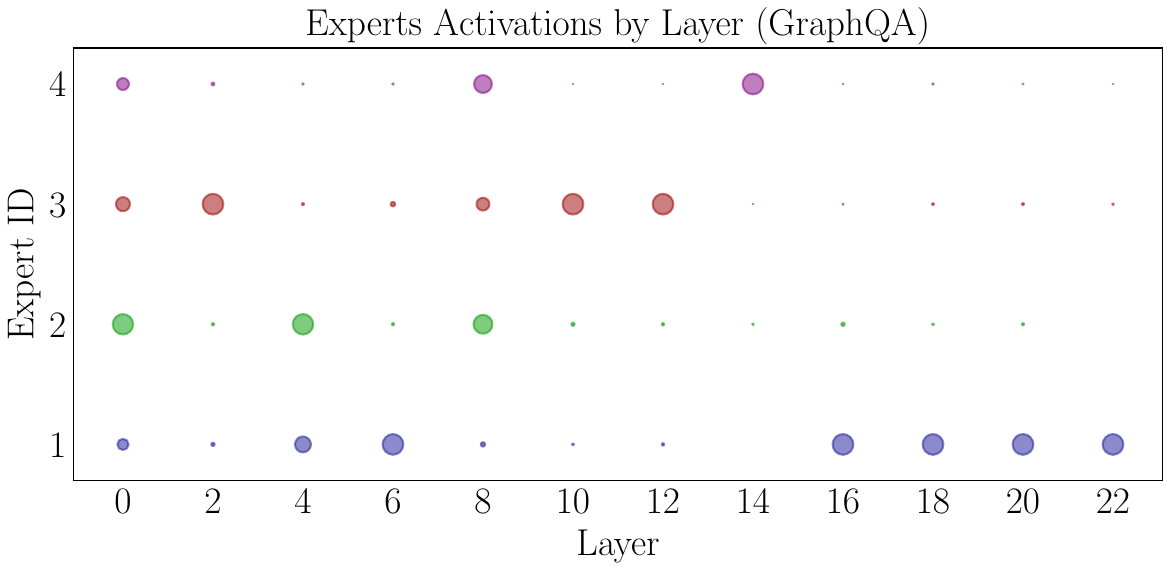}
    }
    \subfigure[LLaVA-Bench]{
        \includegraphics[width=.45\textwidth]{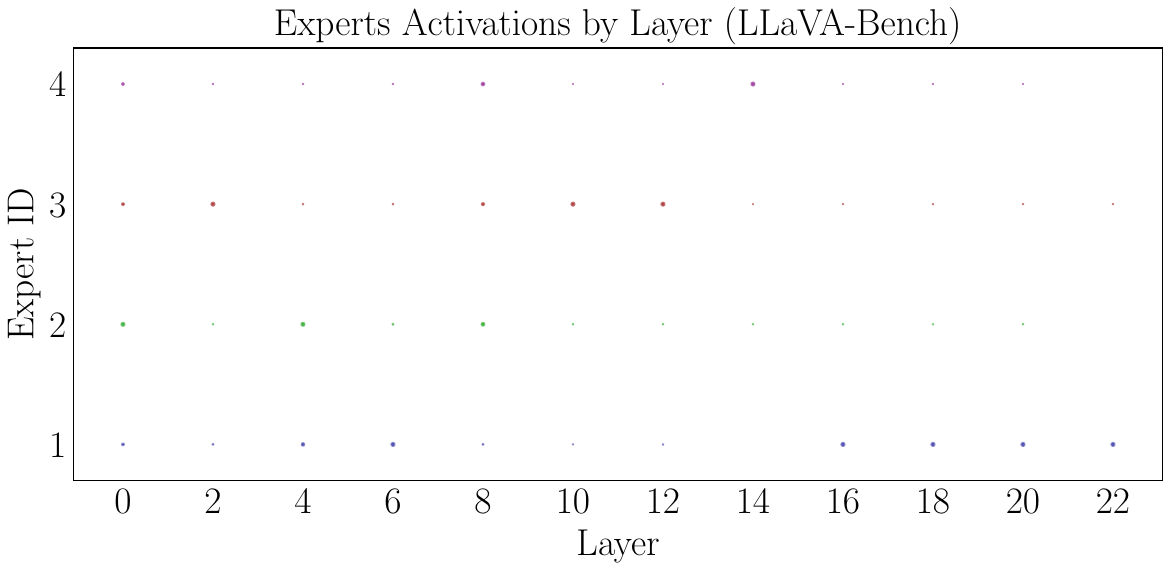}
    }
    \\
    \subfigure[MM-Vet]{
        \includegraphics[width=.45\textwidth]{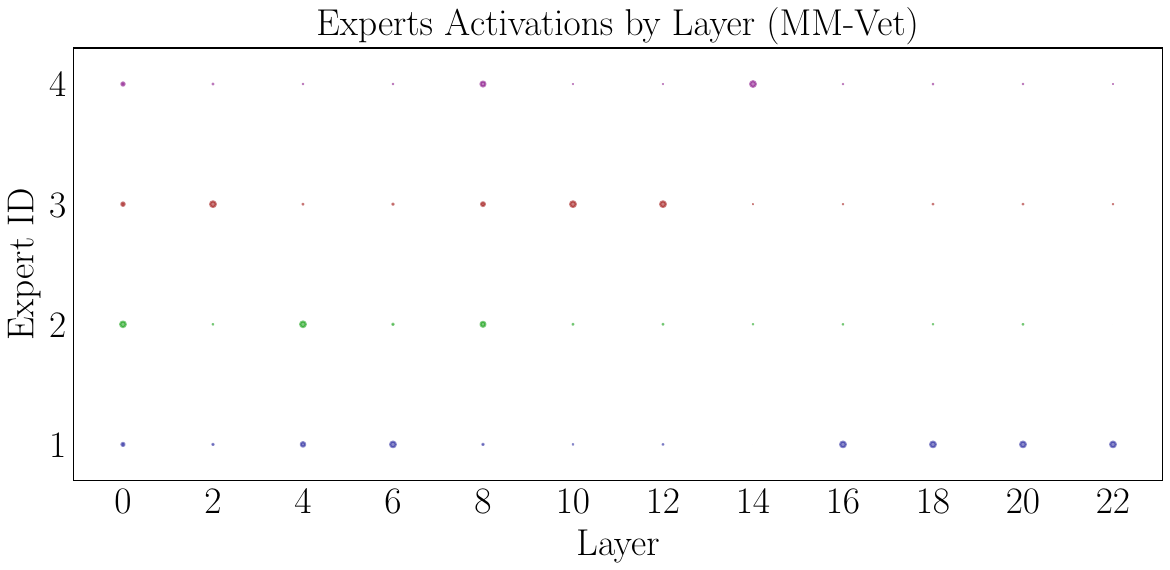}
    }
    \subfigure[MMBench]{
        \includegraphics[width=.45\textwidth]{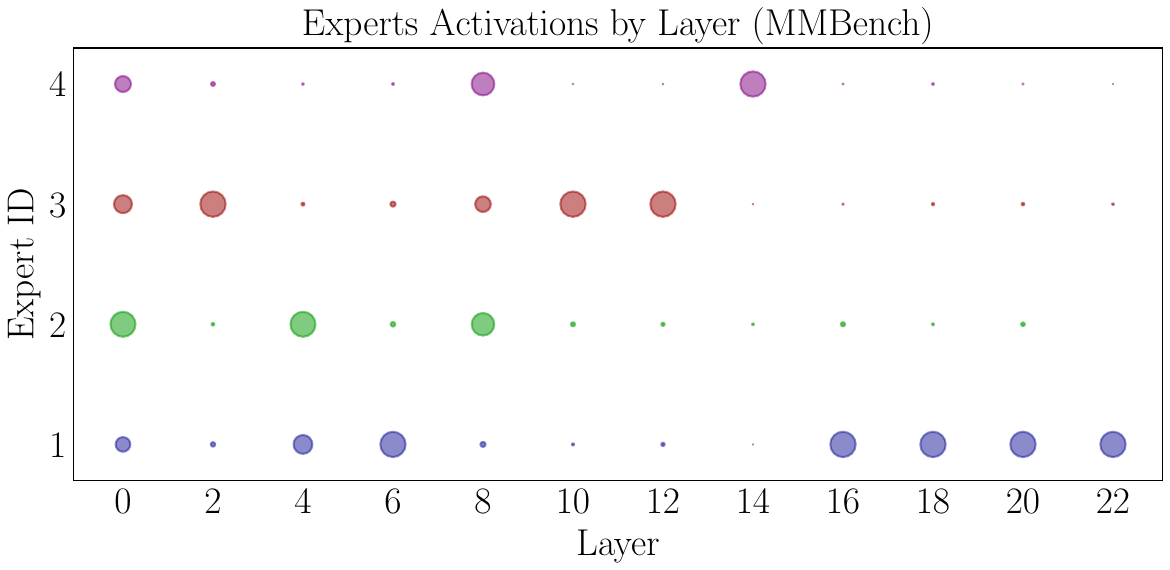}
    }
    \\
    \subfigure[MME]{
        \includegraphics[width=.45\textwidth]{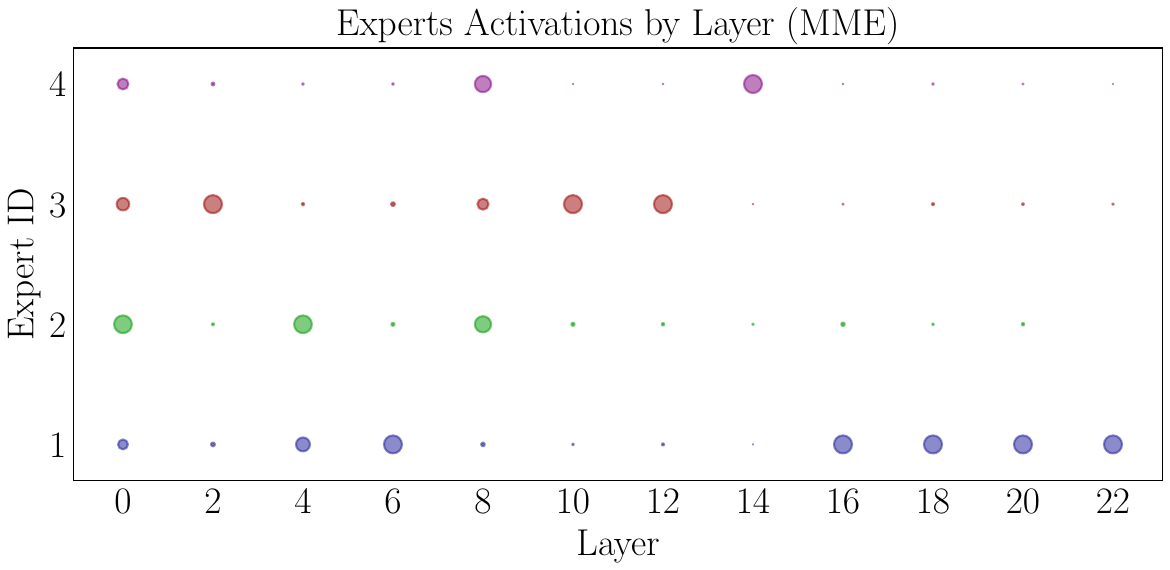}
    }
    \subfigure[POPE]{
        \includegraphics[width=.45\textwidth]{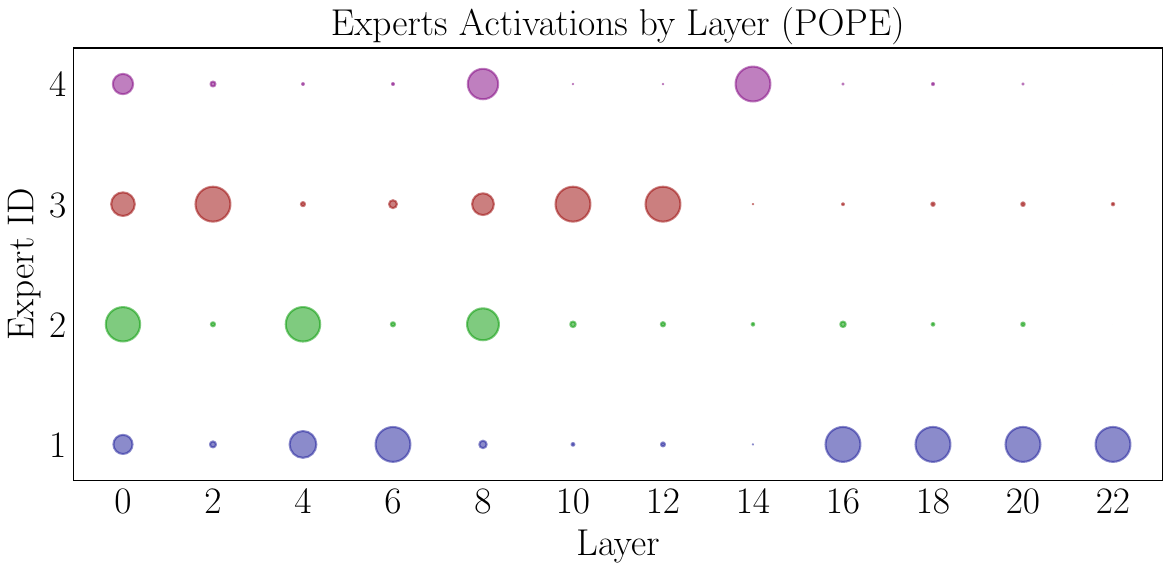}
    }
    \caption{\textbf{Activation frequency of experts on various MoE layers and evaluation tasks using StableLM as backbone.}}
    \label{fig:activation-frequency-stablelm}
\end{figure}

\begin{figure}
    \centering
    \subfigure[ScienceQA]{
        \includegraphics[width=.45\textwidth]{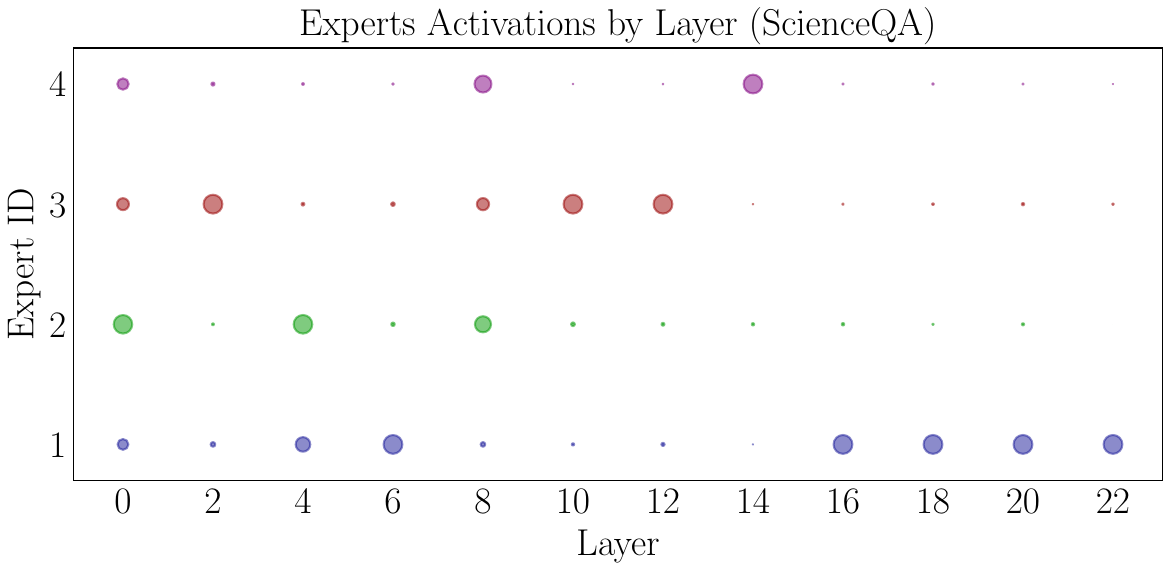}
    }
    \subfigure[TextVQA]{
        \includegraphics[width=.45\textwidth]{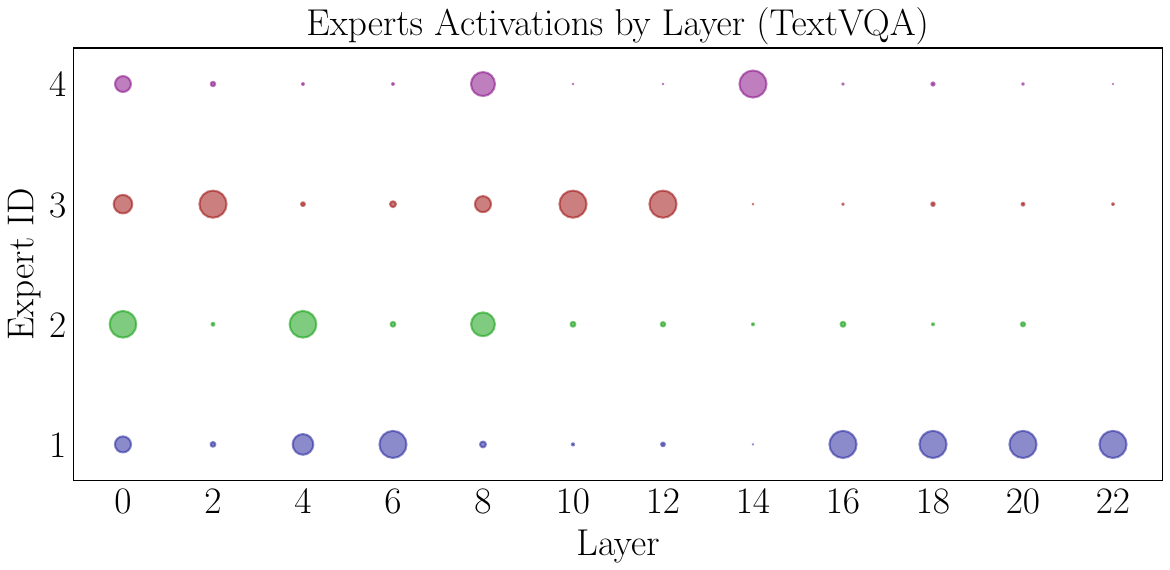}
    }
    \\
    \subfigure[VisWiz]{
        \includegraphics[width=.45\textwidth]{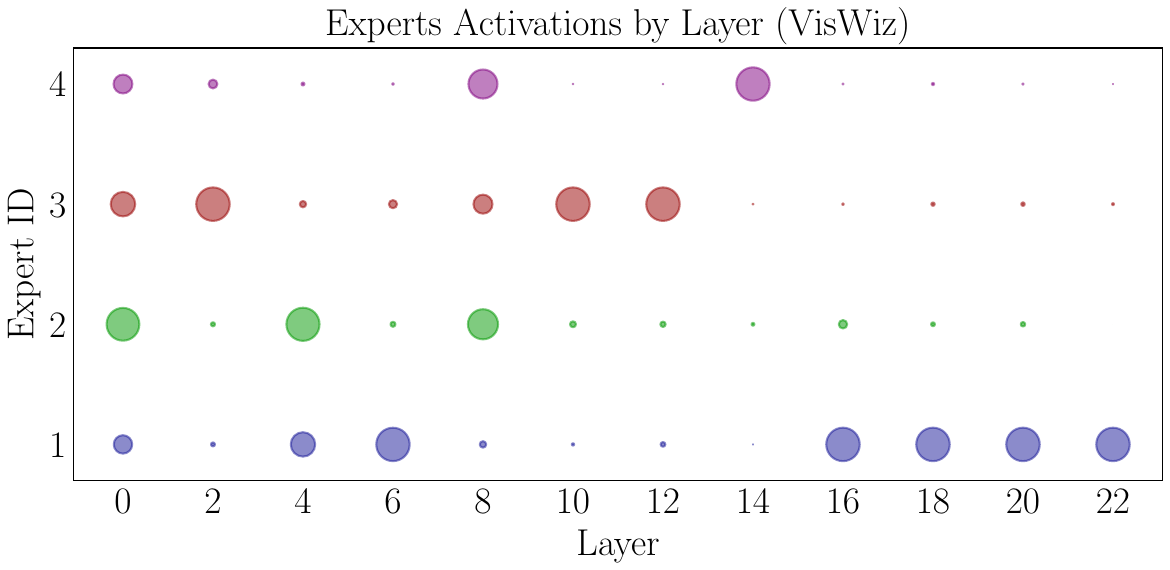}
    }
    \caption{\textbf{Activation frequency of experts on various MoE layers and evaluation tasks using StableLM as backbone.}}
    \label{fig:activation-frequency-stablelm-continue}
\end{figure}

\begin{figure}
    \centering
    \subfigure[GraphQA]{
        \includegraphics[width=.45\textwidth]{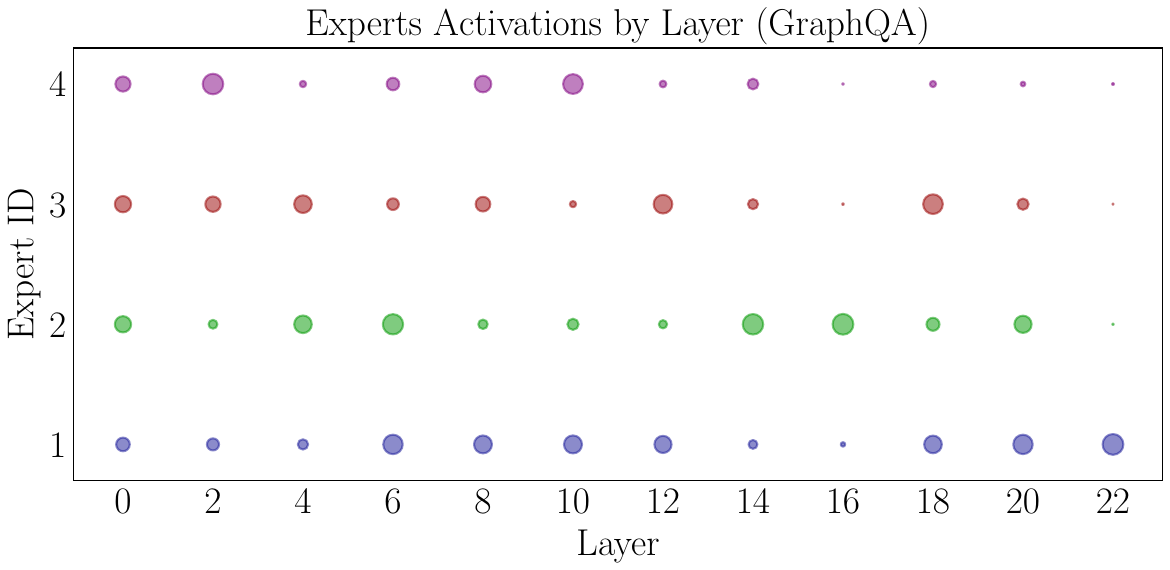}
    }
    \subfigure[LLaVA-Bench]{
        \includegraphics[width=.45\textwidth]{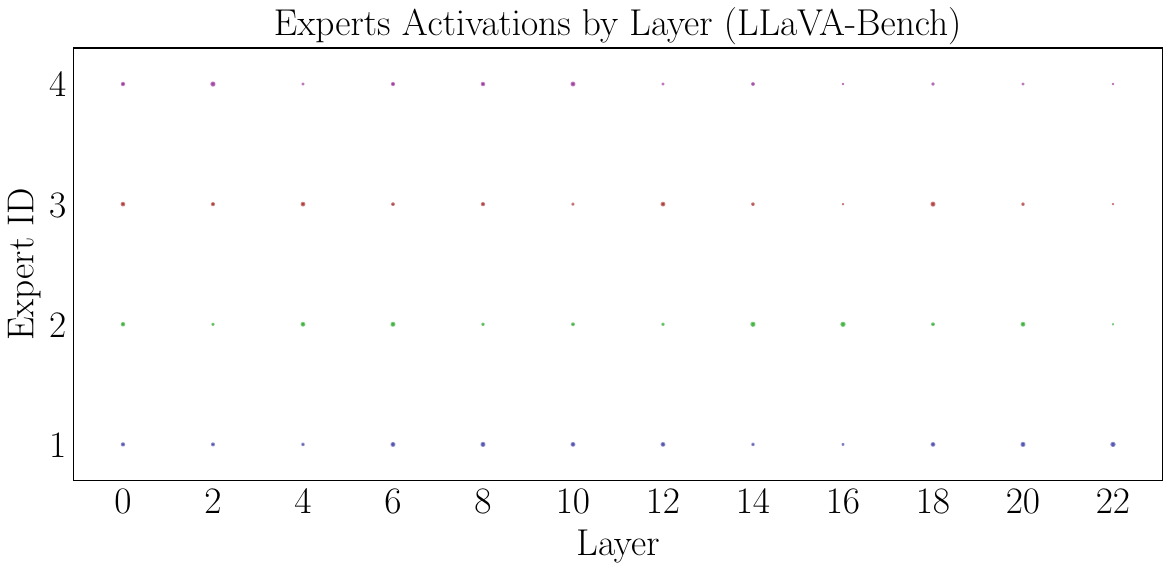}
    }
    \\
    \subfigure[MM-Vet]{
        \includegraphics[width=.45\textwidth]{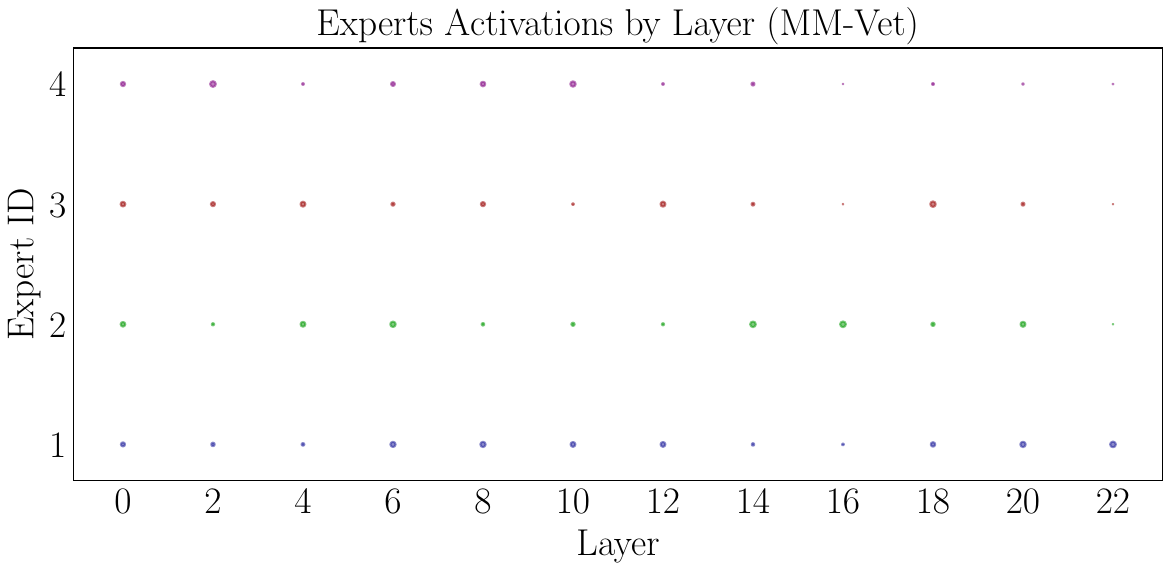}
    }
    \subfigure[MMBench]{
        \includegraphics[width=.45\textwidth]{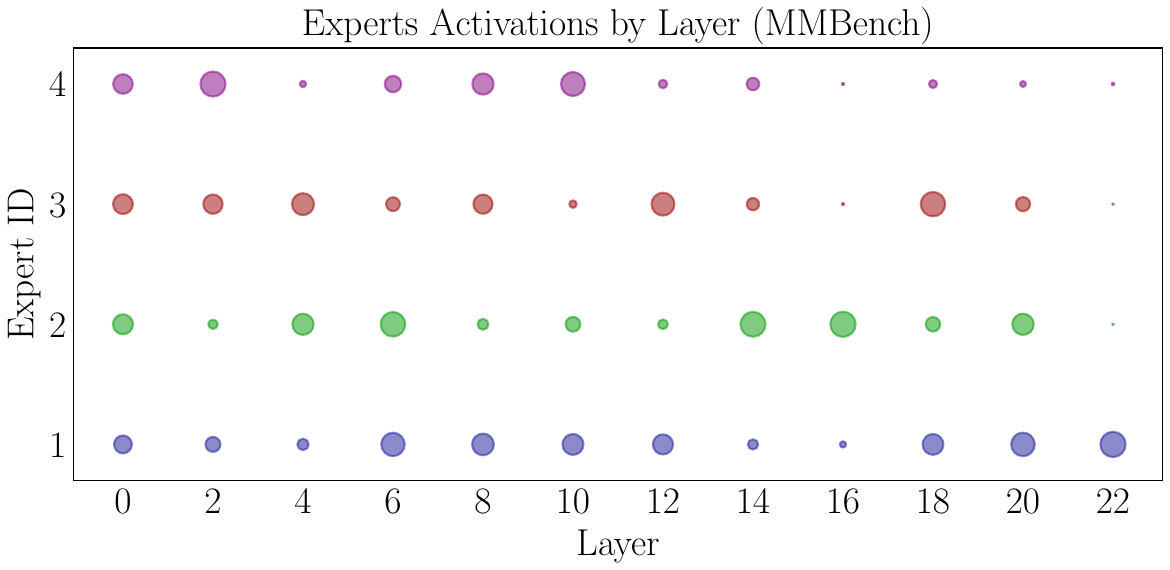}
    }
    \\
    \subfigure[MME]{
        \includegraphics[width=.45\textwidth]{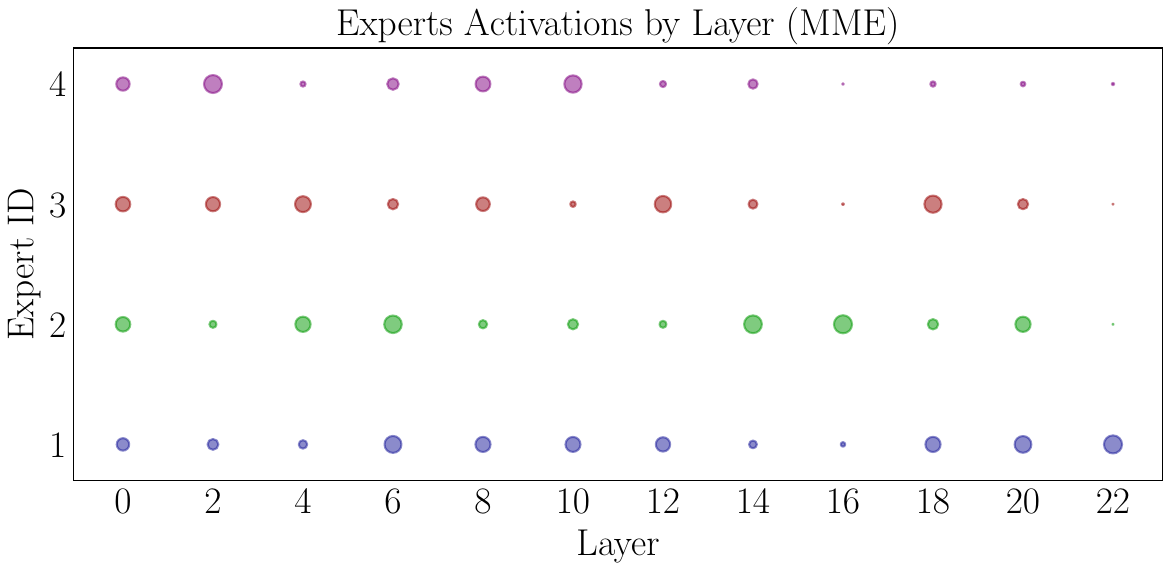}
    }
    \subfigure[POPE]{
        \includegraphics[width=.45\textwidth]{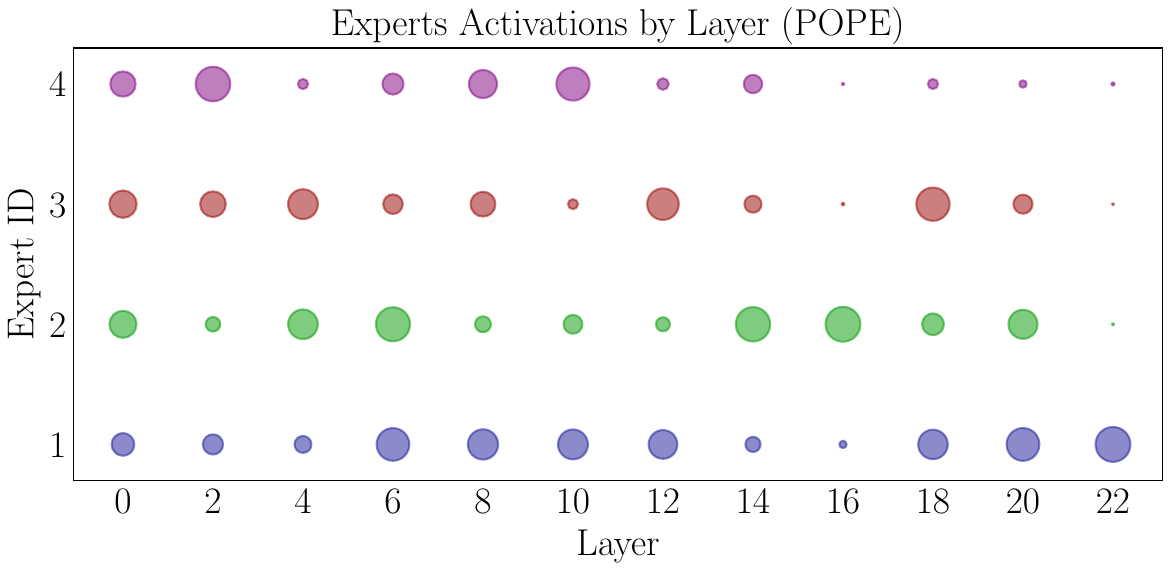}
    }
    \caption{\textbf{Activation frequency of experts on various MoE layers and evaluation tasks using Qwen as backbone.}}
    \label{fig:activation-frequency-qwen}
\end{figure}

\begin{figure}
    \centering
    \subfigure[ScienceQA]{
        \includegraphics[width=.45\textwidth]{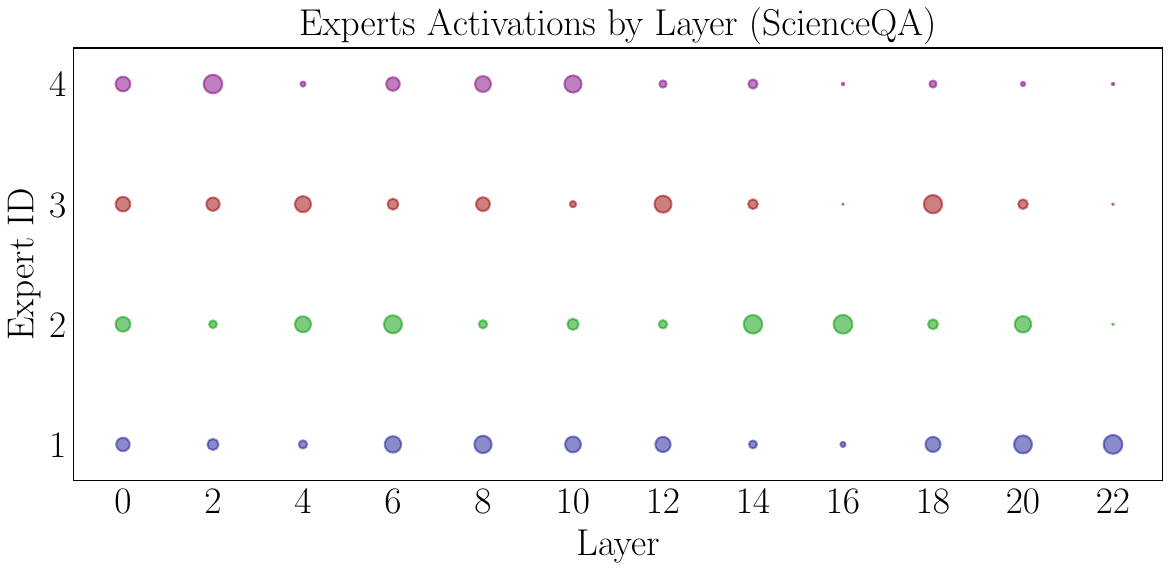}
    }
    \subfigure[TextVQA]{
        \includegraphics[width=.45\textwidth]{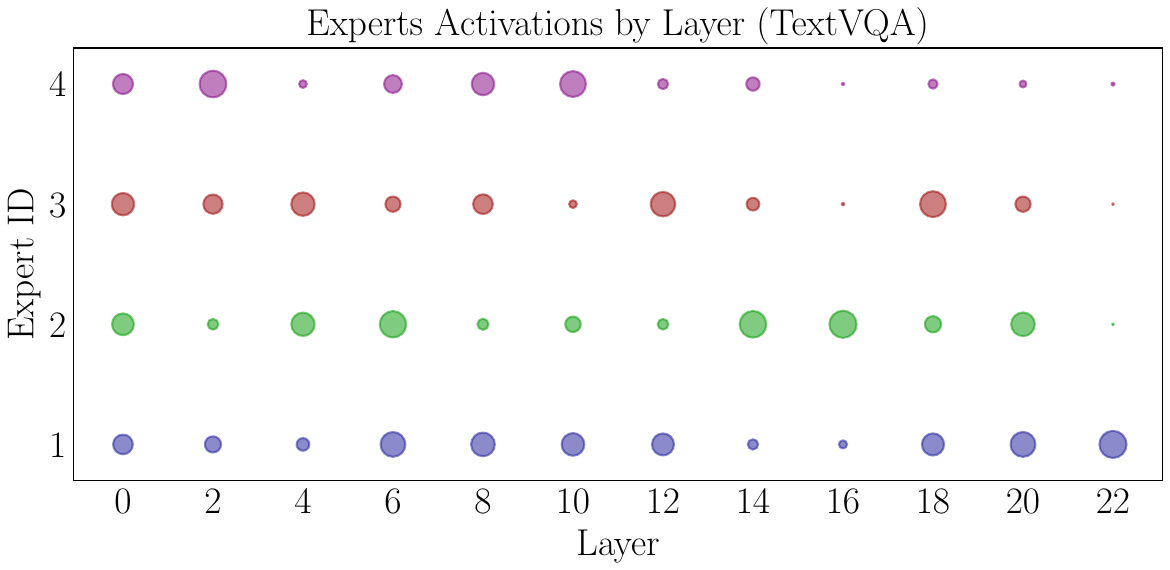}
    }
    \\
    \subfigure[VisWiz]{
        \includegraphics[width=.45\textwidth]{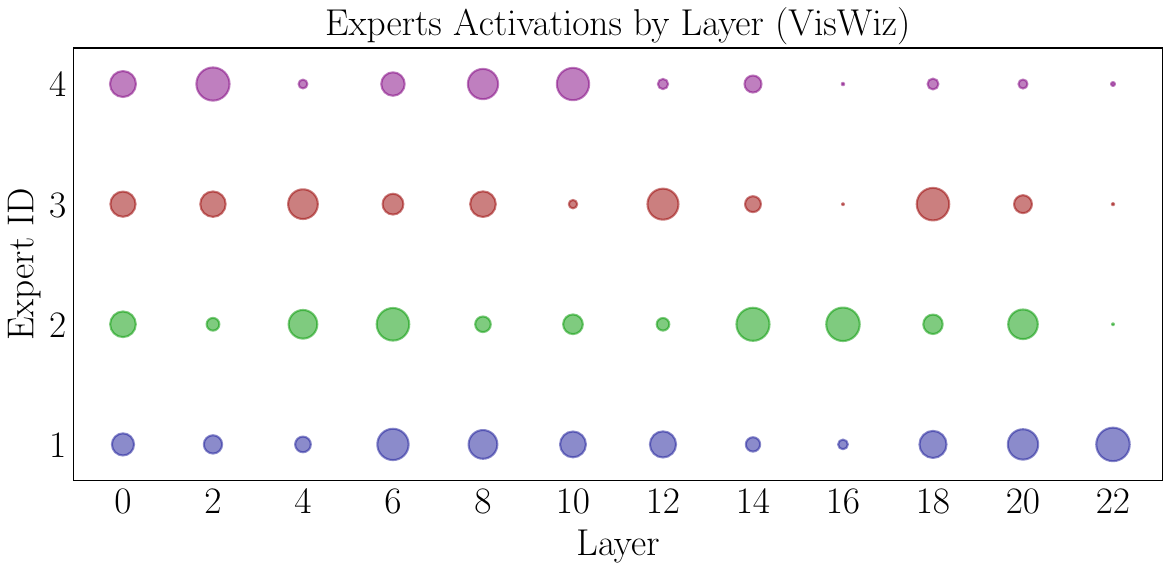}
    }
    \caption{\textbf{Activation frequency of experts on various MoE layers and evaluation tasks using Qwen as backbone.}}
    \label{fig:activation-frequency-qwen-continue}
\end{figure}

\begin{figure}
    \centering
    \subfigure[GraphQA]{
        \includegraphics[width=.45\textwidth]{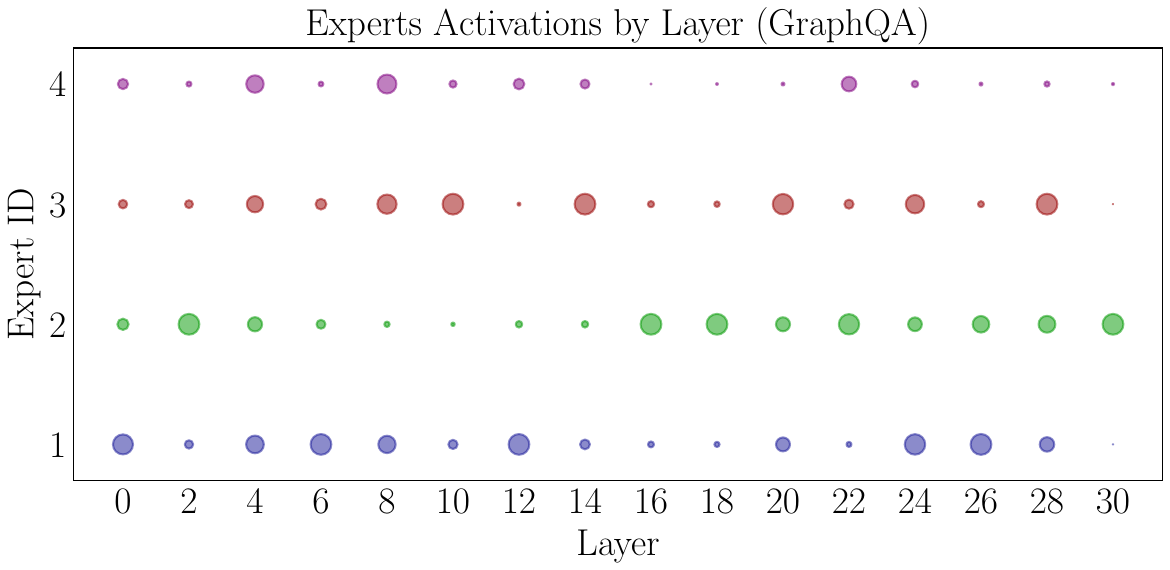}
    }
    \subfigure[LLaVA-Bench]{
        \includegraphics[width=.45\textwidth]{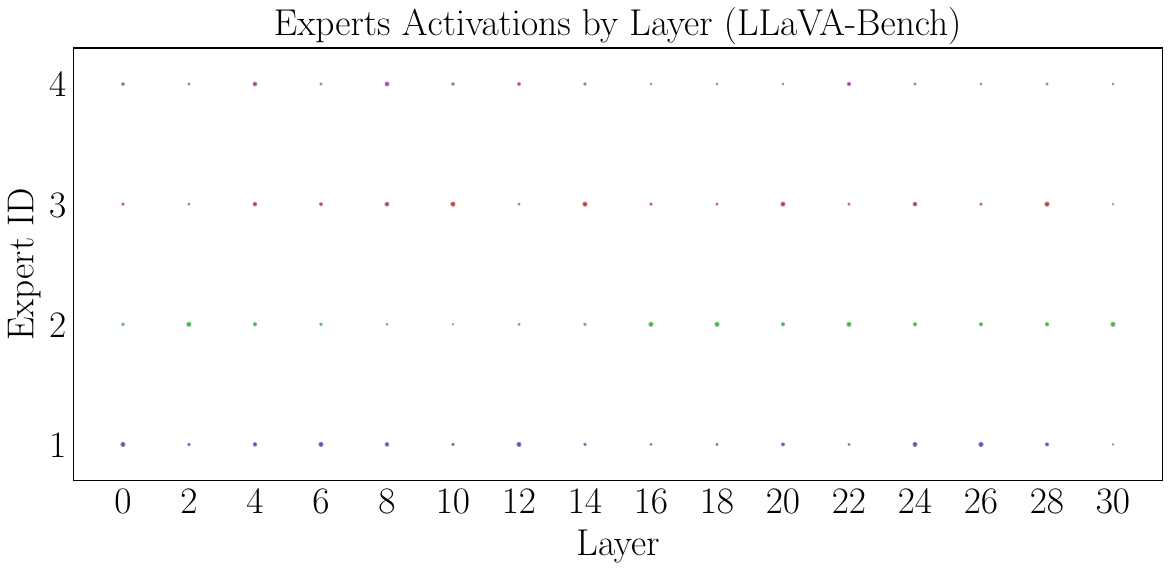}
    }
    \\
    \subfigure[MM-Vet]{
        \includegraphics[width=.45\textwidth]{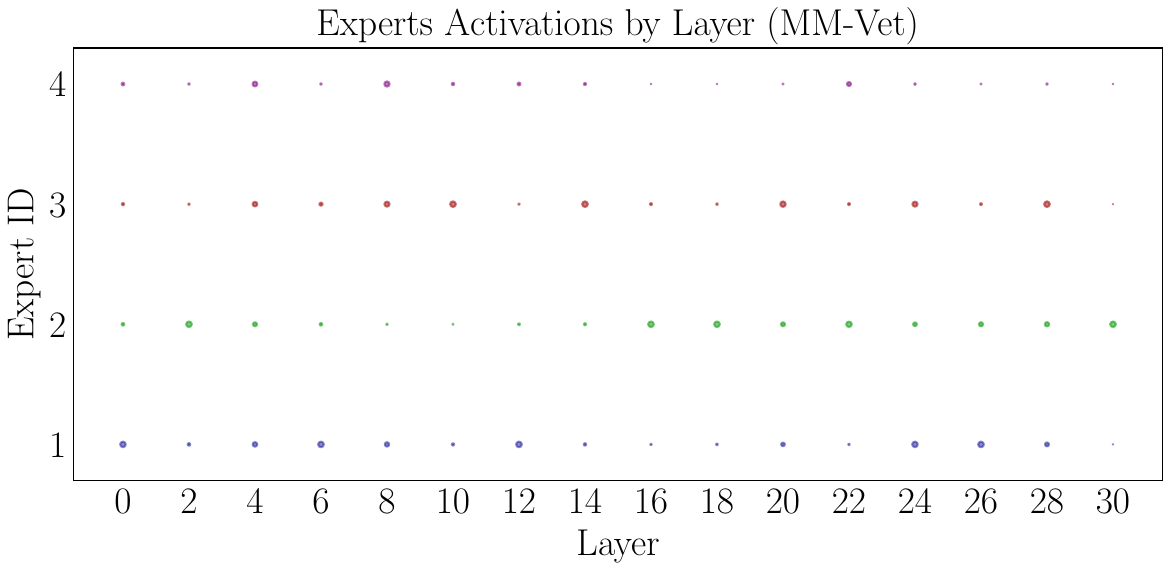}
    }
    \subfigure[MMBench]{
        \includegraphics[width=.45\textwidth]{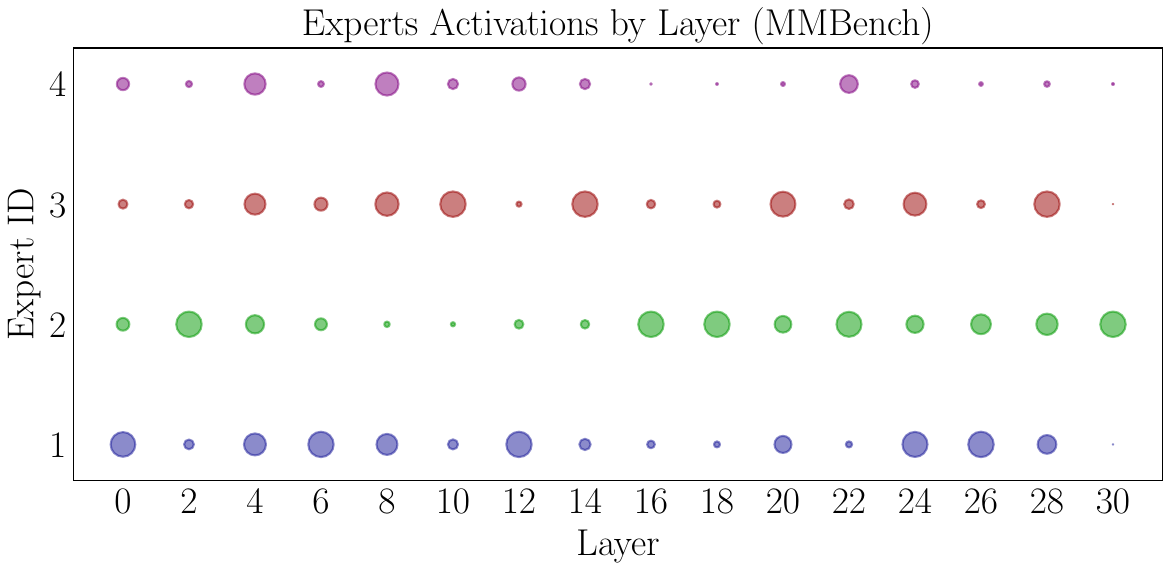}
    }
    \\
    \subfigure[MME]{
        \includegraphics[width=.45\textwidth]{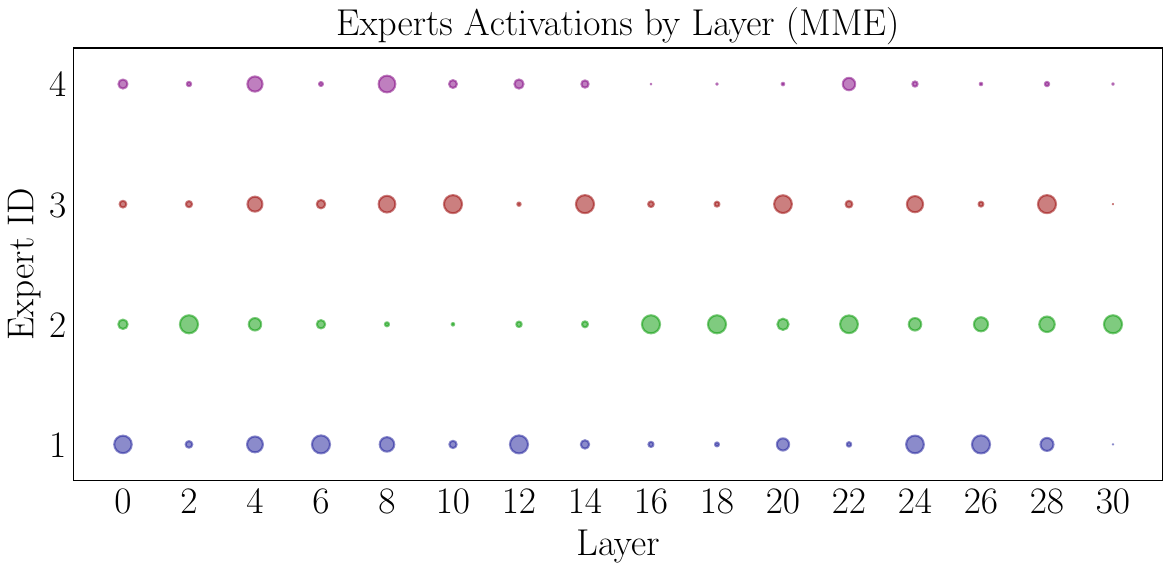}
    }
    \subfigure[POPE]{
        \includegraphics[width=.45\textwidth]{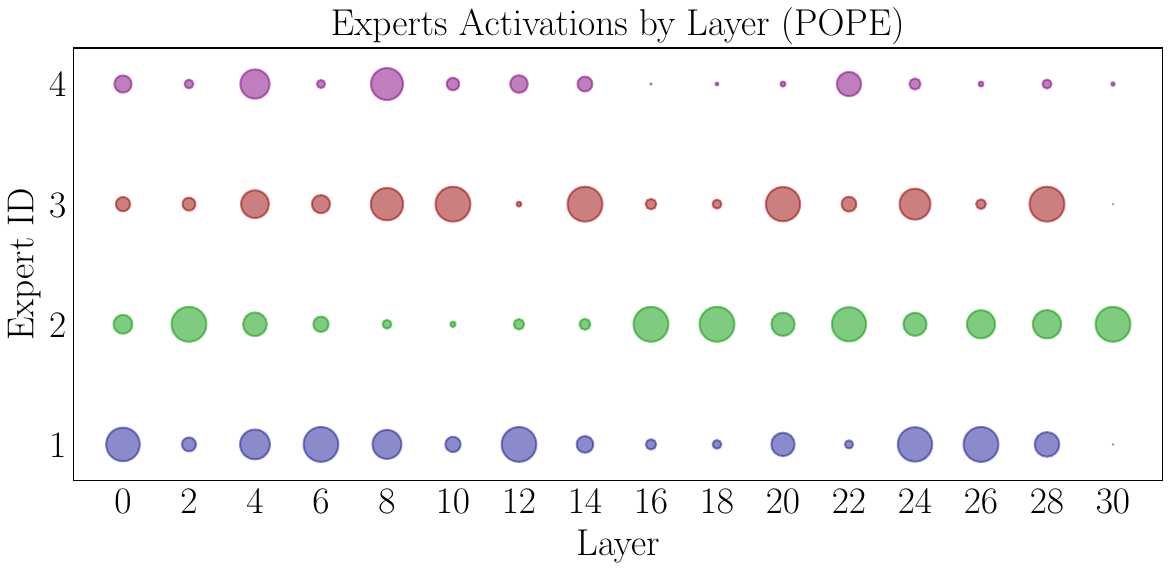}
    }
    \caption{\textbf{Activation frequency of experts on various MoE layers and evaluation tasks using Phi-2 as backbone.}}
    \label{fig:activation-frequency-phi}
\end{figure}

\begin{figure}
    \centering
    \subfigure[ScienceQA]{
        \includegraphics[width=.45\textwidth]{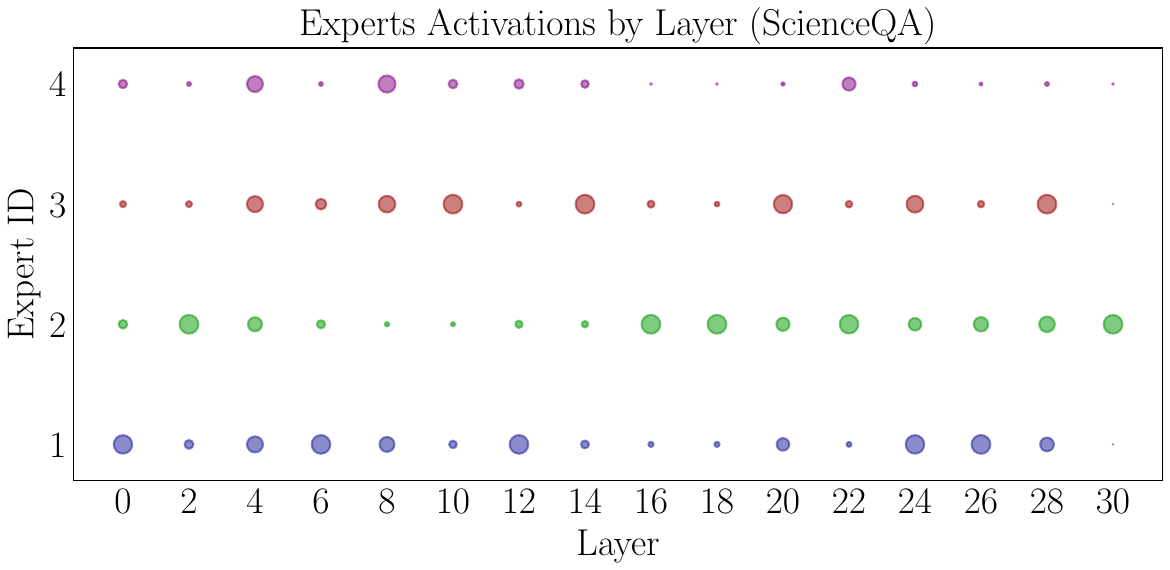}
    }
    \subfigure[TextVQA]{
        \includegraphics[width=.45\textwidth]{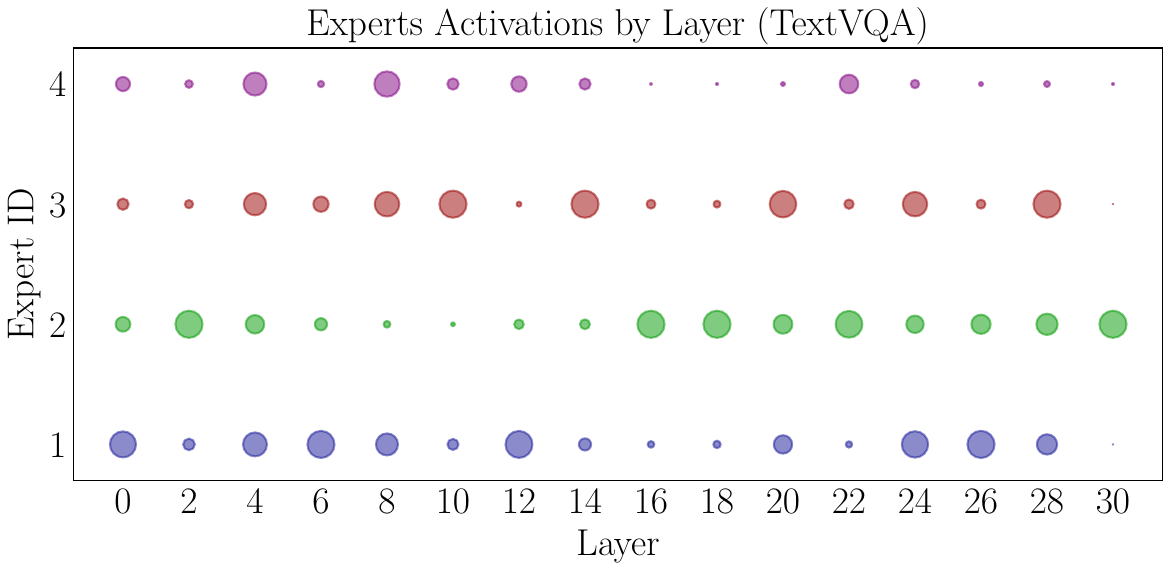}
    }
    \\
    \subfigure[VisWiz]{
        \includegraphics[width=.45\textwidth]{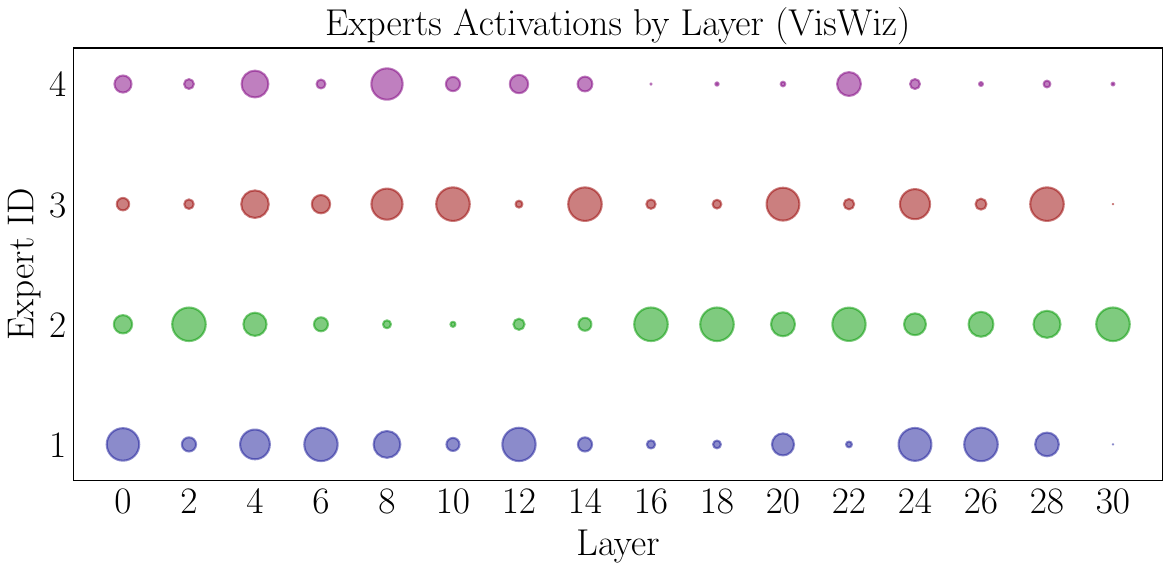}
    }
    \caption{\textbf{Activation frequency of experts on various MoE layers and evaluation tasks using Phi-2 as backbone.}}
    \label{fig:activation-frequency-phi-continue}
\end{figure}

We present the activation frequency of experts across various MoE layers and evaluation tasks using different backbones: StableLM-1.6B (Figures~\ref{fig:activation-frequency-stablelm} and~\ref{fig:activation-frequency-stablelm-continue}), Qwen-1.8B (Figures~\ref{fig:activation-frequency-qwen} and~\ref{fig:activation-frequency-qwen-continue}), and Phi-2-2.7B (Figures~\ref{fig:activation-frequency-phi} and~\ref{fig:activation-frequency-phi-continue}).
The results suggest that compared to the StableLM-1.6B backbone, experts are more uniformly activated for models utilizing Qwen-1.8B and Phi-2-2.7B as backbone LLMs.

\begin{figure}[!t]
    \centering
    \includegraphics[width=.55\textwidth]{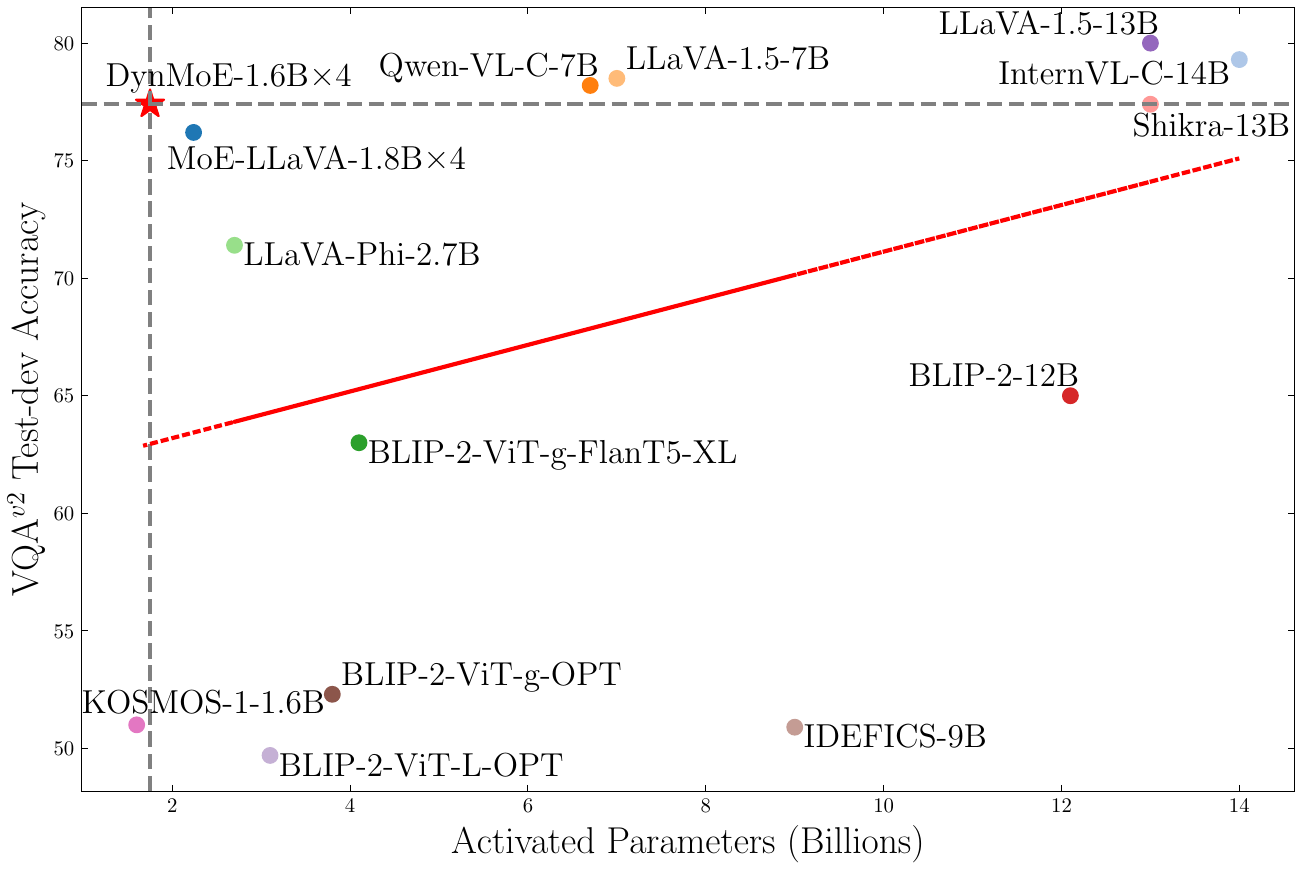}
    \caption{
        \textbf{Comparing the performance efficiency of models.}
        The $x$-axis represents the number of activated parameters, while the $y$-axis shows the performance on the Visual Question Answering (VQA) task.
    }
    \label{fig:illustration-sparse}
\end{figure}

\subsection{Average Top-$k$}

In Figures~\ref{fig:topk-qwen} and~\ref{fig:topk-phi} , we illustrate the average top-$k$ of \algabbr models using Qwen and Phi-2 as backbone LLMs.

    {
        \begin{figure}
            \centering
            \includegraphics[width=\textwidth]{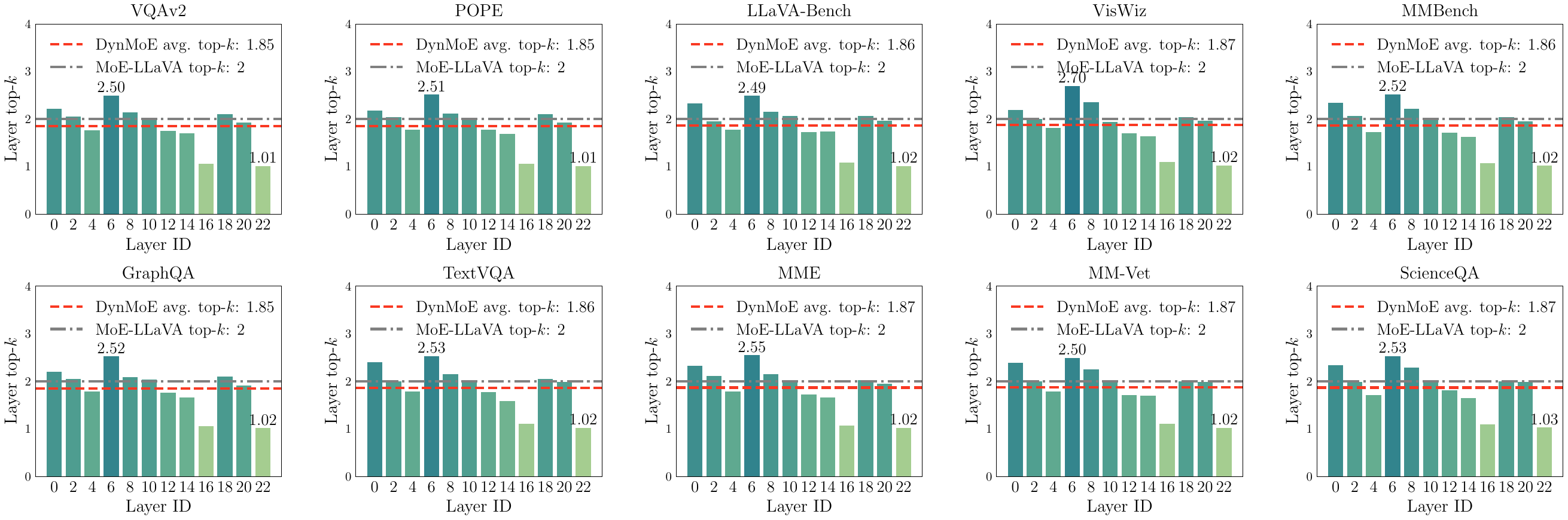}
            \caption{\textbf{Average top-$k$ activated experts of \algabbr on vision-language benchmarks, using Qwen as language backbone.}}
            \label{fig:topk-qwen}
        \end{figure}
    }

    {
        \begin{figure}
            \centering
            \includegraphics[width=\textwidth]{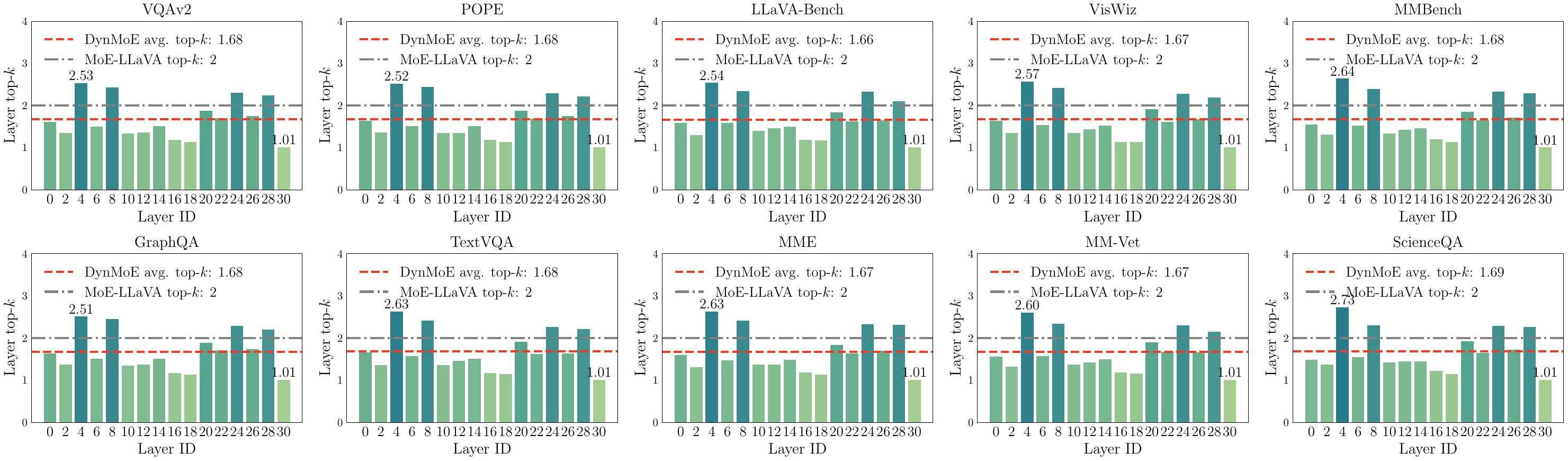}
            \caption{\textbf{Average top-$k$ activated experts of \algabbr on vision-language benchmarks, using Phi-2 as language backbone.}}
            \label{fig:topk-phi}
        \end{figure}
    }

\subsection{Layer-wise Expert Similarity Matrix}

In Figures~\ref{fig:sim-stablelm},~\ref{fig:sim-qwen}, and~\ref{fig:sim-phi}, we illustrate the similarities between various expert representations, specifically, different rows of $\mW_g$ across multiple MoE layers. These comparisons utilize StableLM-1.6B, Qwen-1.8B, and Phi-2-2.7B as the backbone LLMs. The findings demonstrate that these expert representations are nearly orthogonal, suggesting that different experts capture diverse features, which could potentially enhance the model's capacity.

\subsection{Visualization of $\mG$}

In Figures~\ref{fig:gates-stablelm},~\ref{fig:gates-qwen}, and~\ref{fig:gates-phi}, we present the values of the learned threshold $\mG$, employing StableLM-1.6B, Qwen-1.8B, and Phi-2-2.7B as the backbone LLMs. The results reveal that for each MoE layer, there is one expert that is more readily activated. This observation is consistent with the design of Deepseek-MoE~\citep{dai2024deepseekmoe}.

{
\begin{figure}
    \centering
    \includegraphics[width=\textwidth]{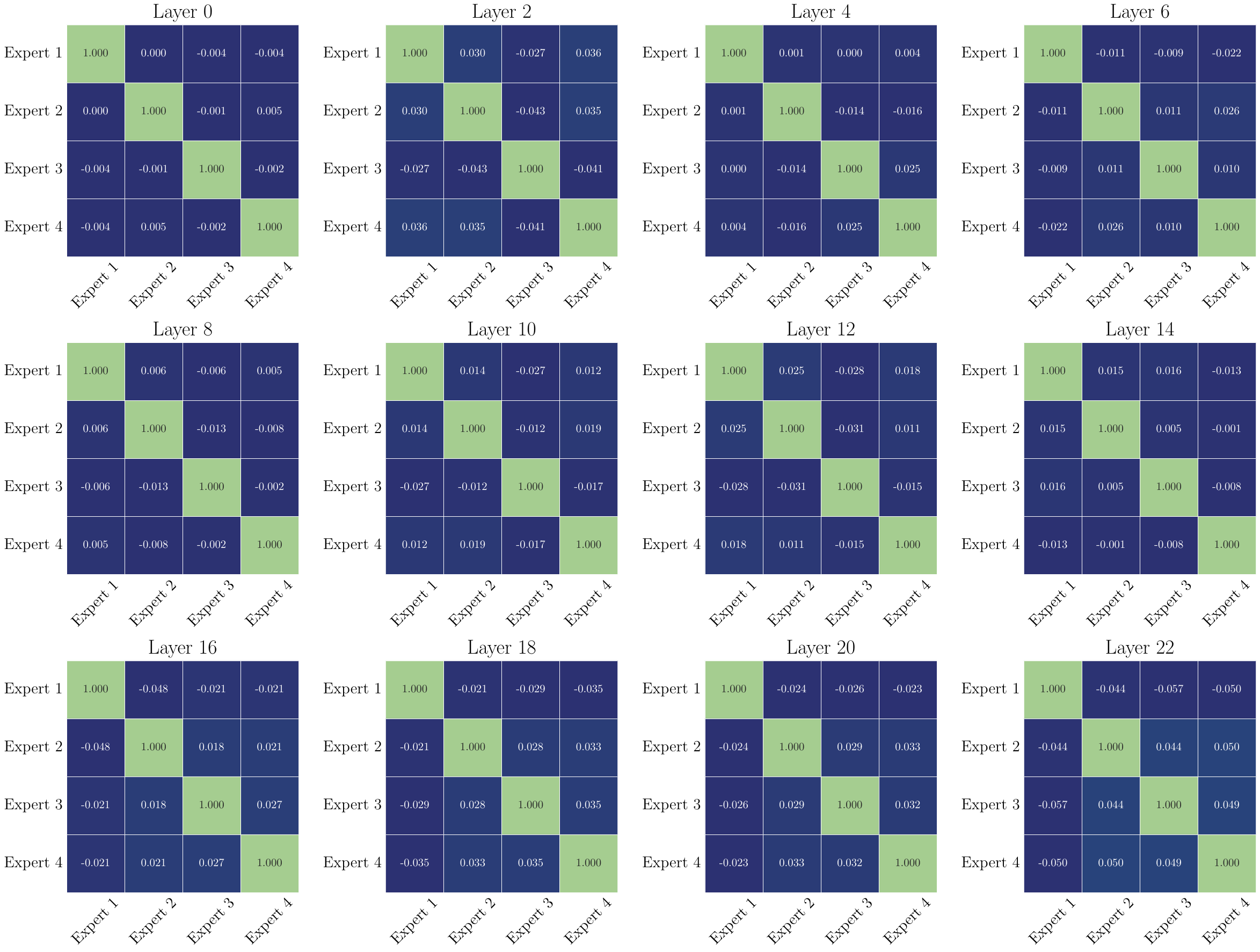}
    \caption{\textbf{Layer-wise expert similarity matrix (StableLM).} We record the experts' cosine similarity per layer during test time. It turns out the cosine similarity between experts is close to 0.}
    \label{fig:sim-stablelm}
\end{figure}

\begin{figure}
    \centering
    \includegraphics[width=\textwidth]{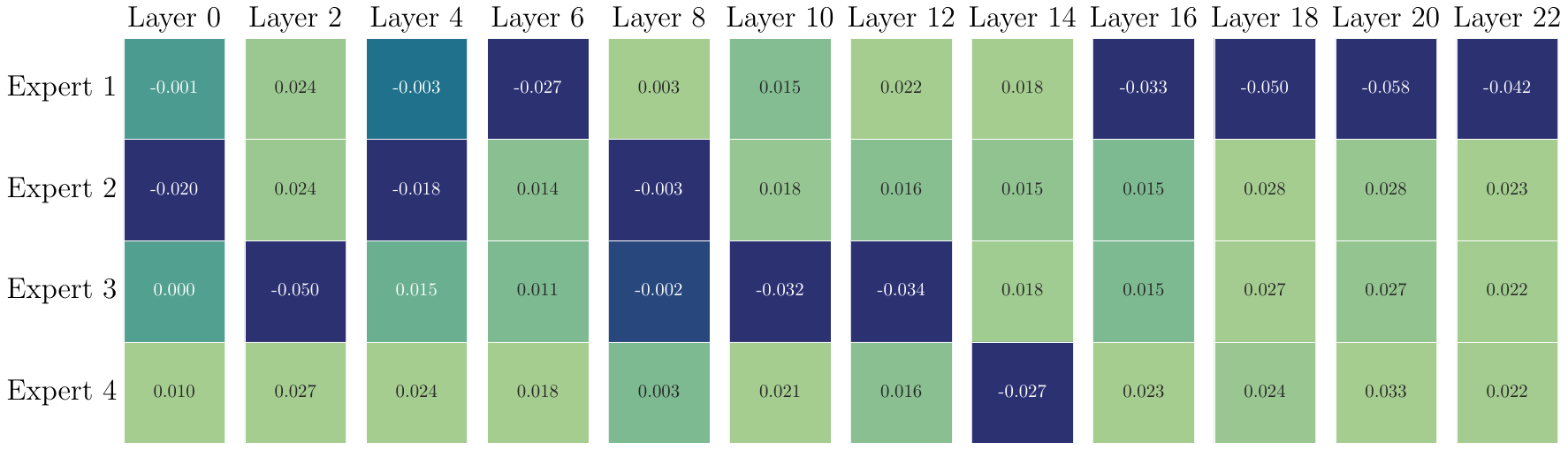}
    \caption{\textbf{Layer-wise expert activation threshold (StableLM).} Darker-colored experts are more likely to be activated compared to lighter-colored experts.}
    \label{fig:gates-stablelm}
\end{figure}
}

{
\begin{figure}
    \centering
    \includegraphics[width=\textwidth]{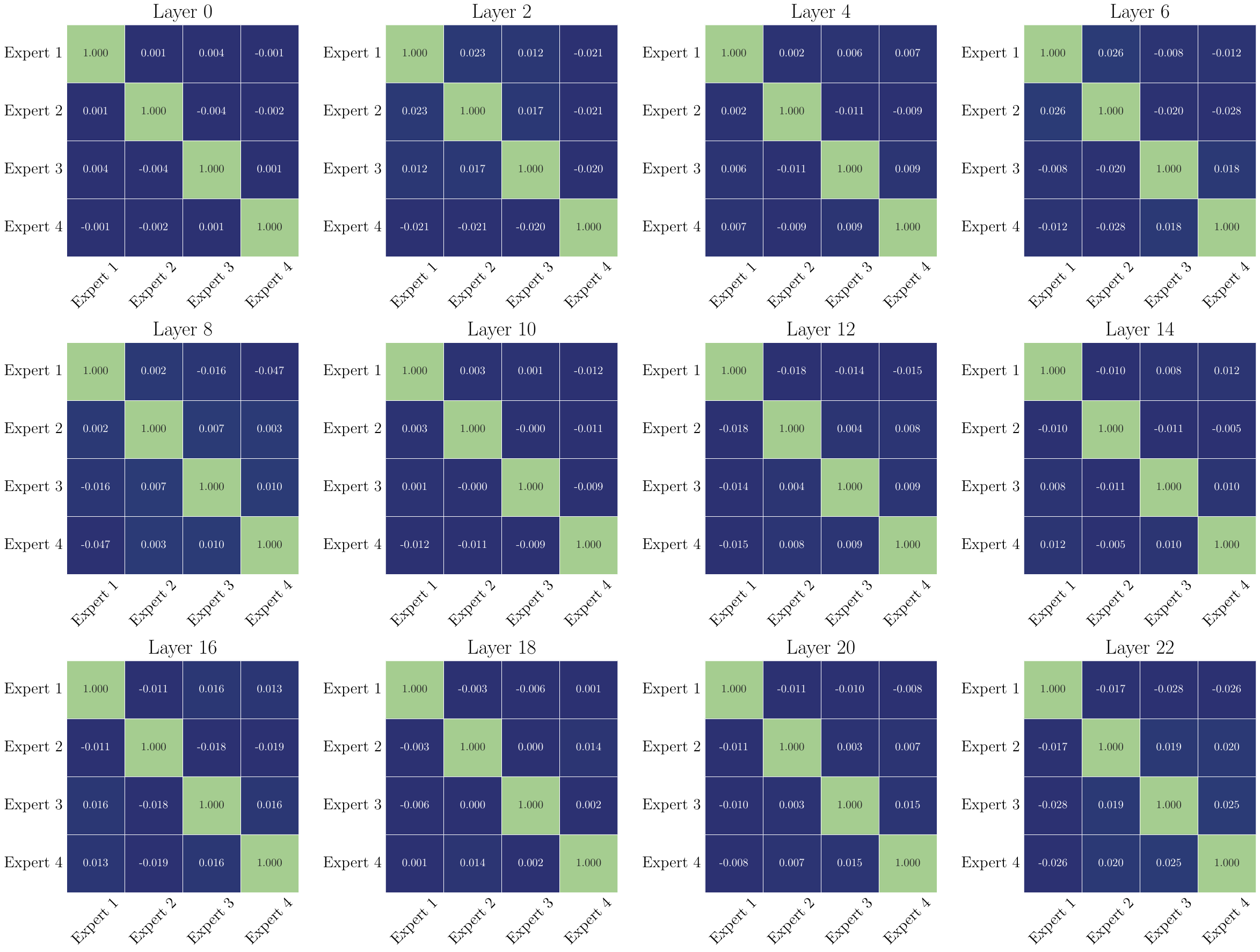}
    \caption{\textbf{Layer-wise expert similarity matrix (Qwen).} We record the experts' cosine similarity per layer during test time. It turns out the cosine similarity between experts is close to 0.}
    \label{fig:sim-qwen}
\end{figure}

\begin{figure}
    \centering
    \includegraphics[width=\textwidth]{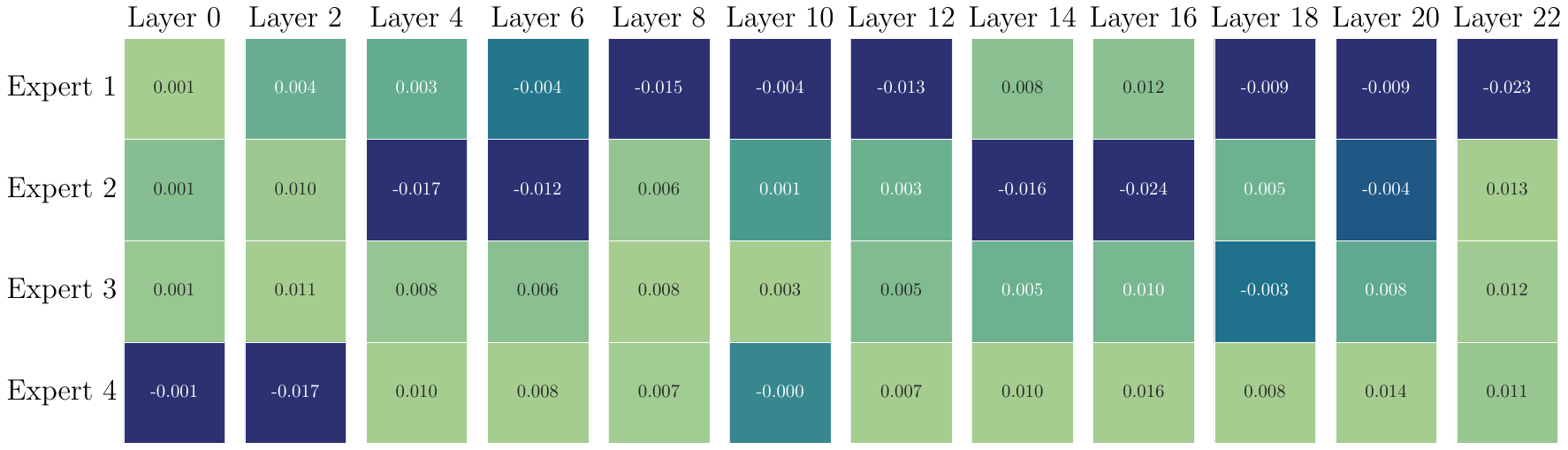}
    \caption{\textbf{Layer-wise expert activation threshold (Qwen).} Darker-colored experts are more likely to be activated compared to lighter-colored experts.}
    \label{fig:gates-qwen}
\end{figure}
}

{
\begin{figure}
    \centering
    \includegraphics[width=\textwidth]{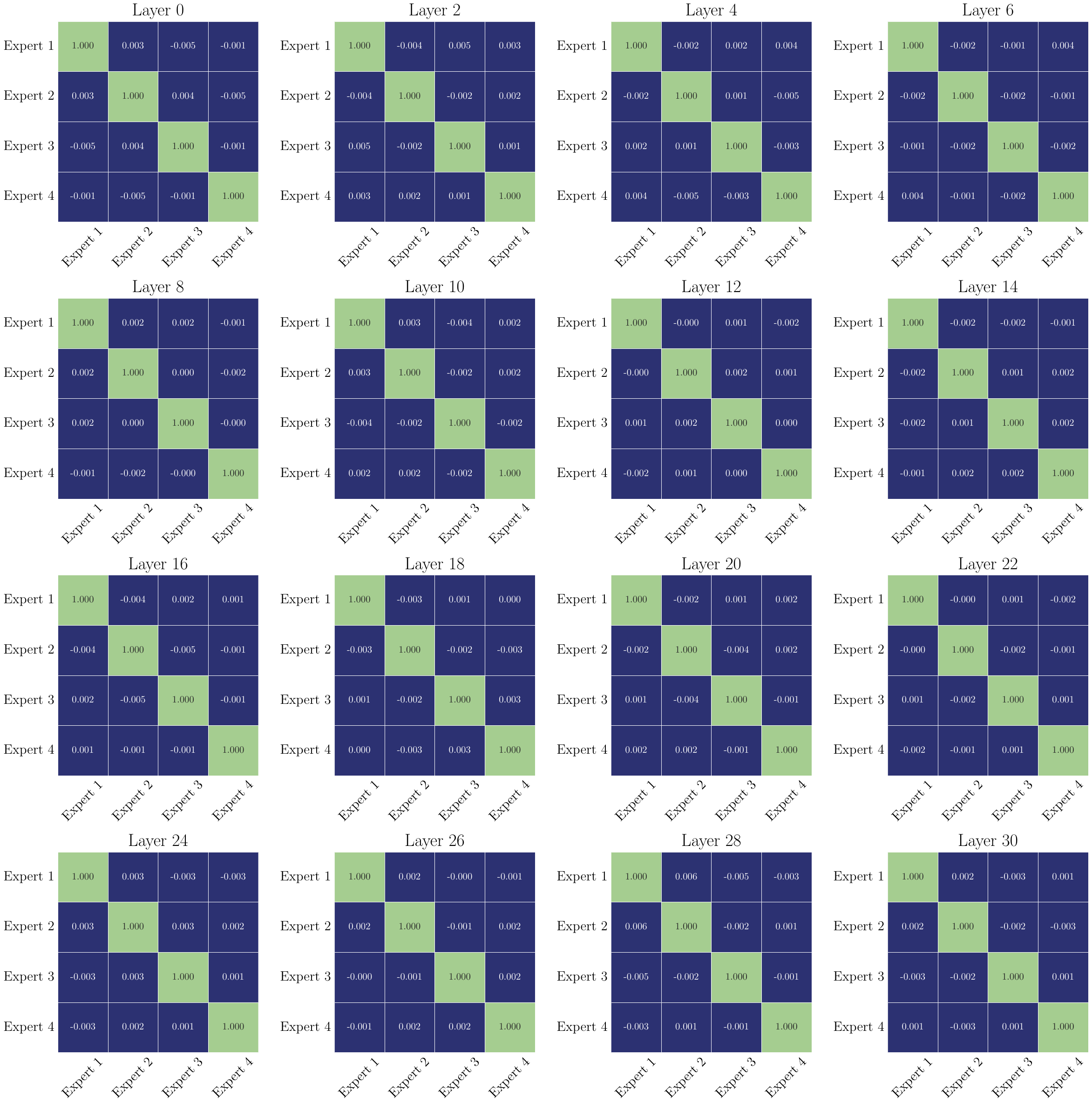}
    \caption{\textbf{Layer-wise expert similarity matrix (Phi-2).} We record the experts' cosine similarity per layer during test time. It turns out the cosine similarity between experts is close to 0.}
    \label{fig:sim-phi}
\end{figure}

\begin{figure}
    \centering
    \includegraphics[width=\textwidth]{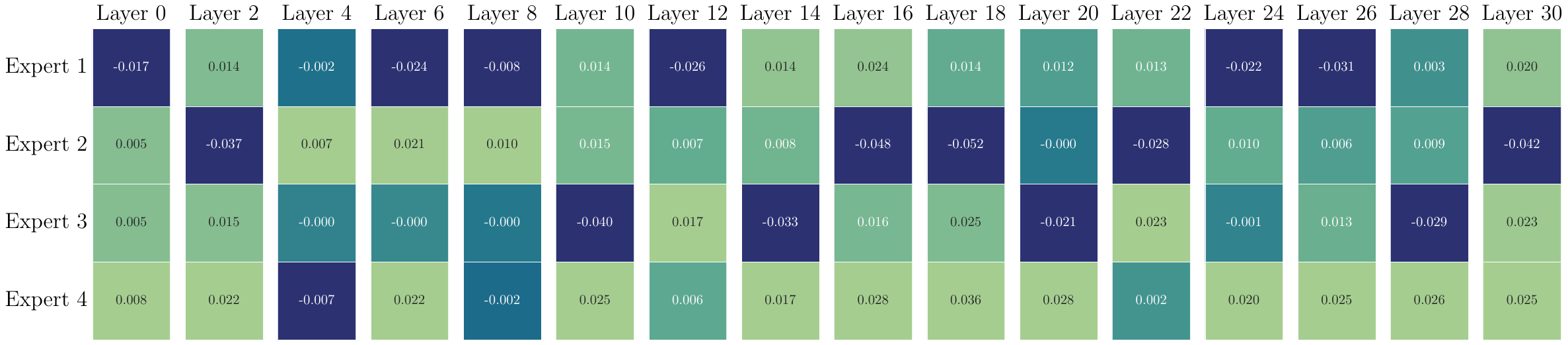}
    \caption{\textbf{Layer-wise expert activation threshold (Phi-2).} Darker-colored experts are more likely to be activated compared to lighter-colored experts.}
    \label{fig:gates-phi}
\end{figure}
}

\begin{figure}
    \centering
    \includegraphics[width=0.6\linewidth]{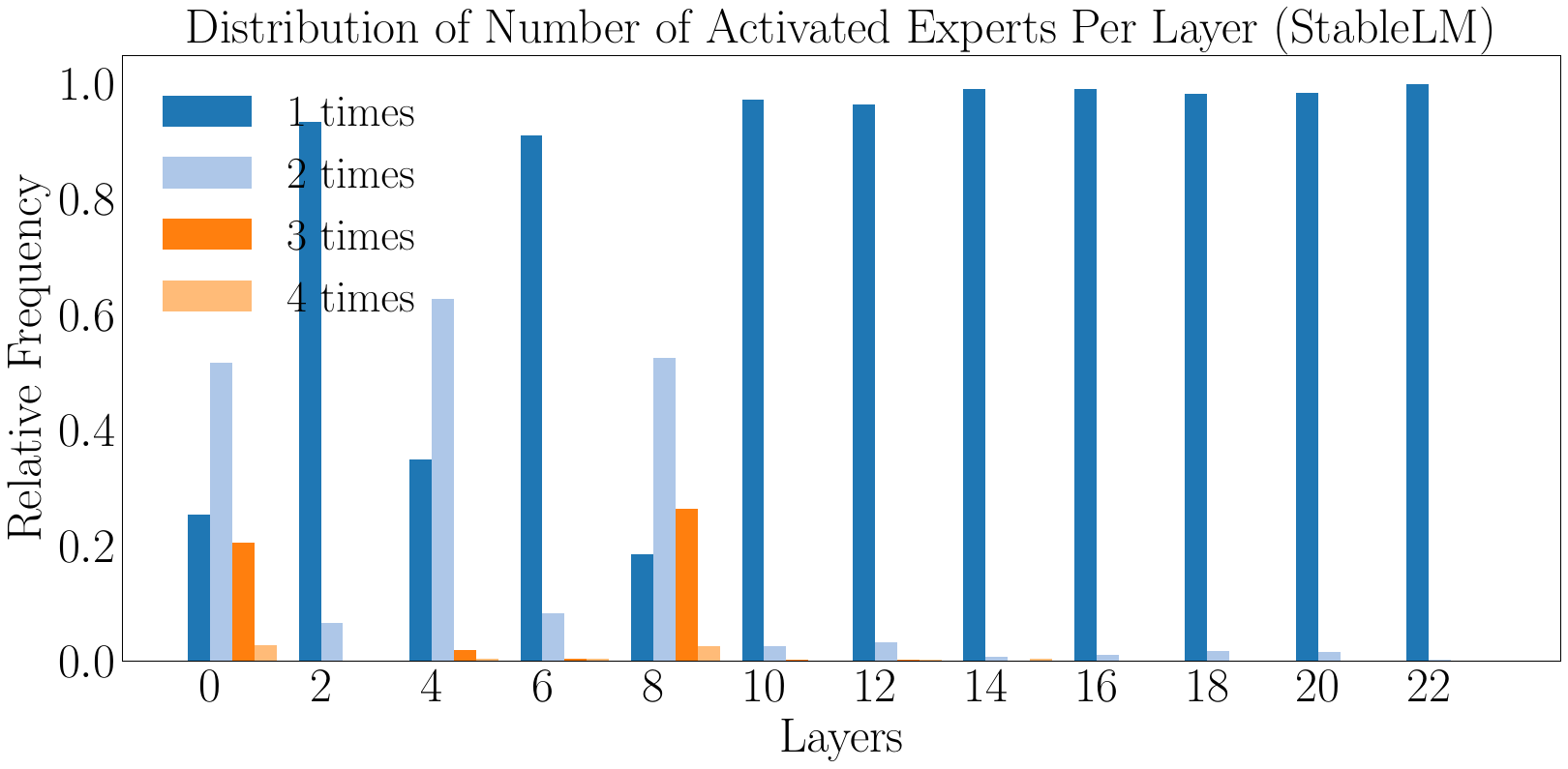} \\
    \includegraphics[width=0.6\linewidth]{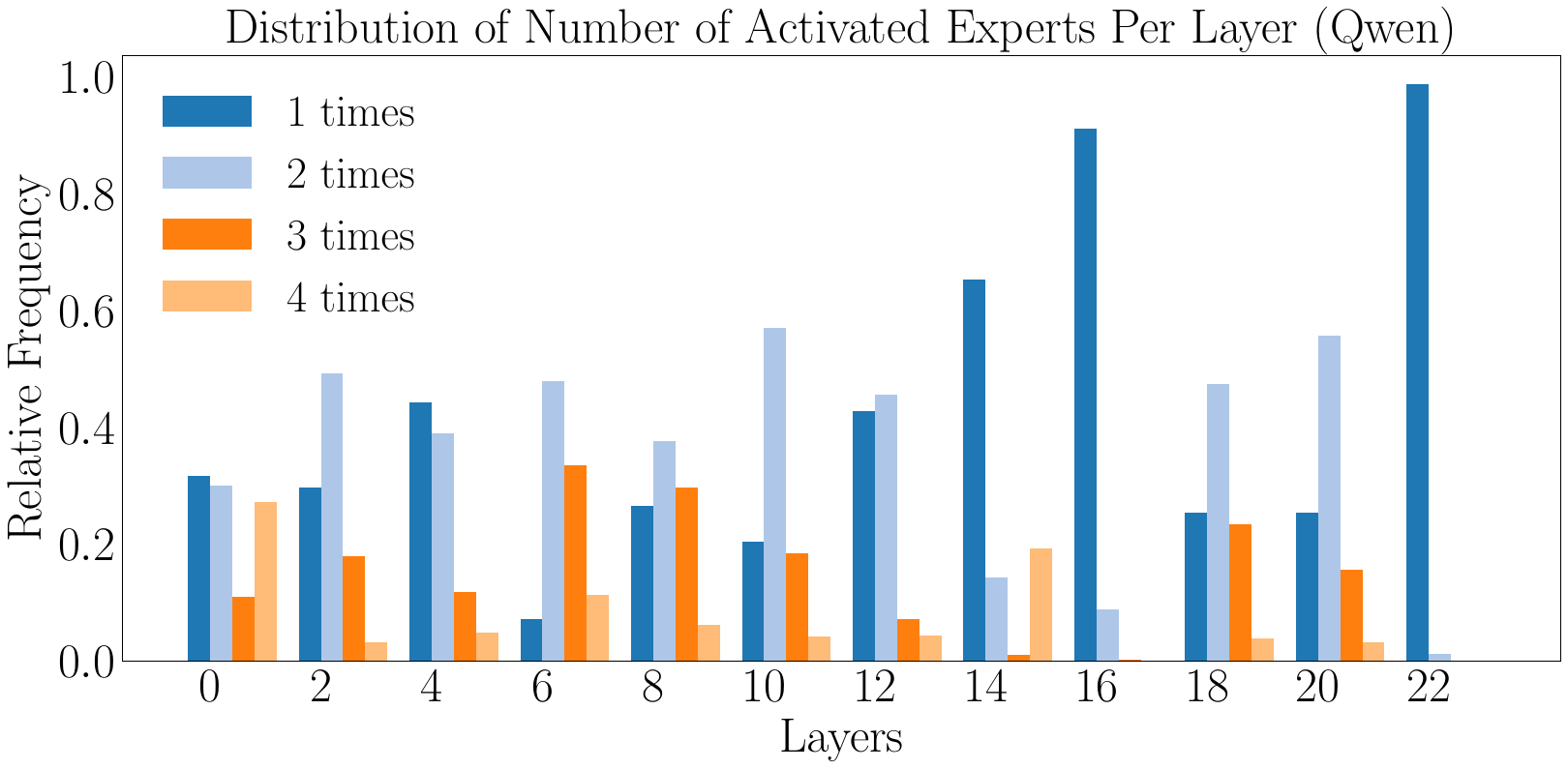} \\
    \includegraphics[width=0.6\linewidth]{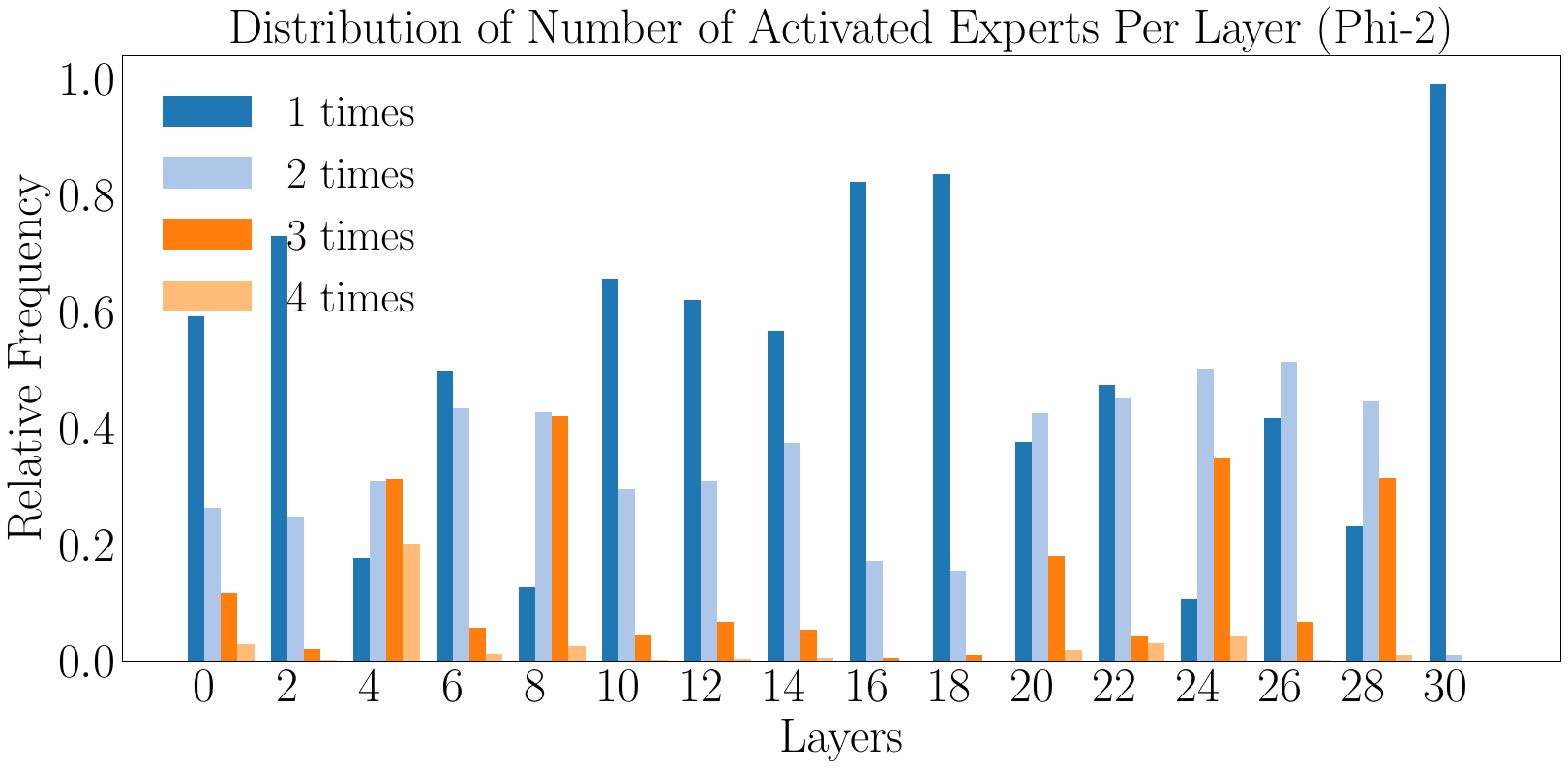}
    \caption{\textbf{Distribution of number of activated experts in each layer.} We report the results of StableLM, Qwen, and Phi-2 models, respectively.}
    \label{fig:distribution-of-number-of-experts}
\end{figure}

\begin{figure}
    \centering
    \includegraphics[width=0.45\linewidth]{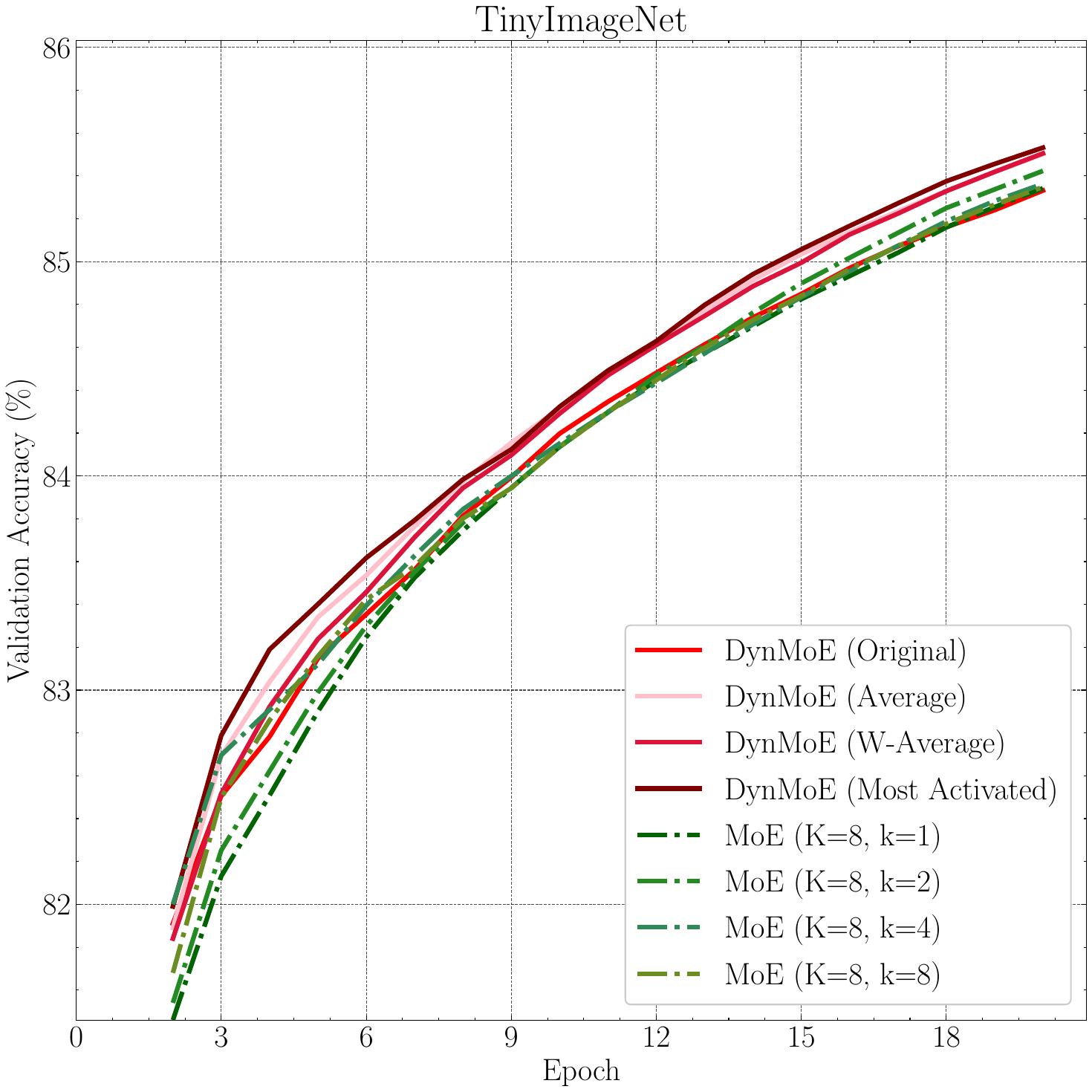}
    \includegraphics[width=0.45\linewidth]{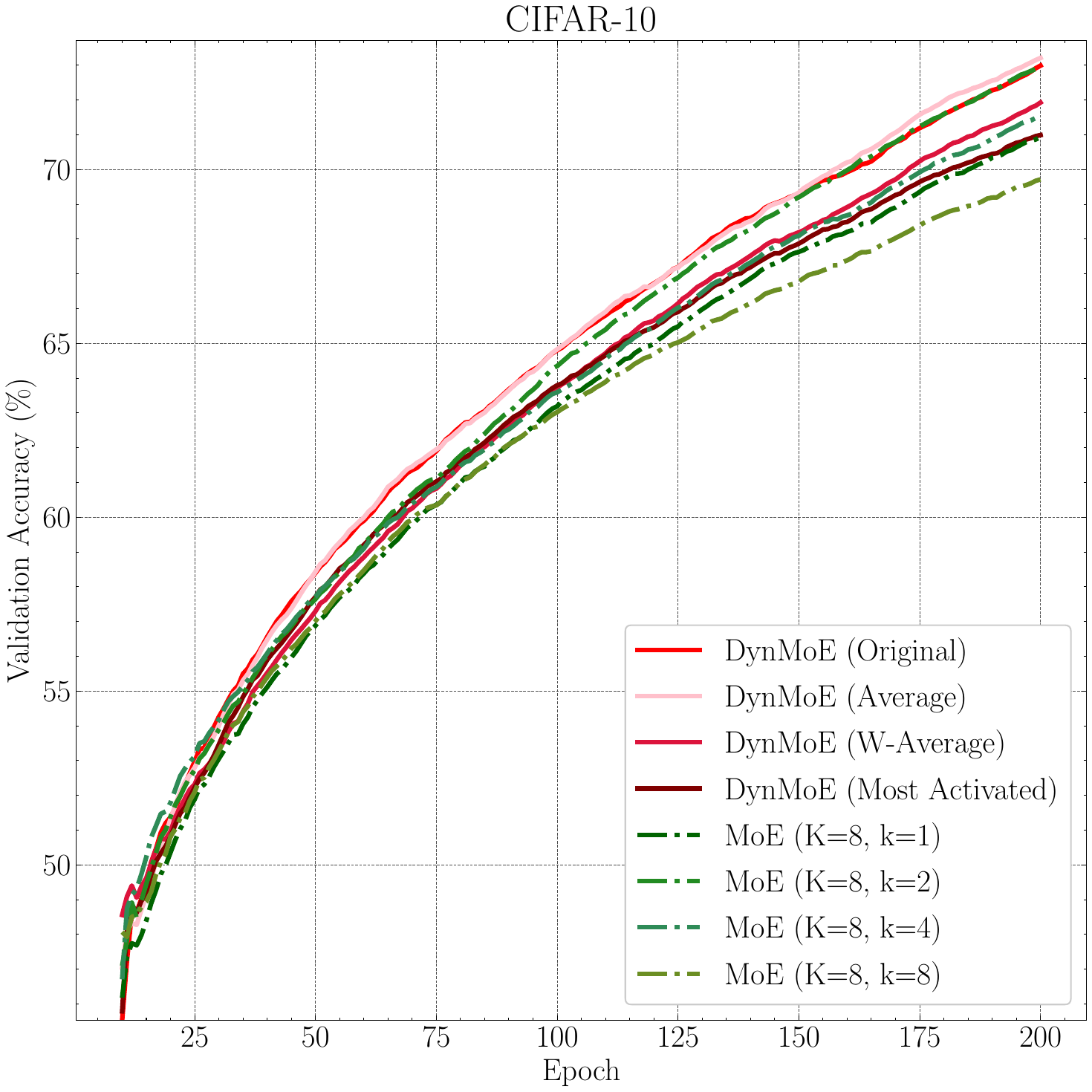}
    \caption{\textbf{Convergence curve on CIFAR10 and TinyImageNet datyasets.} }
    \label{fig:convergence}
\end{figure}

\begin{table}[!t]
    \centering
    \caption{
        \textbf{Efficiency evaluation of \algabbr comparing to MoE-LLaVA.} We conduct experiments on single A100 GPU (80 GB) paired with 16 CPUs using identical environment and identical training/inference configurations. 
        We report the performance of MoE-LLaVA using DeepSpeed's top-2 gating implementation.
        The symbols $\downarrow$ and $\uparrow$ indicate that lower and higher values, respectively, denote better performance.
    }
    \setlength{\tabcolsep}{2pt}
    {\fontsize{10}{8}\selectfont
    \resizebox{1.\textwidth}{!}{  
    \begin{tabular}{l c c c c c}  
        \toprule  
        Model & Training FLOPs $\downarrow$ & Inference FLOPs $\downarrow$ & Inference MACs $\downarrow$ & Memory Usage $\downarrow$ \\[1.5pt]
        & (TFLOPs/step) & (GFLOPs/token) & (GMACs/token) & (GB) \\
        \midrule  
        MoE-LLaVA (StableLM) & 18.23 & 27.62 & 13.34 & 5.98\\
        DynMoE-LLaVA (StableLM) & 17.97 & 25.25 & 12.13 & 5.98 \\
        \midrule  
        MoE-LLaVA (Qwen) & 34.27 & 23.36 & 11.30 & 6.37 \\
        DynMoE-LLaVA (Qwen, Ours) & 34.61 & 22.17 & 10.73 & 6.37 \\
        \midrule  
        MoE-LLaVA (Phi-2) & 63.43 & 46.87 & 22.73 & 10.46 \\
        DynMoE-LLaVA (Phi-2) & 63.36 & 44.92 & 21.72 & 10.46 \\
        \bottomrule  
    \end{tabular}  
    }}
    \label{tab:inference-efficiency-appendix}
\end{table}

\end{document}